\documentclass[10pt,twocolumn,letterpaper]{article}

\usepackage[pagenumbers]{cvpr} %

\usepackage{multirow}
\usepackage{colortbl}
\usepackage{tabularx}
\definecolor{componentrow}{gray}{0.93}
\definecolor{ntHeader}{HTML}{EEF1F4}
\definecolor{ntMethod}{HTML}{E6EEF7}
\definecolor{ntMethodSoft}{HTML}{E6EEF7}
\definecolor{ntRule}{HTML}{000000}
\newcolumntype{G}{>{\cellcolor{ntMethodSoft}}r}
\newcolumntype{M}{>{\cellcolor{ntMethodSoft}}c}
\newcolumntype{L}{>{\cellcolor{ntMethodSoft}}l}
\newcolumntype{Y}{>{\centering\arraybackslash}X}
\newenvironment{nicetable}{\arrayrulecolor{black}%
  \setlength{\aboverulesep}{0pt}\setlength{\belowrulesep}{0pt}%
  \setlength{\extrarowheight}{1.2pt}}{}
\newcommand{\tintrule}{\noalign{\vskip-\cmidrulewidth}}
\newcommand{\venue}[1]{{\fontsize{5.5}{6.5}\selectfont\color{black!60}#1}}
\usepackage{makecell}
\usepackage{placeins}
\usepackage{amsthm}
\newtheorem{proposition}{Proposition}
\DeclareRobustCommand{\pending}[1]{\textcolor{red}{??}}
\newcommand{\method}{GRAIN}
\newcommand{\scp}{SCP}%

\IfFileExists{tables/statistics_macros.tex}{\newcommand{\SWEEPmeanCOCOOne}{+0.93}

\newcommand{\SWEEPmeanCHOne}{+3.26}

\newcommand{\SWEEPmeanVDOne}{+0.65}

\newcommand{\SWEEPwins}{100}
\newcommand{\SWEEPBOOTpos}{70}
\newcommand{\SWEEPBOOTtot}{78}
\newcommand{\SWEEPdetectors}{17}
\newcommand{\SWEEPtotal}{102}
}{}
\providecommand{\SWEEPwins}{\pending{}}
\providecommand{\SWEEPmeanCOCOOne}{\pending{}}

\providecommand{\SWEEPmeanCHOne}{\pending{}}

\providecommand{\SWEEPmeanVDOne}{\pending{}}

\newcommand{\lws}{LWS}
\newcommand{\cgs}{CGS}
\newcommand{\sba}{SBA}
\newcommand{\prf}{PRF}
\newcommand{\figorbox}[4]{\IfFileExists{#1}{\includegraphics[width=#2]{#1}}{%
\fbox{\parbox[c][#3][c]{\dimexpr#2-2\fboxsep-2\fboxrule\relax}{\centering\small #4}}}}

\AtEndPreamble{%
  \crefname{figure}{Figure}{Figures}%
  \Crefname{figure}{Figure}{Figures}%
}

\definecolor{cvprblue}{rgb}{0.21,0.49,0.74}
\usepackage[pagebackref,breaklinks,colorlinks,allcolors=cvprblue]{hyperref}

\def\confName{CVPR}
\def\confYear{2027}

\title{GRAIN: Shared-Budget Complementary Re-observation for Object Detection}

\author{Chengxuan Zhou$^{1}$\quad Shaoyang Zhang$^{2}$\quad Rui Qian$^{3}$\quad Juntao Cheng$^{4}$\quad Feng Liu$^{1}$\quad Wanru Song$^{1}$\\
{\small $^{1}$Nanjing University of Posts and Telecommunications\quad $^{2}$Hohai University\quad $^{3}$Fudan University\quad $^{4}$Shanghai Jiaotong University}
}

\begin{document}
\twocolumn[{%
\renewcommand\twocolumn[1][]{#1}%
\maketitle
\begin{center}

\includegraphics[width=\textwidth]{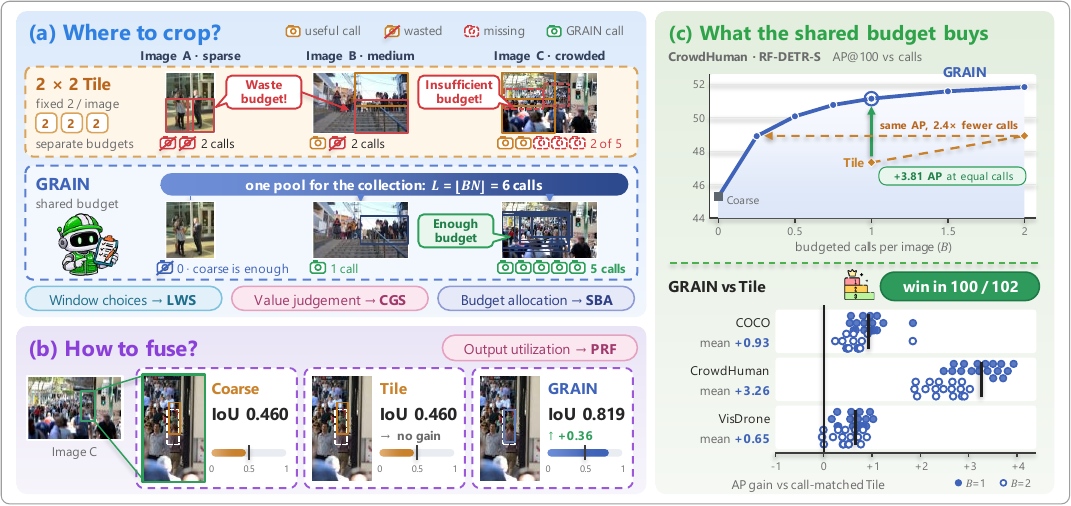}
\captionsetup{hypcap=false}\captionof{figure}{\textbf{Spending Re-observation Where It Pays.} (a)~On three CrowdHuman validation images planned at $B=2$, Tile allocates 2, 2, and 2 crops regardless of content, whereas \method{} allocates 0, 1, and 5 crops. (b)~In the crowded image, the best match for an occluded person rises from IoU 0.46 under Coarse and Tile to 0.82 under \method{}. (c)~In the top plot, RF-DETR-S with \method{} at $B=\frac14$ comes within 0.02 AP@100 of Tile at $B=2$ on CrowdHuman using 2.4$\times$ fewer total calls. In the bottom plot, vertical lines mark mean gains at $B=1$ across \SWEEPdetectors{} detectors on COCO, CrowdHuman, and VisDrone-DET, where \method{} wins \SWEEPwins{} of \SWEEPtotal{} settings.}
\label{fig:teaser}
\end{center}%
}]
\begin{abstract}
Re-running a frozen detector on enlarged image crops recovers small and crowded objects without retraining, yet every crop costs an extra detector call. Existing slicing and adaptive-cropping pipelines plan crops per image in isolation, even though inspection demand varies widely across an image collection. We present \method{}, which unifies cross-image call allocation and bounded-output admission around cross-image comparable resolution-closure gains over boundary-safe crop interiors. Coarse hypotheses carry resolution deficits, crops close a geometric share of those deficits for hypotheses enclosed in their interior positive region, and multiplicative residuals discount covered evidence within a monotone submodular surrogate. Prefix sums and a global max-heap execute this collection-wide plan before crop inference, and Positive-Region Fusion aligns post-inference spatial admission with the planned interior support before class-balanced top-$K$ selection. Across \SWEEPdetectors{} frozen detectors from five families on COCO, CrowdHuman, and VisDrone-DET, \method{} beats call-matched tiling in \SWEEPwins{} of \SWEEPtotal{} AP@100 settings and transfers to AI-TOD-v2. With RF-DETR-S on CrowdHuman, it nearly matches two-crop tiling using one eighth of the crop calls. At fixed budgets, the surrogate orders candidate plans with Spearman correlations of 0.95 to 0.99 while the sweep attains at least 98\% of the exact surrogate optimum. Code will be publicly released.
\end{abstract}
    
\section{Introduction}
Small and densely packed objects remain a primary bottleneck for modern detectors~\cite{cheng2023towards}. Recent works refine label assignment in DEIM~\cite{huang2025deim} and Mr.~DETR~\cite{zhang2025mr}, dynamic queries in PaQ-DETR~\cite{kang2026paq}, real-time routing in YOLO-Master~\cite{lin2026yolo}, and tiny-object features in SET~\cite{sun2025set}, Gaussian modeling~\cite{bian2025feature}, and BDNet~\cite{guan2026bdnet}, yet fixed input resolution caps pixel support: small instances span only a few pixels, and feature pyramids~\cite{lin2017feature} recover only part of the lost detail. Re-evaluating a frozen detector on enlarged crops restores fine detail without retraining~\cite{unel2019power,akyon2022sahi}, but every crop costs an extra detector call.

\begin{figure*}[t]
\centering
\includegraphics[width=\textwidth]{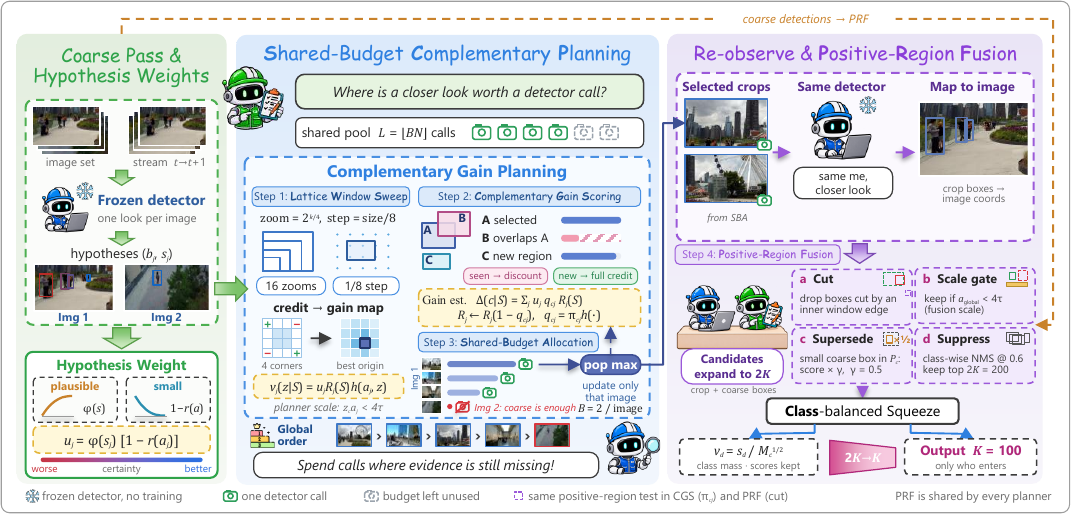}
\caption{\textbf{Overview of \method{}.} A frozen detector first runs one coarse pass over the image collection. \scp{} plans all crop calls before crop inference, as \lws{} builds multi-zoom gain maps from the residual credits of \cgs{} and \sba{} assigns each call to the largest collection gain while refreshing only the chosen image. After the same detector evaluates the selected crops, \prf{} filters inferred crop boxes with the planned interior region, merges views into at most $2K$ candidates, and selects $K$ class-balanced outputs per image.}
\label{fig:overview}
\end{figure*}

Existing training-free pipelines plan crops one image at a time. Uniform slicing in SAHI~\cite{akyon2022sahi} applies a fixed grid to every image, while content-adaptive methods such as DAHI~\cite{suarez2026dahi}, Dynamic Tiling~\cite{nguyen2023dynamic}, and ViCrop-Det~\cite{wang2026vicrop} select crops from single-image density, detection, or attention cues and merge views via suppression or border pruning~\cite{bodla2017softnms,najibi2019autofocus} decoupled from crop selection. Yet inspection demand varies sharply across images in Figure~\ref{fig:teaser}: sparse scenes are resolved by the coarse pass whereas crowded scenes need several zoomed views, so per-image budgets waste calls on easy images and starve dense ones. Across offline collections and batched streams, one shared budget can flow to unresolved scenes.

Sharing a collection budget couples three decisions treated separately in Figure~\ref{fig:teaser}: generating candidate windows, judging marginal crop utility given prior picks, and assigning the next call across images. Inspired by budgeted coverage, we formulate our first research question: \emph{1) \textbf{How can we define a cross-image comparable marginal gain that jointly guides multi-zoom window search, overlap discounting, and shared call allocation across an image collection?}} However, zoomed crops raise accuracy only if their boxes survive into a bounded output set, where boundary-truncated fragments otherwise displace intact predictions in Figure~\ref{fig:teaser}. This motivates our second research question: \emph{2) \textbf{How can we align multi-view spatial admission with planned crop interiors so intact zoomed detections populate a bounded output capacity?}} Taken together, these considerations motivate \method{} to couple cross-image coverage valuation with boundary-aligned fusion.

As shown in Figure~\ref{fig:overview}, \method{} grounds re-observation on cross-image comparable resolution-closure gains over boundary-safe crop interiors. Coarse hypotheses carry resolution deficits, crops close a geometric share of those deficits for interior hypotheses, and multiplicative residuals keep marginal gains calibrated across images and zooms inside a monotone submodular surrogate. Addressing the first question, Shared-Budget Complementary Planning (\scp{}) pairs Complementary Gain Scoring (\cgs{}) with Lattice Window Sweep (\lws{}) and Shared-Budget Allocation (\sba{}), executed via 2D prefix sums and a global max-heap before crop inference. Addressing the second, Positive-Region Fusion (\prf{}) enforces matching interior admission on inferred crop boxes before suppression and class-balanced top-$K$ selection. Our contributions are:
\begin{itemize}[leftmargin=*,nosep]
\item We formulate budgeted collection re-observation around cross-image comparable interior deficit closure and pose two coupled research questions on shared call allocation and boundary-aligned fusion.
\item We propose \method{}, pairing \scp{} (\lws{}, \cgs{}, and \sba{}) with \prf{} under a shared interior containment rule to plan multi-zoom crops and fuse views on any frozen detector.
\item Benchmarks across \SWEEPdetectors{} frozen detectors on COCO, CrowdHuman, and VisDrone-DET win \SWEEPwins{} of \SWEEPtotal{} AP@100 settings and transfer to AI-TOD-v2, supported by end-to-end, streaming, and surrogate diagnostics.
\end{itemize}

\section{Related Work}
\noindent \textbf{Training-Free Re-observation.}
Re-applying a detector to enlarged subregions recovers small objects without modifying weights~\cite{unel2019power}. Uniform slicing in SAHI~\cite{akyon2022sahi} and ASAHI~\cite{zhang2023adaptive} partitions every image into overlapping grids, tying cost to image size rather than content. Content-adaptive pipelines focus compute within one image by estimating density in DAHI~\cite{suarez2026dahi} and DAASD~\cite{liu2026daasd}, gating tiles via proposals in ROI-Gated SAHI~\cite{riyadh2026roi}, refining split boxes in Dynamic Tiling~\cite{nguyen2023dynamic}, or thresholding attention entropy in ViCrop-Det~\cite{wang2026vicrop}. Planning images in isolation prevents shifting calls from sparse to crowded scenes or discounting multi-zoom overlap on a unified scale. \method{} calibrates window gains across all images and zooms under one budget and upgrades external pools in Sec.~\ref{sec:experiments}.

\smallskip \noindent \textbf{Scale-Aware and Learned Re-observation.}
Scale variation motivates multi-resolution pyramids~\cite{lin2017feature,singh2018analysis}. Chip cascades such as SNIPER~\cite{singh2018sniper} and AutoFocus~\cite{najibi2019autofocus} prune background regions during training or inference, while learned proposers regress zoom regions from density maps in DMNet~\cite{li2020density} and coarse estimators~\cite{duan2021coarse}, clusters in ClusDet~\cite{yang2019clustered}, or differentiable warps~\cite{wang2026adaptive}. Active-perception policies learn zoom selection via reinforcement learning in Dynamic Zoom-in~\cite{gao2018dynamic}, PatchDrop~\cite{uzkent2020learning}, AdaZoom~\cite{xu2022adazoom}, grounded cropping~\cite{zhao2026learning}, and ZoomEarth~\cite{liu2026zoomearth}, whereas architectural designs embed refinement via cascaded crops in CZ~Det~\cite{meethal2023cascaded}, sparse queries in QueryDet~\cite{yang2022querydet}, patch packing in UFPMP-Det~\cite{huang2022ufpmp}, dynamic routing~\cite{mnih2014recurrent,figurnov2017spatially,wang2020glance,han2022dynamic,qin2025no}, or early exits in MSDNet~\cite{huang2018multi} and edge schedulers~\cite{jiang2021flexible}. These methods require retraining or custom heads and act per input, whereas \method{} shares crop calls of one frozen detector across an image collection.

\smallskip \noindent \textbf{Submodular Selection and Window Search.}
Selecting crops to resolve uncovered objects forms a budgeted coverage problem. Greedy maximization of monotone submodular functions yields a $1-1/e$ bound~\cite{nemhauser1978analysis,krause2014submodular} accelerated by lazy queues~\cite{minoux1978accelerated,leskovec2007cost}, while Efficient Subwindow Search~\cite{lampert2008beyond} locates one window per image via branch and bound and adaptive submodularity~\cite{golovin2011adaptive} targets sequential replanning. Generic coverage, however, neither calibrates cross-scale detection value nor scales over dense zoom lattices. \method{} constructs a resolution-closure surrogate whose separable structure lets \sba{} refresh one image per step while \lws{} factorizes interior containment into 2D prefix-sum sweeps~\cite{crow1984summed,viola2001rapid}.

\smallskip \noindent \textbf{Fusion of Repeated Observations.}
Overlapping crops create duplicate and boundary-truncated boxes that compete for bounded output slots. Single-view rules such as Soft-NMS~\cite{bodla2017softnms}, Adaptive NMS~\cite{liu2019adaptive}, learned suppression~\cite{hosang2017learning}, and Weighted Boxes Fusion (WBF)~\cite{solovyev2021weighted} ignore crop geometry, while multi-view pipelines use slice merging in SAHI~\cite{akyon2022sahi}, border bands in AutoFocus~\cite{najibi2019autofocus} and DAHI~\cite{suarez2026dahi}, expert calibration in MoCaE~\cite{oksuz2023mocae}, or dense-map fusion in SCAHI~\cite{becker2026scahi}, which requires dense probability heads absent in sparse query detectors. \prf{} uses only sparse boxes and scores, aligning interior admission with crop planning before class-balanced top-$K$ selection.

\section{Method}
\label{sec:method}\label{sec:prelim}

\noindent\textbf{Problem Statement.}
Given $N$ images, a frozen detector runs one coarse pass per image returning hypotheses $\{(b_j,s_j)\}$ with boxes $b_j$ and confidences $s_j$. Re-observation allocates at most $L=\lfloor BN\rfloor$ extra calls to enlarged crops across the collection, where mean budget $B$ may be fractional so easy scenes receive zero crops and dense ones receive several. The same detector evaluates selected crops, merging views into at most $K$ outputs per image via boundary-aligned admission.

\smallskip\noindent\textbf{Overview.}\label{sec:overview}
In Figure~\ref{fig:overview}, \method{} plans over image collections while supporting batched and causal streaming execution in Sec.~\ref{sec:online}. Section~\ref{sec:coverage} constructs a cross-image comparable deficit-closure surrogate over coarse hypotheses, Sec.~\ref{sec:scp} executes multi-zoom planning via \scp{} before crop inference, and Sec.~\ref{sec:prf} merges views through \prf{} under a shared interior rule.

\subsection{Re-observation as Evidence Coverage}
\label{sec:coverage}

\noindent\textbf{Hypothesis Weight.}
Planning extracts an untruncated pool of coarse hypotheses surviving a low confidence threshold and class-wise NMS~\cite{neubeck2006efficient}. Empty scenes trigger no calls, whereas weak or crowd-merged responses cue local crops whose hypothesis-free admission in Sec.~\ref{sec:prf} also recovers adjacent missed instances in Sec.~\ref{sec:planner-analysis}. To prioritize unresolved candidates without chasing low-score clutter, we pair confidence-modulated salience with pixel-support saturation $r(a)=a^2/(a^2+\tau^2)$ at target scale $\tau$. At input scale $a_j=\sqrt{A_jW_DH_D/(W_iH_i)}$ for box $b_j$ of area $A_j$, image size $(W_i,H_i)$, and detector input $(W_D,H_D)$, hypothesis $j$ carries weight
\begin{equation}
u_j=\phi(s_j)\,[1-r(a_j)],\qquad \phi(s)=\sqrt{s}\,(1-s/2).
\label{eq:weight}
\end{equation}
Here $1-r(a_j)$ measures coarse resolution deficit, while $\phi(s_j)$ suppresses background noise, peaks at ambiguous confidence $s_j=2/3$, and retains about half weight at $s_j=1$ so confident crowd hypotheses still cue local~zoom.

\smallskip\noindent\textbf{Observation Quality.}
A window $c$ at zoom $z_c$ magnifies enclosed objects by $z_c$, while its positive region $\mathcal P_c$ contracts $c$ inward by two pixels on inner crop edges and exempts image borders, valuing hypothesis $j$ via
\begin{equation}
q_{cj}=\pi_{cj}\,h(a_j,z_c),\qquad \pi_{cj}=\mathbb{1}[\,b_j\subseteq\mathcal P_c\,],
\label{eq:quality}
\end{equation}
where full-box containment in $\mathcal P_c$ excludes inner-edge fragments without shrinking interior support. Derived directly from $r(\cdot)$ without extra shape parameters, closure $h(a,z)$ is
\begin{equation}
h(a,z)=\begin{cases}
\dfrac{(z^2-1)a^2}{z^2a^2+\tau^2},&za<4\tau,\\[6pt]
0,&za\ge4\tau.
\end{cases}
\label{eq:closure}
\end{equation}
For $za<4\tau$, $h(a,z)=[r(za)-r(a)]/[1-r(a)]$ telescopes $u_jh(a_j,z_c)=\phi(s_j)[r(z_ca_j)-r(a_j)]$ into added resolution credit, while the $4\tau$ guard avoids over-magnifying objects past $16/17$ saturation. Only $\tau$ varies across~benchmarks.

\smallskip\noindent\textbf{Coverage Objective.}
Motivated by diminishing-returns set coverage~\cite{nemhauser1978analysis,krause2014submodular}, which lacks cross-scale detectability calibration, we cast collection re-observation around a deterministic geometric surrogate:
\begin{equation}
F(S)=\sum_j u_j\Bigl(1-\prod_{c\in S}(1-q_{cj})\Bigr).
\label{eq:coverage}
\end{equation}
Rather than assuming independent hit probabilities, $q_{cj}\in[0,1]$ encodes fractional deficit closure and $\prod_{c\in S}(1-q_{cj})$ tracks remaining unresolved demand before crop inference. Over window pools $\mathcal C_i$, planning solves
\begin{equation}
\max_{S\subseteq\bigcup_i\mathcal C_i}\ F(S)\quad\text{subject to}\quad |S|\le L.
\label{eq:sba-objective}
\end{equation}

\begin{proposition}
\label{prop:greedy}
For fixed $u_j\ge0$ and $q_{cj}\in[0,1]$, $F$ is monotone submodular. Exact greedy selection under $|S|\le L$ returns $S_g$ satisfying $F(S_g)\ge(1-1/e)\,F(S^\star)$ relative to an optimal subset $S^\star$ over the same pool.
\end{proposition}
Proposition~\ref{prop:greedy}, proved in Appendix~\ref{sec:proof}, establishes greedy near-optimality on the surrogate $F$ over any fixed candidate pool, while Section~\ref{sec:planner-analysis} empirically verifies both \lws{}'s tightness against the exact optimum on $F$ and $F$'s fixed-budget rank fidelity to~AP@100.

\subsection{Shared-Budget Complementary Planning}
\label{sec:scp}
Submodularity alone neither calibrates multi-zoom utility across scenes nor searches dense grids efficiently. \scp{} couples \cgs{} with \sba{} and \lws{} to achieve~both.

\smallskip\noindent\textbf{Complementary Gain Scoring.}
For hypothesis $j$, \cgs{} tracks residual demand $R_j(S)=\prod_{d\in S}(1-q_{dj})$:
\begin{equation}
R_j(\varnothing)=1,\qquad R_j(S\cup\{c\})=R_j(S)(1-q_{cj}).
\label{eq:cgs-state}
\end{equation}
Evaluating $\Delta(c\mid S)=F(S\cup\{c\})-F(S)$ gives
\begin{equation}
\Delta(c\mid S)=\sum_j u_j\,q_{cj}\,R_j(S)\,.
\label{eq:gain}
\end{equation}
Equation~\eqref{eq:gain} unifies deficit $u_j$, interior closure $q_{cj}$, and residual demand $R_j(S)$ on one cross-image scale. Rounded windows never repeat, and contracting $R_j$ by $1-q_{cj}$ steers calls away from near-duplicates without hard overlap bans that block complementary crowd views in Sec.~\ref{sec:analysis}.

\smallskip\noindent\textbf{Shared-Budget Allocation.}
\label{sec:sba}
Because windows observe only their host image, $F$ separates additively across images and gains under \cgs{} depend solely on $S_i=S\cap\mathcal C_i$, factoring each step into local search and global selection:
\begin{equation}
\begin{aligned}[t]
c_i^\star&\in\arg\max\nolimits_{c\in\mathcal C_i\setminus S_i}\Delta(c\mid S_i),\\
i^\star&\in\arg\max\nolimits_{i}\Delta(c_i^\star\mid S_i),\quad
S_{i^\star}\leftarrow S_{i^\star}\cup\{c_{i^\star}^\star\}.
\end{aligned}
\label{eq:sba-step}
\end{equation}
\sba{} indexes each image's top window in a lazy max-heap~\cite{minoux1978accelerated}, updating residuals and refreshing only $i^\star$ in $O(\log N)$ heap time with linear memory up to $Q_B\le L$ calls. Greedy order nests plans in $B$, admits per-image caps for tail latency, and induces a gain threshold for streaming in Sec.~\ref{sec:online}.

\smallskip\noindent\textbf{Lattice Window Sweep.}\label{sec:lws}
Inspired by integral-image accumulation~\cite{crow1984summed,viola2001rapid}, \lws{} solves the inner search of Eq.~\eqref{eq:sba-step} across zooms $\mathcal Z=\{2^{k/4}:k=1,\ldots,16\}$ by factorizing scale closure from spatial containment. On image $(W_i,H_i)$, zoom-$z$ windows of size $(W_i/z,H_i/z)$ step by one eighth of a side with clipped borders, giving origins whose $\mathcal P_c$ encloses $b_j$ credit
\begin{equation}
v_j(z\mid S)=u_j\,R_j(S)\,h(a_j,z).
\label{eq:sweep-credit}
\end{equation}
In origin space, $b_j\subseteq\mathcal P_c$ forms a dual axis-aligned rectangle located by binary search, turning boundary-aware scoring across $G_z$ origins into four corner impulses per hypothesis and one 2D sweep in $O(n\log G_z+G_z)$ time instead of $O(nG_z)$ pairs. Scanning $z\in\mathcal Z$ selects position and zoom, centers near-ties, and recomputes rounded gain before heap insertion, and \cgs{} with \sba{} upgrades external pools in~Sec.~\ref{sec:experiments}.

\subsection{Positive-Region Fusion}
\label{sec:prf}
Unlike border-pruning and slice-merging rules~\cite{najibi2019autofocus,akyon2022sahi} that filter crops independently of planning, \prf{} enforces a geometric contract between planned credit and bounded-output admission. At full-image scale $a_{\rm g}(b)=\sqrt{|b|W_DH_D/(W_iH_i)}$ for mapped box $b$, \prf{} \emph{expands} views to $2K$ candidates and \emph{squeezes} them to~$K$.

\smallskip\noindent\textbf{Expand.}
\begin{enumerate}[leftmargin=*,nosep]
\item \emph{Cut.} Mapped crop boxes require $b\subseteq\mathcal P_c$ from Eq.~\eqref{eq:quality}, rejecting edge-clamped fragments via full-box containment while admitting all intact interior detections, including newly resolved objects without coarse~matches.
\item \emph{Scale Gate.} Mapped crop boxes require $a_{\rm g}(b)<4\tau$, leaving large objects to the coarse~pass.
\item \emph{Supersede.} Coarse boxes with $a_{\rm g}(b)<2\tau$ inside an executed $\mathcal P_c$ have scores scaled by $\gamma=0.5$, ceding NMS priority to sharper crop detections while surviving as fallbacks if a crop~misses.
\item \emph{Suppress.} Coarse and crop boxes undergo class-wise NMS at IoU 0.6, keeping top-$2K$ candidates~$\mathcal D_{2K}^{(i)}$.
\end{enumerate}
Sharing $\mathcal P_c$ and $\tau$, planning gates coarse hypotheses at projected crop scale $z_ca_j<4\tau$, whereas \prf{} gates inferred crop boxes at full-image scale $a_{\rm g}(b)<4\tau$.

\smallskip\noindent\textbf{Squeeze.}
Multi-crop zoom multiplies frequent crowded classes and can evict rare ones under raw top-$K$ truncation. For candidate $d$ of predicted class $k_d$, squeezing prioritizes output slots by collection class mass:
\begin{equation}
M_k=\sum_{i=1}^{N}\sum_{d\in\mathcal D_{2K}^{(i)}}^{k_d=k}s_d,\qquad
v_d=\frac{s_d}{\sqrt{\max(M_{k_d},\epsilon_M)}},
\label{eq:squeeze}
\end{equation}
with $\epsilon_M=10^{-6}$. Each image retains the $K=100$ candidates with largest $v_d$ while outputting original fused scores $s_d$ unchanged, preserving calibration and reducing to score top-$K$ on single-class datasets. Executed on CPU boxes and scores, \prf{} keeps $\gamma$, NMS IoU, and the squeeze exponent fixed alongside~$\tau$.

\section{Experiments}
\label{sec:experiments}

\begin{table*}[t]
\caption{\textbf{Main Results.} We report full-validation AP@100 for \SWEEPdetectors{} frozen detectors including Large variants on COCO, CrowdHuman, and VisDrone-DET. Coarse runs the detector once, whereas Tile applies $2\times2$ SAHI slicing~\cite{akyon2022sahi}. Both Tile and \method{} share \prf{} and spend $B$ crop calls per image on average. Bold marks the stronger method per budget, including rounded ties. Wins and summary statistics use unrounded AP across all \SWEEPdetectors{} detectors.}
\label{tab:family}
\centering\scriptsize

\setlength{\tabcolsep}{3.2pt}
\resizebox{\textwidth}{!}{\begin{nicetable}
\begin{tabular}{ll>{\hspace{5pt}}rrrGG>{\hspace{5pt}}rrrGG>{\hspace{5pt}}rrrGG}
\toprule
\rowcolor{ntHeader}
 & & \multicolumn{5}{c}{COCO val2017} & \multicolumn{5}{c}{CrowdHuman validation} & \multicolumn{5}{c}{VisDrone-DET validation} \\
\tintrule\cmidrule(lr){3-7}\cmidrule(lr){8-12}\cmidrule(lr){13-17}
\rowcolor{ntHeader}
Family & Size & & \multicolumn{2}{c}{Tile} & \multicolumn{2}{c}{\cellcolor{ntMethod}\method{}} & & \multicolumn{2}{c}{Tile} & \multicolumn{2}{c}{\cellcolor{ntMethod}\method{}} & & \multicolumn{2}{c}{Tile} & \multicolumn{2}{c}{\cellcolor{ntMethod}\method{}} \\
\rowcolor{ntHeader}
 & & \multirow{-2}{*}{Coarse} & $B{=}1$ & $B{=}2$ & \cellcolor{ntMethod}$B{=}1$ & \cellcolor{ntMethod}$B{=}2$ & \multirow{-2}{*}{Coarse} & $B{=}1$ & $B{=}2$ & \cellcolor{ntMethod}$B{=}1$ & \cellcolor{ntMethod}$B{=}2$ & \multirow{-2}{*}{Coarse} & $B{=}1$ & $B{=}2$ & \cellcolor{ntMethod}$B{=}1$ & \cellcolor{ntMethod}$B{=}2$ \\
\midrule
\multirow{4}{*}{RF-DETR~\cite{robinson2026rf} \venue{(ICLR'26)}} & N & 47.90 & 50.54 & 51.75 & \textbf{52.39} & \textbf{53.59} & 44.18 & 46.37 & 47.96 & \textbf{50.08} & \textbf{50.82} & 23.12 & 28.16 & 29.96 & \textbf{28.87} & \textbf{30.48} \\
 & S & 52.65 & 54.49 & 55.33 & \textbf{55.73} & \textbf{56.24} & 45.27 & 47.32 & 48.91 & \textbf{51.13} & \textbf{51.82} & 23.38 & 28.63 & 30.46 & \textbf{29.46} & \textbf{30.84} \\
 & M & 54.53 & 55.84 & 56.46 & \textbf{56.94} & \textbf{57.24} & 45.74 & 47.80 & 49.32 & \textbf{51.48} & \textbf{52.18} & 23.68 & 28.77 & 30.52 & \textbf{29.53} & \textbf{31.29} \\
 & L & 56.37 & 57.06 & 57.36 & \textbf{58.02} & \textbf{58.08} & 46.88 & 48.65 & 50.08 & \textbf{52.15} & \textbf{52.76} & 25.11 & 29.68 & \textbf{31.24} & \textbf{29.97} & 31.23 \\
\midrule
\multirow{4}{*}{D-FINE~\cite{peng2025d} \venue{(ICLR'25)}} & N & 42.60 & 43.98 & 44.75 & \textbf{45.08} & \textbf{45.47} & 41.38 & 44.04 & 45.82 & \textbf{47.97} & \textbf{48.84} & 19.60 & 22.69 & 23.85 & \textbf{23.65} & \textbf{24.50} \\
 & S & 48.28 & 48.98 & 49.40 & \textbf{49.54} & \textbf{49.87} & 46.14 & 47.93 & 49.41 & \textbf{51.28} & \textbf{51.88} & 23.17 & 26.17 & 27.38 & \textbf{26.78} & \textbf{27.79} \\
 & M & 52.07 & 52.71 & 53.18 & \textbf{53.72} & \textbf{53.89} & 48.76 & 50.22 & 51.57 & \textbf{53.43} & \textbf{53.95} & 26.83 & 29.85 & 31.02 & \textbf{30.41} & \textbf{31.32} \\
 & L & 53.77 & 54.53 & 54.96 & \textbf{55.21} & \textbf{55.35} & 49.85 & 51.43 & 52.79 & \textbf{54.61} & \textbf{55.11} & 26.73 & 30.38 & 31.69 & \textbf{31.12} & \textbf{32.09} \\
\midrule
\multirow{3}{*}{RT-DETRv4~\cite{liao2026rt} \venue{(ECCV'26)}} & S & 49.63 & 50.25 & 50.66 & \textbf{51.06} & \textbf{51.18} & 47.80 & 49.17 & 50.44 & \textbf{52.23} & \textbf{52.76} & 24.70 & 27.70 & 28.80 & \textbf{28.00} & \textbf{28.99} \\
 & M & 53.44 & 53.94 & 54.25 & \textbf{54.60} & \textbf{54.74} & 50.06 & 51.04 & 52.19 & \textbf{53.94} & \textbf{54.38} & 27.09 & 29.97 & 31.17 & \textbf{30.77} & \textbf{31.62} \\
 & L & 55.23 & 55.84 & 56.28 & \textbf{56.38} & \textbf{56.52} & 50.67 & 51.73 & 52.89 & \textbf{54.42} & \textbf{54.79} & 26.97 & 30.00 & 30.99 & \textbf{30.55} & \textbf{31.54} \\
\midrule
\multirow{2}{*}{ECDet~\cite{liu2026edgecrafter} \venue{(TMLR'26)}} & S & 51.52 & 52.46 & 53.09 & \textbf{53.50} & \textbf{53.84} & 46.17 & 48.24 & 49.78 & \textbf{51.84} & \textbf{52.62} & 25.67 & 29.28 & 30.51 & \textbf{30.16} & \textbf{31.32} \\
 & M & 54.03 & 55.01 & 55.41 & \textbf{55.63} & \textbf{55.88} & 48.01 & 49.79 & 51.31 & \textbf{53.30} & \textbf{53.98} & 24.69 & 28.38 & 29.66 & \textbf{29.41} & \textbf{30.55} \\
\midrule
\multirow{4}{*}{YOLO26~\cite{jocher2026yolo26} \venue{(arXiv'26)}} & N & 40.17 & 41.25 & 41.88 & \textbf{42.26} & \textbf{42.59} & 41.50 & 43.44 & 44.78 & \textbf{46.83} & \textbf{47.43} & 16.10 & 19.29 & 20.60 & \textbf{20.16} & \textbf{21.31} \\
 & S & 47.66 & 48.67 & 49.31 & \textbf{49.67} & \textbf{49.93} & 45.57 & 46.97 & 48.12 & \textbf{49.81} & \textbf{50.23} & 20.24 & 23.59 & \textbf{24.87} & \textbf{23.87} & \textbf{24.87} \\
 & M & 52.46 & 53.19 & 53.59 & \textbf{54.02} & \textbf{54.09} & 47.93 & 48.85 & 49.91 & \textbf{51.51} & \textbf{51.84} & 24.44 & 27.40 & \textbf{28.61} & \textbf{27.57} & 28.58 \\
 & L & 53.77 & 54.31 & 54.79 & \textbf{55.15} & \textbf{55.40} & 48.77 & 49.80 & 50.76 & \textbf{52.28} & \textbf{52.64} & 25.90 & 29.18 & 30.44 & \textbf{29.97} & \textbf{30.95} \\
\bottomrule
\end{tabular}
\end{nicetable}
 
 }
\end{table*}

\subsection{Experimental Setup}
\noindent \textbf{Datasets and Metrics.}
We evaluate on Microsoft COCO val2017~\cite{lin2014microsoft}, CrowdHuman validation~\cite{shao2018crowdhuman}, VisDrone-DET validation and test-dev~\cite{zhu2021detection}, and AI-TOD-v2 test~\cite{xu2022detecting}, reporting AP@100 at $K{=}100$ alongside VisDrone-DET AP$_S$ and CrowdHuman MR$^{-2}$~\cite{dollar2012pedestrian} and JI~\cite{chu2020detection}.

\smallskip \noindent \textbf{Baselines.}
Plans spend at most $L{=}\lfloor BN\rfloor$ crop calls. Because no baseline is global-budget-equivalent to ours, we construct Tile to select the $B\in\{1,2\}$ slices with largest summed coarse confidence from a $2{\times}2$ SAHI~\cite{akyon2022sahi} grid at 20\% overlap under shared \prf{}. DAHI~\cite{suarez2026dahi} and Dynamic Tiling~\cite{nguyen2023dynamic} are budget-adapted identically and tested with \cgs{} and~\sba{}.

\smallskip \noindent \textbf{Implementation Details.}
We evaluate \SWEEPdetectors{} frozen detectors across RF-DETR~\cite{robinson2026rf}, D-FINE~\cite{peng2025d}, RT-DETRv4~\cite{liao2026rt}, ECDet~\cite{liu2026edgecrafter}, and YOLO26~\cite{jocher2026yolo26}. Using RF-DETR-S, the sole domain scale $\tau$ is fixed once on training data to 24, 32, and 12 on COCO, CrowdHuman, and VisDrone-DET, and 12 on AI-TOD-v2 validation, leaving all other constants unchanged. Accuracy uses PyTorch, while end-to-end latency sums TensorRT FP16 passes and amortized CPU planning and fusion on an RTX 4090.

\subsection{Main Results}

\noindent \textbf{Comparison with Call-Matched Tiling.}
To isolate crop allocation across detector families, Table~\ref{tab:family} compares \method{} against call-matched Tile under shared \prf{}. \method{} wins \SWEEPwins{} of \SWEEPtotal{} settings, with positive paired bootstrap intervals in \SWEEPBOOTpos{} of \SWEEPBOOTtot{} 13-detector settings. At $B{=}1$, mean gains reach \SWEEPmeanCOCOOne{}, \SWEEPmeanCHOne{}, and \SWEEPmeanVDOne{} AP@100 on COCO, CrowdHuman, and VisDrone-DET, letting one-crop \method{} beat two-crop Tile for every detector on COCO and CrowdHuman while transferring to VisDrone-DET test-dev ($+0.92$ and $+0.70$ AP@100) and AI-TOD-v2 test (23.95 and 24.07 versus 23.43 and 23.60 for Tile at $B{=}1,2$). Redirecting calls via \cgs{} and \sba{} toward concentrated deficits $1-r(a_j)$ pays off most where crowd density varies sharply, whereas the two small losses within 0.03 occur at $B{=}2$ dense VisDrone-DET~scenes.

\begin{table}[t]
\caption{\textbf{Planner-Pool Comparison.} We evaluate frozen RF-DETR-S at $B=2$ on CrowdHuman and VisDrone-DET with shared \prf{}. Rows marked $^\dagger$ are budget-adapted, DAHI averages three seeds, and +CGS+SBA applies our scoring and allocation to each pool with mean end-to-end latency in milliseconds.}
\label{tab:similar-main}
\centering\scriptsize\setlength{\tabcolsep}{1.1pt}
\begin{nicetable}
\begin{tabularx}{\columnwidth}{lYcYYYcY}
\toprule
\rowcolor{ntHeader}
 & \multicolumn{4}{c}{CrowdHuman} & \multicolumn{3}{c}{VisDrone-DET} \\
\tintrule\cmidrule(lr){2-5}\cmidrule(l){6-8}
\rowcolor{ntHeader}
Method & AP$\uparrow$ & MR$^{-2}\downarrow$ & JI$\uparrow$ & ms & AP$\uparrow$ & AP$_S\uparrow$ & ms \\
\midrule
Tile~\cite{akyon2022sahi} & 48.91 & 74.10 & 76.63 & 8.1 & 30.46 & 21.65 & 8.4 \\
DAHI$^\dagger$~\cite{suarez2026dahi} & 49.00 & 73.60 & 76.68 & 13.1 & 29.35 & 20.87 & 14.7 \\
Dynamic Tiling$^\dagger$~\cite{nguyen2023dynamic} & 50.26 & 73.53 & 77.61 & 9.9 & 30.04 & 21.64 & 9.8 \\
\midrule
SAHI+CGS+SBA & 49.48 & 73.92 & 77.48 & 8.3 & \textbf{30.85} & 21.88 & 8.9 \\
DAHI+CGS+SBA & 51.55 & \textbf{71.65} & 78.61 & 17.6 & 30.71 & 22.81 & 21.2 \\
Dynamic Tiling+CGS+SBA & 50.84 & 72.99 & 78.45 & 11.4 & 30.82 & 22.72 & 11.1 \\
\midrule
\rowcolor{ntMethod}
\cellcolor{ntMethod}\textbf{\method{}} & \textbf{51.82} & 72.08 & \textbf{78.98} & 9.5 & 30.84 & \textbf{22.88} & 10.0 \\
\bottomrule
\end{tabularx}
\end{nicetable}

\end{table}

\smallskip \noindent \textbf{Comparison with Adaptive Cropping.}
Because adaptive proposers already focus windows within an image, Table~\ref{tab:similar-main} compares planning across pools under shared \prf{}. \method{} leads accuracy on CrowdHuman and AP$_S$ on VisDrone-DET at lower latency than density or refinement proposers, while adding \cgs{} and \sba{} lifts each baseline pool by 0.39 to 2.55 AP@100 and gives DAHI+\cgs{}+\sba{} the lowest miss rate. Single-image heuristics exhaust quotas on easy scenes and revisit resolved hypotheses, whereas deficit closure and residuals $R_j(S)$ suppress redundancy and reallocate calls across the split without proposer overhead.

\begin{table}[t]
\caption{\textbf{End-to-End Deployment Portability.} We compare complete training-free pipelines on frozen RF-DETR-S across COCO, CrowdHuman, and VisDrone-DET in AP@100, where Calls count all global, proposer, and crop passes and $^\dagger$ marks baselines adapted to frozen RF-DETR-S. Paired \method{} rows match each baseline's exact call total, including ViCrop-Det under default $\mu=0.3$ and training-only gate calibration.}
\label{tab:native-main}
\centering\scriptsize\setlength{\tabcolsep}{0.6pt}
\begin{nicetable}
\begin{tabularx}{\columnwidth}{lYYYYYY}
\toprule
\rowcolor{ntHeader}
 & \multicolumn{2}{c}{COCO} & \multicolumn{2}{c}{CrowdHuman} & \multicolumn{2}{c}{VisDrone-DET} \\
\tintrule\cmidrule(lr){2-3}\cmidrule(lr){4-5}\cmidrule(l){6-7}
\rowcolor{ntHeader}
Pipeline & AP & Calls & AP & Calls & AP & Calls \\
\midrule
Coarse & 52.65 & 1.00 & 45.27 & 1.00 & 23.38 & 1.00 \\
\rowcolor{ntMethod}
\method{} $B{=}2$ & 56.24 & 3.00 & 51.82 & 3.00 & 30.84 & 3.00 \\
\midrule
SAHI~\cite{akyon2022sahi} & 41.67 & 5.00 & 19.31 & 5.00 & 26.59 & 5.00 \\
\rowcolor{ntMethod}
\method{} $B{=}4$ & \textbf{56.49} & 5.00 & \textbf{52.07} & 5.00 & 31.97 & 5.00 \\
\midrule
DAHI~\cite{suarez2026dahi} & 32.28 & 10.62 & 26.12 & 9.42 & 25.86 & 11.62 \\
Dynamic Tiling~\cite{nguyen2023dynamic} & 39.15 & 11.01 & 38.36 & 11.43 & 31.83 & 11.66 \\
\rowcolor{ntMethod}
\method{} $B{=}10$ & 56.33 & 11.00 & 52.03 & 11.00 & \textbf{32.25} & 11.00 \\
\midrule
ASAHI$^\dagger$~\cite{zhang2023adaptive} & 33.46 & 6.92 & 28.01 & 7.97 & 29.55 & 7.21 \\
\rowcolor{ntMethod}
\method{} (ASAHI calls) & \textbf{56.47} & 6.92 & \textbf{52.06} & 7.97 & \textbf{32.32} & 7.21 \\
\midrule
ROI-Gated SAHI$^\dagger$~\cite{riyadh2026roi} & 49.39 & 1.99 & 27.72 & 14.58 & 28.44 & 7.79 \\
\rowcolor{ntMethod}
\method{} (ROI calls) & \textbf{55.72} & 1.99 & \textbf{51.99} & 14.58 & \textbf{32.35} & 7.79 \\
\midrule
ViCrop-Det (default)$^\dagger$~\cite{wang2026vicrop} & 47.74 & 1.06 & 39.54 & 1.04 & 22.59 & 1.00 \\
\rowcolor{ntMethod}
\method{} (ViCrop calls) & \textbf{53.52} & 1.06 & \textbf{46.63} & 1.04 & \textbf{23.69} & 1.00 \\
\midrule
DAASD$^\dagger$~\cite{liu2026daasd} & 51.98 & 2.14 & 30.62 & 10.70 & 29.26 & 8.34 \\
\rowcolor{ntMethod}
\method{} (DAASD calls) & \textbf{55.83} & 2.14 & \textbf{52.04} & 10.70 & \textbf{32.39} & 8.34 \\
\midrule
ViCrop-Det (calibrated)$^\dagger$ & 36.44 & 3.18 & 25.98 & 3.67 & 23.01 & 3.78 \\
\rowcolor{ntMethod}
\method{} (ViCrop calibrated) & \textbf{56.28} & 3.18 & \textbf{51.99} & 3.67 & \textbf{31.43} & 3.78 \\
\bottomrule
\end{tabularx}
\end{nicetable}

\end{table}

\smallskip \noindent \textbf{End-to-End Deployment Portability.}
To test complete plug-and-play deployment, Table~\ref{tab:native-main} evaluates pipelines retaining native proposal and merging rules at matched calls $T$ via $B=(T-N)/N$. \method{} wins every matched pair and beats 11.66-call Dynamic Tiling on VisDrone-DET in five calls, achieving 15.8\%, 25.9\%, and 77.1\% productive crop shares ($+0.96$, $+1.70$, and $+2.15$ net AP@100 per extra call) versus 11.8\%, 19.1\%, and 60.0\% ($-2.74$, $-6.49$, and $+0.80$) for native SAHI. In unaligned pipelines, border fragments and merged crowd boxes fill the 100-output cap and push accuracy below Coarse on COCO and CrowdHuman, whereas shared $\mathcal P_c$ admission in \prf{} protects intact coarse boxes as residuals $R_j(S)$ saturate near $B{=}4$, ensuring reliable end-to-end portability.

\subsection{Accuracy--Cost Trade-Off}

\begin{figure}[t]
\centering
\includegraphics[width=\columnwidth]{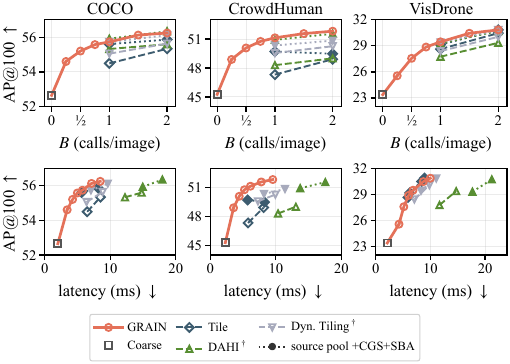}
\caption{\textbf{Accuracy versus Budget and Latency.} We plot AP@100 of frozen RF-DETR-S against crop calls on top and end-to-end TensorRT latency on the bottom across COCO, CrowdHuman, and VisDrone-DET. Dotted curves with filled markers add \cgs{} and \sba{} to baseline pools ($^\dagger$ budget-adapted).}
\label{fig:fractional}
\end{figure}

\noindent \textbf{Budget.}
Tight compute envelopes motivate testing fractional crop budgets across $B\in\{\frac14,\frac12,\frac34,1,1.5,2\}$ in the top row of Figure~\ref{fig:fractional}. Gains rise steepest below one call per image, where \method{} at $B{=}\frac14$ on CrowdHuman comes within 0.02 AP@100 of two-crop Tile at one eighth of its crop calls and nears two-crop Tile by $B{=}\frac12$ on COCO. Because \sba{}'s heap pops the largest collection gain $\Delta(c\mid S)$ first, early calls resolve severe deficits while leaving sparse scenes uncropped to front-load returns.

\smallskip \noindent \textbf{Latency.}
To measure runtime efficiency without file I/O noise, the bottom row of Figure~\ref{fig:fractional} and Table~\ref{tab:similar-main} report end-to-end latency per image. \method{} traces the upper-left Pareto frontier near 10\,ms per image at $B{=}2$, with single-thread CPU planning adding under 1.8\,ms even on dense VisDrone-DET scenes. Because \lws{} scores each zoom via four corner impulses and one 2D sweep while \sba{} refreshes only the chosen image $i^\star$, CPU planning and box-level \prf{} remain minor beside GPU inference, converting call savings directly into speedups.

\begin{table}[t]
\caption{\textbf{Equal-Cost Alternatives.} We compare AP@100 of frozen RF-DETR-S with input upscaling and larger models on COCO, CrowdHuman, and VisDrone-DET. Calls and Pixels are relative to one native pass, listing COCO first in RF-DETR-M$^{a}$ and RF-DETR-L$^{b}$, with Side $\times s$ scaling sides and $\cup$ merging by~NMS.}
\label{tab:equal-cost}
\centering\scriptsize\setlength{\tabcolsep}{1.8pt}
\begin{nicetable}
\begin{tabularx}{\columnwidth}{Xccccc}
\toprule
\rowcolor{ntHeader}
Pipeline & Calls & Pixels & COCO & CrowdHuman & VisDrone-DET \\
\midrule
Coarse (native) & 1 & 1 & 52.65 & 45.27 & 23.38 \\
Side $\times 1.25$ & 1 & 1.56 & 54.68 & 47.98 & 24.86 \\
Side $\times 1.5$ & 1 & 2.25 & 56.18 & 49.51 & 24.69 \\
Side $\times 2$ & 1 & 4 & \textbf{57.02} & 50.49 & 22.70 \\
Native $\cup$ side $\times 1.5$ & 2 & 3.25 & 54.87 & 45.41 & 25.97 \\
Model one size up$^{a}$ & 1 & 1.27/1 & 54.53 & 45.74 & 23.68 \\
Model two sizes up$^{b}$ & 1 & 1.89/1.21 & 56.37 & 46.88 & 25.11 \\
\midrule
Tile $B{=}1$ & 2 & 2 & 54.49 & 47.32 & 28.63 \\
\rowcolor{ntMethod}
\method{} $B{=}1$ & 2 & 2 & 55.73 & 51.13 & 29.46 \\
\rowcolor{ntMethod}
\method{} $B{=}2$ & 3 & 3 & 56.24 & \textbf{51.82} & \textbf{30.84} \\
\bottomrule
\end{tabularx}
\end{nicetable}
 
\end{table}

\smallskip \noindent \textbf{Equal-Cost Alternatives.}
Because single-pass upscaling or larger models compete with cropping~\cite{becker2026scahi}, Table~\ref{tab:equal-cost} compares selective re-observation with uniform input and backbone scaling. On COCO, where medium and large objects dominate, full-image upscaling or RF-DETR-L slightly edges out \method{} at $B{=}1$ by refining global context in one pass. On CrowdHuman and VisDrone-DET, however, \method{} at $B{=}1$ beats every enlarged input, multi-scale union, and larger model by at least 0.64 and 3.49 AP@100 using half the pixels of $2{\times}$ upscaling. Uniform scaling spends quadratic pixels on background, whereas \method{} focuses native passes on localized deficits $1-r(a_j)$ in crowded scenes.

\smallskip \noindent \textbf{Online and Batched Allocation.}\label{sec:online}
We evaluate the streaming controllers when sequential requests preclude an offline split. Planning in arrival blocks of $M{=}16$ images stays within 0.05 AP@100 of offline \sba{}, while a causal quantile controller using prefix budgets and running class masses shifts AP@100 by $-0.10$ on average across 78 settings and per-image normalization still beats Tile across budgets. Because \cgs{} places gains on a unified scale, images couple only via the global threshold and class masses $M_k$, which stabilize over short streams.

\subsection{Ablation Study}
\label{sec:analysis}

\begin{figure*}[t]
\centering
\includegraphics[width=\textwidth]{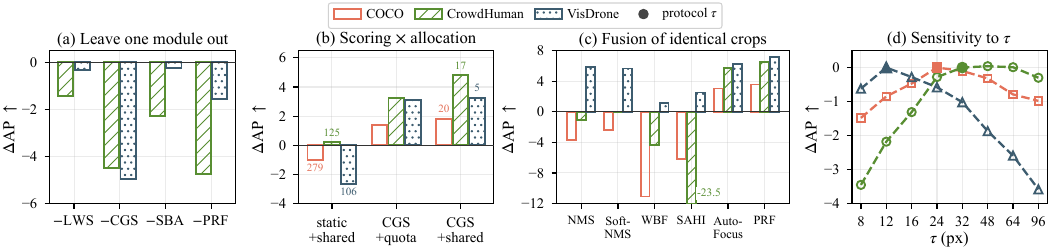}
\caption{\textbf{Ablations.} We analyze AP@100 of frozen RF-DETR-S across COCO, CrowdHuman, and VisDrone-DET. (a)~We remove one module at a time at $B{=}1$ corresponding to Table~\ref{tab:ablation}. (b)~We cross scoring and allocation rules at $B{=}2$ under \prf{} relative to static counts with per-image quotas, annotating maximum single-image calls. (c)~We compare fusion rules on identical $B{=}2$ crops relative to Coarse. (d)~We sweep target scale $\tau$ on training subsets relative to the filled protocol $\tau$.}
\label{fig:ablation}
\end{figure*}

\begin{table}[t]
\caption{\textbf{Module Combinations.} We evaluate frozen RF-DETR-S at $640{\times}640$ and $B=1$ on CrowdHuman and VisDrone-DET under an equal collection budget. Unchecked cells revert to $2{\times}2$ grid windows, static counts, round-robin allocation, and plain NMS, with bold marking the best score per metric.}
\label{tab:ablation}
\centering\scriptsize\setlength{\tabcolsep}{1pt}
\begin{nicetable}
\begin{tabularx}{\columnwidth}{cccc|YYYYY|YYYYY}
\toprule
\rowcolor{ntHeader}
\multicolumn{4}{c|}{Modules} & \multicolumn{5}{c|}{CrowdHuman} & \multicolumn{5}{c}{VisDrone-DET} \\
\tintrule\cmidrule(r){1-4}\cmidrule(lr){5-9}\cmidrule(l){10-14}
\rowcolor{ntHeader}
\rotatebox{90}{\lws{}} & \rotatebox{90}{\cgs{}} & \rotatebox{90}{\sba{}} & \rotatebox{90}{\prf{}} & AP & AP$_{50}$ & AP$_S$ & AP$_M$ & AP$_L$ & AP & AP$_{50}$ & AP$_S$ & AP$_M$ & AP$_L$ \\
\midrule
 &  &  &  & 40.95 & 73.92 & 34.91 & 42.47 & 42.04 & 27.46 & 45.79 & 18.99 & 37.64 & 59.20 \\
\midrule
$\checkmark$ &  &  &  & 43.20 & 76.68 & 32.62 & 42.97 & 45.96 & 25.73 & 43.87 & 16.71 & 37.57 & 59.14 \\
$\checkmark$ & $\checkmark$ &  &  & 43.26 & 75.80 & 37.86 & 44.31 & 44.26 & 27.76 & 46.11 & 20.03 & 37.05 & 60.07 \\
 & $\checkmark$ & $\checkmark$ &  & 43.61 & 75.55 & 36.41 & 42.19 & 46.26 & 27.78 & 46.13 & 19.46 & 37.76 & 60.32 \\
$\checkmark$ & $\checkmark$ & $\checkmark$ &  & 46.38 & 78.47 & 40.53 & 45.83 & 48.04 & 27.89 & 46.37 & 20.34 & 37.07 & 60.04 \\
$\checkmark$ & $\checkmark$ &  & $\checkmark$ & 48.85 & 82.05 & 41.08 & 49.32 & 50.84 & 29.22 & 48.57 & 21.05 & 39.20 & 60.39 \\
$\checkmark$ &  & $\checkmark$ & $\checkmark$ & 46.65 & 80.89 & 32.21 & 44.80 & 51.03 & 24.49 & 42.19 & 14.81 & 36.93 & \textbf{61.18} \\
 & $\checkmark$ & $\checkmark$ & $\checkmark$ & 49.70 & 82.96 & 41.48 & 50.02 & 51.71 & 29.13 & 48.58 & 20.45 & \textbf{39.76} & 61.05 \\
\midrule
\rowcolor{ntMethod}
\cellcolor{ntMethod}$\checkmark$ & \cellcolor{ntMethod}$\checkmark$ & \cellcolor{ntMethod}$\checkmark$ & \cellcolor{ntMethod}$\checkmark$ & \textbf{51.13} & \textbf{83.86} & \textbf{46.16} & \textbf{52.24} & \textbf{52.14} & \textbf{29.46} & \textbf{48.91} & \textbf{21.40} & 39.31 & 61.02 \\
\bottomrule
\end{tabularx}
\end{nicetable}

\end{table}

\noindent \textbf{Module Contributions.}
To isolate each module's contribution under a fixed budget, Table~\ref{tab:ablation} and Figure~\ref{fig:ablation}a toggle \lws{}, \cgs{}, \sba{}, and \prf{} at $B{=}1$. Most combinations improve over grid crops with NMS, while dropping \cgs{} or \prf{} causes the steepest losses exceeding 4.4 AP@100 on CrowdHuman, followed by \sba{} and \lws{}. Empirical closure or disabling the $4\tau$ guard alters $B{=}2$ AP@100 by at most 0.21, whereas fixed zoom $z{=}2$ loses 1.10 AP@100 on CrowdHuman. Multi-zoom search and cross-image sharing thus pay off fully only when \cgs{} calibrates deficits and \prf{} guards the bounded output set.

\smallskip \noindent \textbf{Shared Allocation and Complementary Gain.}
To separate deficit scoring from generic submodular counting, Figure~\ref{fig:ablation}b crosses scoring rules with quotas and shared allocation. Although static counts remain monotone submodular, sharing calls under static counts lowers accuracy on COCO and VisDrone-DET by piling up to 279 calls onto one image, whereas \cgs{} caps concentration at 20 calls with positive scoring--allocation interactions on all datasets. Setting $R_j{=}1$ degrades AP@100 sharply, and hard overlap rejection trails smooth residuals by up to 1.45 AP@100 because contracting $R_j(S)$ deters near-duplicates without blocking adjacent crowd~views.

\smallskip \noindent \textbf{Boundary-Aligned Fusion.}\label{sec:fusion}
To isolate post-inference admission on fixed views, Figure~\ref{fig:ablation}c compares fusion rules on identical $B{=}2$ crops. Plain NMS drops below Coarse on COCO and CrowdHuman because confident border fragments displace intact boxes under suppression and top-$K$ truncation. Full-box containment in $\mathcal P_c$ reverses this drop, letting \prf{} beat Soft-NMS~\cite{bodla2017softnms}, WBF~\cite{solovyev2021weighted}, SAHI~\cite{akyon2022sahi} merging, and AutoFocus~\cite{najibi2019autofocus} border pruning across all benchmarks, with dropping the $\mathcal P_c$ cut explaining nearly the entire loss toward NMS. Crossing Tile and \method{} with NMS and \prf{} further shows that \method{} leads under both fusions while \prf{} turns Tile's second-crop NMS regression into positive gains, and balanced squeezing adds 0.70 AP@100 on VisDrone-DET while preserving fused scores~$s_d$.

\smallskip \noindent \textbf{Target Scale.}
Because $\tau$ sets half-saturation in $r(a)$ for weight $u_j$ and closure $h(a,z)$, Figure~\ref{fig:ablation}d sweeps $\tau$ across $\{8,12,16,24,32,48,64,96\}$ on training subsets. Training and validation AP@100 shift by at most 0.62 and 0.86 across neighboring steps around the chosen $\tau$, revealing a stable plateau per domain, whereas extreme grid values vary accuracy by up to 3.57 points and push VisDrone-DET below Tile at $\tau{=}8$. Overly small $\tau$ treats medium-small targets as resolved while overly large $\tau$ spends calls on saturated objects, validating coarse training sweeps.

\subsection{Planner Analysis}
\label{sec:planner-analysis}

\begin{table}[t]
\caption{\textbf{Planner Diagnostics.} We audit surrogate fidelity over all 23 plans and 10 fixed-$B{=}1$ plans, demand skew, and tail latency on frozen RF-DETR-S at $B{=}1$ unless noted, reporting end-to-end TensorRT p95 latency.}
\label{tab:validity}
\centering\footnotesize\setlength{\tabcolsep}{2.0pt}
\begin{nicetable}
\begin{tabularx}{\columnwidth}{Xccc}
\toprule
\rowcolor{ntHeader}
Check & COCO & CrowdHuman & VisDrone-DET \\
\midrule
Spearman, all / fixed $B{=}1$ & 0.98 / 0.98 & 0.96 / 0.95 & 0.98 / 0.98 \\
AUROC, \cgs{} / static & 0.74 / 0.61 & 0.75 / 0.58 & 0.73 / 0.57 \\
\lws{} / exact optimum & 0.993 & 0.984 & 0.996 \\
\sba{} gain over quota & +0.77 & +2.57 & +0.26 \\
Call Gini / uncropped & 0.75 / 62\% & 0.74 / 60\% & 0.40 / 27\% \\
p95 (ms), $B{=}2$ / cap 4 & 24.9 / 14.6 & 26.4 / 17.0 & 17.5 / 17.5 \\
\bottomrule
\end{tabularx}
\end{nicetable}
 
\end{table}

\noindent \textbf{Surrogate Fidelity.}
Because higher budgets raise both coverage and accuracy, Table~\ref{tab:validity} audits rank alignment between $F(S)$ and AP@100 across budgets and strictly within equal-budget plans. Fixed-budget Spearman correlation stays at 0.95 to 0.99 at $B{=}1$ and $B{=}2$ (and at least 0.93 excluding Tile), coarse-only \cgs{} gain separates productive calls far better than static counts in AUROC while lifting productive rates monotonically across gain deciles, and \lws{} attains at least 98.4\% of the exact two-window optimum while a $28\times$ finer pool adds at most 0.56 AP@100. Interior deficit closure thus tracks downstream detection gains closely at fixed cost with negligible discretization~loss.

\smallskip \noindent \textbf{Uneven Demand and Coarse Support.}
To explain where cross-image sharing helps most, Table~\ref{tab:validity} links \sba{}'s gain over quotas to collection demand skew, proving largest on COCO and CrowdHuman where roughly 60\% of scenes need zero crops and smaller on uniformly busy VisDrone-DET. At $B{=}2$, \method{} retains 97.1\%, 96.9\%, and 89.1\% of coarse-matched objects while recovering 34.0\%, 41.0\%, and 28.9\% of coarse-missed targets across the three datasets, including 24.0\% to 27.0\% of targets without any matching coarse hypothesis, while sequential crop confirmation adds only 0.15 AP@100 on VisDrone-DET. Permissive one-shot cues and hypothesis-free $\mathcal P_c$ admission thus capture nearly all sequential gain.

\smallskip \noindent \textbf{Bounded Tail Latency.}
Because unconstrained sharing can stack calls onto extreme scenes and lengthen worst-case latency, Table~\ref{tab:validity} evaluates per-image heap caps during \sba{}. Capping each image at four crops at $B{=}2$ cuts end-to-end p95 latency by 36\% to 41\% on COCO and CrowdHuman while shifting AP@100 by at most 0.25 points. Multiplicative residual attenuation $R_j(S)$ concentrates each scene's highest-utility crops into its earliest picks, so redirecting extreme tail calls to the next global heap candidates bounds tail latency with negligible accuracy cost (Appendix~\ref{app:round3}).

\section{Conclusion}\label{sec:conclusion}
We presented \method{}, a training-free re-observation framework unifying cross-image crop allocation with boundary-aligned multi-view fusion on frozen detectors. To overcome isolated per-image cropping and boundary-truncated box competition, \method{} models coarse resolution deficits and interior positive-region residuals inside a monotone submodular surrogate, plans multi-zoom windows via 2D prefix-sum sweeps and a lazy max-heap, and merges views through matching interior admission and class-balanced top-$K$ selection. Across detection transformers such as RF-DETR and convolutional detectors such as YOLO26, \method{} consistently outperforms uniform slicing such as SAHI, content-adaptive cropping such as Dynamic Tiling, complete slicing systems, and single-pass scaling where small and crowded objects concentrate. Fixed-budget surrogate diagnostics, streaming controllers, and per-image heap caps further support offline and streaming~deployment.

\FloatBarrier
\label{review:body-end}
\clearpage

\section*{LLM Usage Statement}
\noindent We used OpenAI Codex and Anthropic Claude to assist with manuscript drafting, editing, and reviewer-style feedback, as well as literature and citation checks, code development, experiment design, execution and analysis, and figure preparation. Manuscript sources and author-generated feedback were processed by these tools under author oversight. No generative AI was used to generate or annotate the datasets. The authors carefully verified all the assisted outputs and take full responsibility for the submission.

{
    \footnotesize
    \setlength{\bibsep}{0pt}
    \linespread{0.955}\selectfont
    \bibliographystyle{ieeenat_fullname}
    \bibliography{references/references}
}

\clearpage
\label{review:supplement-start}
\twocolumn[{%
  \centering
  {\Large\bfseries GRAIN: Shared-Budget Complementary Re-observation for Object Detection\par}
  \vspace{0.5em}
  {\Large Supplementary Material\par}
  \vspace{1em}
}]

\captionsetup{hypcap=false,font=small,skip=4pt}
\captionsetup[table]{position=top,skip=4pt}
\captionsetup[figure]{position=bottom,skip=4pt}
\captionsetup[sub]{font=footnotesize,skip=2pt}
\flushbottom
\clubpenalty=10000
\widowpenalty=10000
\displaywidowpenalty=10000
\setlength{\abovedisplayskip}{5.5pt plus 2.5pt minus 1.5pt}
\setlength{\belowdisplayskip}{5.5pt plus 2.5pt minus 1.5pt}
\setlength{\abovedisplayshortskip}{3pt plus 2pt minus 1pt}
\setlength{\belowdisplayshortskip}{4pt plus 2pt minus 1pt}
\renewcommand{\topfraction}{0.92}
\renewcommand{\bottomfraction}{0.88}
\renewcommand{\textfraction}{0.06}
\renewcommand{\floatpagefraction}{0.82}
\renewcommand{\dbltopfraction}{0.92}
\renewcommand{\dblfloatpagefraction}{0.82}
\setcounter{topnumber}{4}
\setcounter{bottomnumber}{3}
\setcounter{totalnumber}{6}
\setcounter{dbltopnumber}{3}
\setlength{\textfloatsep}{8pt plus 3pt minus 2pt}
\setlength{\floatsep}{7pt plus 2.5pt minus 2pt}
\setlength{\dbltextfloatsep}{8pt plus 3pt minus 2pt}
\setlength{\dblfloatsep}{7pt plus 2.5pt minus 2pt}

\newtheorem{remark}{Remark}[section]

\definecolor{ntYes}{HTML}{3B7A28}
\definecolor{ntPart}{HTML}{B26A15}
\newcommand{\cmark}{\textcolor{ntYes}{$\checkmark$}}
\newcommand{\pmark}{\textcolor{ntPart}{$\circ$}}
\newcommand{\xmark}{\textcolor{black!38}{$\times$}}

\newcounter{grainalgorithm}
\newcommand{\algind}{\hspace*{1.1em}}
\newenvironment{grainalg}[2]{%
  \par\noindent\begin{minipage}{\columnwidth}%
  \refstepcounter{grainalgorithm}%
  \label{#2}%
  \scriptsize
  \hrule height \heavyrulewidth\vspace{2.5pt}%
  \noindent\textbf{Algorithm~\thegrainalgorithm} \textbf{#1}\par\vspace{2pt}%
  \hrule height \lightrulewidth\vspace{2.5pt}%
  \begin{enumerate}[leftmargin=1.65em,itemsep=1.55pt,topsep=0.8pt,parsep=0pt,label={\scriptsize\arabic*:}]%
}{%
  \end{enumerate}%
  \vspace{2pt}\hrule height \heavyrulewidth%
  \end{minipage}\par\vspace{4pt plus 2pt minus 1pt}%
}

\appendix
\numberwithin{equation}{section}
\makeatletter
\newcommand{\graincontentssection}[2]{%
  \par\addvspace{2.4pt}%
  \@dottedtocline{1}{0em}{1.5em}{\numberline{\textbf{\ref*{#1}}}\hyperref[#1]{\textbf{#2}}}{\textbf{\pageref{#1}}}}
\newcommand{\graincontentssubsection}[2]{%
  \@dottedtocline{2}{1.4em}{2.3em}{\numberline{\ref*{#1}}\hyperref[#1]{#2}}{\pageref{#1}}}
\makeatother

\noindent\textbf{\large Contents}\par\vspace{2pt}
{\footnotesize\noindent\textbf{Roadmap.} Appendices~\ref{app:theory}--\ref{app:coverage-to-ap} present the resolution-deficit formulation, submodular guarantees (including the proof of Proposition~\ref{prop:greedy} in Appendix~\ref{sec:proof}), discrete sweeps, and four-channel AP coupling behind Section~\ref{sec:method} and Table~\ref{tab:validity}. Appendix~\ref{app:protocol} extends Tables~\ref{tab:family}--\ref{tab:equal-cost} and Figure~\ref{fig:fractional} with evaluation protocols, official capacities, scale breakdowns, deployment profiling, and online streaming. Appendices~\ref{app:implementation}--\ref{app:symbol-table} consolidate the unified algorithm, canonical constants, and core notation. Appendix~\ref{app:round3} expands Figure~\ref{fig:ablation}, Table~\ref{tab:ablation}, and Table~\ref{tab:validity} with module synergies, budget scaling, target-scale calibration, and per-image latency caps. Appendices~\ref{app:full-related-work}--\ref{app:societal-impacts} conclude with expanded related work and method taxonomy, qualitative visual controls, and societal impacts.\par}\vspace{2pt}
\begingroup
\small
\setlength{\parskip}{0.8pt}
\graincontentssection{app:theory}{Resolution Deficit, Complementary Coverage, and Guarantees}
\graincontentssubsection{app:symbols}{Definitions and Modeling Assumptions}
\graincontentssubsection{app:closure-derivation}{Closure and Marginal-Gain Derivations}
\graincontentssubsection{sec:proof}{Fixed-Pool Greedy Guarantee}
\graincontentssubsection{sec:implementation-gap}{Additive Search Error and Per-Image Caps}
\graincontentssection{app:discrete-inference}{Continuous Windows to Discrete Inference}
\graincontentssubsection{app:coordinates}{Coordinate Mapping and Boundary Rules}
\graincontentssubsection{app:algorithmic-identities}{Dual-Origin 2D Difference-Array Sweep}
\graincontentssubsection{app:offline-threshold}{Global Threshold Equivalence of SBA}
\graincontentssubsection{app:fusion-order}{Within-Class Order Invariance of PRF Squeeze}
\graincontentssection{app:coverage-to-ap}{From Coverage to Detection AP}
\graincontentssubsection{sec:rank-alignment-theory}{Four-Channel AP Decomposition and Conditional Bounds}
\graincontentssubsection{app:theory-link-audits}{Empirical Link Calibration, Plan Ranking, and AP Attribution}
\graincontentssection{app:protocol}{More Quantitative Results}
\graincontentssubsection{app:checkpoint-protocol}{Datasets, Frozen Checkpoints, and Evaluation Protocol}
\graincontentssubsection{app:revision-evidence}{Cross-Detector Scaling and Official-Limit Evaluation}
\graincontentssubsection{app:latency}{Native Pipelines and Deployment Efficiency}
\graincontentssubsection{app:online-effective}{Causal Online Streaming Without Test-Split Statistics}
\graincontentssection{app:implementation}{GRAIN Algorithm and Protocol Constants}
\graincontentssection{app:symbol-table}{Symbols Table}
\graincontentssection{app:round3}{Additional Ablation Studies}
\graincontentssubsection{app:factorial-controls}{Module Interactions and Fixed-View Fusion Controls}
\graincontentssubsection{app:review-allocation}{Budget Scaling and Complete-Grid Analysis}
\graincontentssubsection{app:tau}{Target Scale, Label-Free Calibration, and Hyperparameter Sensitivity}
\graincontentssubsection{app:call-caps}{Per-Image Call Caps and Streaming Block Horizon}
\graincontentssection{app:full-related-work}{Full Related Work}
\graincontentssection{app:current-qualitative}{More Qualitative Results}
\graincontentssection{app:societal-impacts}{Societal Impacts}
\par
\endgroup
\vspace{4pt plus 2pt minus 1pt}

\section{Resolution Deficit, Complementary Coverage, and Guarantees}
\label{app:theory}
\label{app:coverage-guarantees}

Before invoking the detector on any crop, the planning stage must predict how much detection utility a candidate window will contribute across images of varying dimensions and object densities using only coarse-pass outputs. Relying on raw hypothesis counts or uncalibrated confidence sums fails across images because large objects are already resolved by the coarse pass, low-confidence background clutter inflates false demand, and overlapping crops double-count the same targets. In this section, we study how to map coarse bounding boxes and scores into a cross-image comparable resolution-deficit weight, derive the parameter-free closure kernel and multiplicative residual updates that make collection-wide planning submodular, prove the $(1-1/e)$ greedy cardinality guarantee on the coverage surrogate $F(S)$, and bound step-level discretization errors and per-image latency caps $H_{\max}$.

\subsection{Definitions and Modeling Assumptions}
\label{app:symbols}

Throughout our analysis, let image $i\in\{1,\ldots,N\}$ have native spatial resolution $(W_i,H_i)$, and let the frozen detector operate on inputs of size $(W_D,H_D)$. Coarse predictions and candidate crop windows are expressed in the native image coordinate frame $[0,W_i]\times[0,H_i]$, with canvas-clipped border evaluation evaluated in Appendix~\ref{app:revision-evidence}.

\smallskip\noindent\textbf{(D1) Coarse Hypotheses and Effective Input Scale.}
Because benchmark images vary in native dimensions $(W_i,H_i)$ while the frozen backbone processes a fixed input canvas $(W_D,H_D)$, raw pixel area in source coordinates does not reflect the spatial support an object receives inside the detector. We therefore study the effective input scale of each coarse hypothesis $j$ on image $i(j)$ with bounding box $b_j=[x_{j,1},y_{j,1},x_{j,2},y_{j,2}]$, box area $A_j=|b_j|=(x_{j,2}-x_{j,1})(y_{j,2}-y_{j,1})$, and confidence score $s_j\ge s_{\min}$, with all frozen protocol constants listed once in Table~\ref{tab:constants}, defined by area preservation as
\begin{equation}
a_j=\sqrt{\frac{A_j\,W_D H_D}{W_{i(j)}H_{i(j)}}}.
\label{eq:app-input-scale}
\end{equation}
Under direct stretch resizing to $(W_D,H_D)$, Eq.~\eqref{eq:app-input-scale} gives the exact geometric-mean box side length seen by the detector backbone, whereas under isotropic letterbox resizing with stride padding (as in YOLO26) it serves as a consistent area-preserving scale proxy during planning before post-inference fusion inverts each detector's exact preprocessing in Appendix~\ref{app:coordinates}. Expressing scale through $a_j$ places hypotheses from arbitrary image resolutions onto one common detector-input ruler.

\smallskip\noindent\textbf{(D2) Confidence Salience and Resolution Deficit.}
A coarse hypothesis warrants re-observation only if it is both plausible enough to represent a real object or crowd cue and small enough to suffer from limited pixel support. To formalize this trade-off, we pair the Hill pixel-support saturation function $r(a)=a^2/(a^2+\tau^2)\in[0,1)$ at domain half-saturation scale $\tau>0$ with the confidence salience curve $\phi(s)=\sqrt{s}(1-s/2)$, assigning each coarse hypothesis the initial resolution-deficit weight
\begin{equation}
u_j=\phi(s_j)\,[1-r(a_j)]=\sqrt{s_j}\Bigl(1-\frac{s_j}{2}\Bigr)\frac{\tau^2}{a_j^2+\tau^2}\ge0.
\label{eq:app-weight-expanded}
\end{equation}
Differentiating $\phi(s)$ yields $\phi'(s)=(1-3s/2)/(2\sqrt{s})$, which vanishes and changes sign at $s=2/3$. This structure directly supports our planning objective in three regimes: $\phi(s)$ vanishes at $s=0$ to suppress low-score background clutter, peaks at $s=2/3$ to prioritize ambiguous candidates near the detection boundary, and retains half of its unit-scale weight at $s=1$ ($\phi(1)=1/2$). Retaining non-zero salience on confident small detections ($a_j\lesssim\tau$) ensures that resolved cluster anchors still trigger local zoom views that sharpen their own bounding boxes and uncover adjacent missed instances, while large objects ($a_j\gg\tau$) are automatically discounted by the deficit factor $1-r(a_j)=\tau^2/(a_j^2+\tau^2)$.

\smallskip\noindent\textbf{(D3) Interior Positive Region $\mathcal P_c$.}
When an internal crop border cuts across an object, the detector observes only a truncated fragment whose inferred box often carries high confidence and corrupts multi-view suppression. To prevent the planner from crediting windows that bisect targets, we restrict valid observation support for any candidate crop window $c=[x_{c,1},y_{c,1},x_{c,2},y_{c,2}]\in\mathcal C_{i(c)}$ (with $q_{cj}=0$ whenever $i(c)\ne i(j)$) to an interior positive region $\mathcal P_c=(\underline{x}_c,\overline{x}_c)\times(\underline{y}_c,\overline{y}_c)$ that contracts internal crop edges inward by $\delta_{\mathrm{img}}$ native pixels while exempting edges within $\delta_{\mathrm{img}}$ of the outer source canvas:
\begin{equation}
\begin{aligned}[b]
\underline{x}_c&=\begin{cases}x_{c,1}+\delta_{\mathrm{img}},&x_{c,1}>\delta_{\mathrm{img}},\\ -\infty,&x_{c,1}\le\delta_{\mathrm{img}},\end{cases}\\[2pt]
\overline{x}_c&=\begin{cases}x_{c,2}-\delta_{\mathrm{img}},&x_{c,2}<W_{i(c)}-\delta_{\mathrm{img}},\\ +\infty,&x_{c,2}\ge W_{i(c)}-\delta_{\mathrm{img}},\end{cases}
\end{aligned}
\label{eq:app-positive-region}
\end{equation}
with $\underline{y}_c$ and $\overline{y}_c$ defined identically along $[y_{c,1},y_{c,2}]$ against $H_{i(c)}$. Binary interior containment $\pi_{cj}=\mathbf{1}[b_j\subseteq\mathcal P_c]$ holds if and only if $\underline{x}_c<x_{j,1}$, $x_{j,2}<\overline{x}_c$, $\underline{y}_c<y_{j,1}$, and $y_{j,2}<\overline{y}_c$. Introducing $\mathcal P_c$ in Eq.~\eqref{eq:app-positive-region} establishes the shared geometric rule between planning and fusion: a window accrues planning credit for hypothesis $j$ if and only if an intact detection at that location is eligible to pass interior admission in Positive-Region Fusion (\prf{}).

\smallskip\noindent\textbf{(A1) Scale-Transfer Monotonicity.}
Because coarse and crop passes share frozen detector weights, magnifying window $c$ to $(W_D,H_D)$ scales any contained object's effective input side length from $a_j$ to $z_ca_j$, monotonically increasing its pixel support toward $r(z_ca_j)$ up to the domain saturation ceiling $4\tau$.

\smallskip\noindent\textbf{(A2) Multiplicative Residual Independence.}
Under repeated observation by a subset of windows $S\subseteq\mathcal C=\bigcup_{i=1}^N\mathcal C_i$, the remaining unresolved fraction of hypothesis $j$'s initial deficit contracts multiplicatively across covering views prior to observing downstream crop detections.

\subsection{Closure and Marginal-Gain Derivations}
\label{app:closure-derivation}

We next study how to quantify the fraction of an object's resolution deficit closed by a candidate window at zoom $z>1$ without introducing extra shape hyperparameters. Under Assumption~(A1), magnifying an object of effective input scale $a>0$ by zoom factor $z>1$ increases its Hill pixel support by
\begin{equation}
\begin{aligned}[b]
r(za)-r(a)&=\frac{z^2a^2}{z^2a^2+\tau^2}-\frac{a^2}{a^2+\tau^2}\\
&=\frac{(z^2-1)a^2\tau^2}{(z^2a^2+\tau^2)(a^2+\tau^2)}.
\end{aligned}
\label{eq:app-saturation-diff}
\end{equation}
Because the object's initial coarse deficit is $1-r(a)=\tau^2/(a^2+\tau^2)$, dividing the support gain in Eq.~\eqref{eq:app-saturation-diff} by $1-r(a)$ cancels the factor $\tau^2/(a^2+\tau^2)$ identically and yields the normalized fractional closure kernel
\begin{equation}
h(a,z)=\frac{r(za)-r(a)}{1-r(a)}=\frac{(z^2-1)a^2}{z^2a^2+\tau^2},\quad za<4\tau,
\label{eq:app-closure-frac}
\end{equation}
paired with a hard saturation cutoff $h(a,z)=0$ once the magnified scale reaches $za\ge4\tau$. Combining spatial admission from Eq.~\eqref{eq:app-positive-region} with Eq.~\eqref{eq:app-closure-frac} defines the interior observation quality $q_{cj}$ and single-view weighted closure:
\begin{equation}
\begin{aligned}[b]
q_{cj}&=\pi_{cj}\,h(a_j,z_c),\\
u_j q_{cj}&=\pi_{cj}\,\phi(s_j)\,[r(z_ca_j)-r(a_j)],\; z_ca_j<4\tau.
\end{aligned}
\label{eq:app-closure}
\end{equation}
This derivation shows why $h(a,z)$ requires no auxiliary tuning constants: for any interior hypothesis below the $4\tau$ ceiling, $q_{cj}\in(0,1)$ is strictly increasing in zoom $z_c$, and the product $u_j q_{cj}$ telescopes directly into the salience-weighted increase in pixel support $\phi(s_j)[r(z_ca_j)-r(a_j)]$.

To plan multiple windows per image without double-counting already covered targets or imposing rigid spatial overlap bans, we invoke Assumption~(A2) to track the remaining unresolved demand $R_j(S)$ and total collection coverage $F(S)$ over any candidate subset $S\subseteq\mathcal C$:
\begin{equation}
R_j(S)=\prod_{c\in S}(1-q_{cj}),\qquad F(S)=\sum_j u_j\,[1-R_j(S)],
\label{eq:app-remaining}
\end{equation}
where $R_j(\varnothing)=1$, $F(\varnothing)=0$, and $F(S)=\sum_{i=1}^N F_i(S_i)$ separates additively across images with $S_i=S\cap\mathcal C_i$. Evaluating the marginal increment of adding a new window $c\notin S$ updates residuals solely on host image $i(c)$ via $R_j(S\cup\{c\})=R_j(S)(1-q_{cj})$ and yields the Complementary Gain Scoring (\cgs{}) identity:
\begin{equation}
\begin{aligned}[b]
\Delta(c\mid S)&=F(S\cup\{c\})-F(S)\\
&=\sum_{j:i(j)=i(c)} u_j\,q_{cj}\,R_j(S_{i(c)}).
\end{aligned}
\label{eq:app-marginal-gain}
\end{equation}
Equation~\eqref{eq:app-marginal-gain} provides two foundational properties for \method{}. First, multiplicative residual contraction smoothly discounts repeated views of the same target, steering subsequent picks toward uncovered neighbors without blocking complementary multi-scale views. Second, because $\Delta(c\mid S)$ is measured in unified deficit-closure units across all images and depends only on local history $S_{i(c)}$, Shared-Budget Allocation (\sba{}) ranks candidate windows globally across a collection and spends fractional budgets strictly where unresolved demand is highest.

\begin{remark}[Coupled $4\tau$ Planner and Fusion Guards]
In Eq.~\eqref{eq:app-closure}, the planner enforces $z_ca_j<4\tau$ on projected coarse-hypothesis scale (and $\widehat z_ca_j<4\tau$ after integer pixel rounding), directly mirroring the post-inference \prf{} scale gate $a_{\rm g}(b)<4\tau$ evaluated on the full-image input scale of returned crop detections. Sensitivity controls in Appendix~\ref{app:factorial-controls} and Appendix~\ref{app:tau} confirm that this coupled $4\tau$ boundary prevents high-zoom calls from re-observing already resolved objects while avoiding the $0.87$ to $2.27$ AP@100 recall loss incurred when tightening the shared gate to $2\tau$.
\end{remark}

\subsection{Fixed-Pool Greedy Guarantee}
\label{sec:proof}

To establish the optimization properties of cross-image budget sharing over any fixed finite candidate window pool $\mathcal C=\bigcup_{i=1}^N\mathcal C_i$, Main Proposition~\ref{prop:greedy} states that with nonnegative deficit weights $u_j\ge0$ and bounded closure qualities $q_{cj}\in[0,1]$, exact greedy selection under a collection call budget $|S|\le L$ achieves at least $(1-1/e)$ of the optimal collection coverage $F(S^\star)$.

\begin{proof}[Proof of Main Proposition~\ref{prop:greedy}]
We verify the three structural conditions required for cardinality-constrained monotone submodular maximization.
\begin{enumerate}[leftmargin=*,itemsep=2pt]
\item \textbf{Monotonicity.} Because $u_j\ge0$ and $q_{cj}\in[0,1]$ for every window-hypothesis pair $(c,j)$, each retention factor $1-q_{cj}$ lies in $[0,1]$, ensuring $R_j(S)=\prod_{d\in S}(1-q_{dj})\in[0,1]$ for any subset $S\subseteq\mathcal C$. By Eq.~\eqref{eq:app-marginal-gain}, adding any window $c\notin S$ yields a nonnegative marginal gain $\Delta(c\mid S)=\sum_j u_j q_{cj} R_j(S)\ge0$, so $F(S)\le F(T)$ whenever $S\subseteq T$.
\item \textbf{Submodularity.} Let $S\subseteq T\subseteq\mathcal C$ and consider any candidate window $c\notin T$. Because every window $c'\in T\setminus S$ multiplies hypothesis $j$'s residual by $1-q_{c'j}\in[0,1]$, the remaining demand satisfies $R_j(T)=R_j(S)\prod_{c'\in T\setminus S}(1-q_{c'j})\le R_j(S)$ for every $j$. Multiplying by $u_j q_{cj}\ge0$ and summing over $j$ preserves the inequality:
\[
\begin{aligned}
\Delta(c\mid T)&=\sum_j u_j q_{cj} R_j(T)\\
&\le\sum_j u_j q_{cj} R_j(S)=\Delta(c\mid S).
\end{aligned}
\]
\item \textbf{Cardinality Guarantee.} Since $F(\varnothing)=0$ is normalized, monotone, and submodular, exact greedy selection under $|S|\le L$ satisfies $F(S_g)\ge[1-(1-1/L)^L]F(S^\star)\ge(1-1/e)F(S^\star)$ relative to any optimal subset $|S^\star|\le L$ by the classical theorem of Nemhauser et al.~\cite{nemhauser1978analysis}, which also follows as the zero-error ($\epsilon_t=0$) case of Proposition~\ref{prop:approx-greedy} below.
\end{enumerate}
\end{proof}
This result confirms that diminishing returns arise intrinsically from multiplicative deficit contraction $R_j(S)$, providing a worst-case optimality floor for collection-wide greedy scheduling over any fixed window pool.

\subsection{Additive Search Error and Per-Image Caps}
\label{sec:implementation-gap}

In practical deployment, \method{} departs from idealized unconstrained greedy search in two ways: \lws{} selects candidate windows at nominal zoom $z$ before recomputing gain at integer-rounded zoom $\widehat z_c$, and real-time deployments optionally enforce a per-image call cap $|S_i|\le H_{\max}$ to bound worst-case frame latency. Propositions~\ref{prop:approx-greedy} and~\ref{prop:capped-greedy} bound both departures on the fixed pool.

\begin{proposition}[Additive Search Error Under Inexact Steps]
\label{prop:approx-greedy}
Fix candidate pool $\mathcal C$, weights $u_j\ge0$, qualities $q_{cj}\in[0,1]$, and integer collection budget $L\ge1$. Starting from $S_0=\varnothing$, let step $t\in\{1,\ldots,L\}$ append one unused window (or take a zero-gain dummy step upon termination) with realized gain $g_t=F(S_t)-F(S_{t-1})$ and step shortfall $\epsilon_t=\max(\{0\}\cup\{\Delta(c\mid S_{t-1}):c\in\mathcal C\setminus S_{t-1}\})-g_t$. Relative to any fixed-pool optimal subset $S^\star\subseteq\mathcal C$ satisfying $|S^\star|\le L$,
\begin{equation}
\begin{aligned}[b]
F(S_L)\ge{}&\left[1-\left(1-\frac1L\right)^L\right]F(S^\star)\\
&-\sum_{t=1}^{L}\left(1-\frac1L\right)^{L-t}\epsilon_t.
\end{aligned}
\label{eq:additive-search-error}
\end{equation}
\end{proposition}
\begin{proof}
1.~\emph{One-step recursion.} Monotonicity and submodularity of $F$ together with $|S^\star\setminus S_{t-1}|\le L$ imply
\[
F(S^\star)-F(S_{t-1})\le\sum_{c\in S^\star\setminus S_{t-1}}\Delta(c\mid S_{t-1})\le L(g_t+\epsilon_t).
\]
2.~\emph{Geometric unrolling.} Defining the optimality gap $\delta_t=F(S^\star)-F(S_t)$ and substituting $g_t=\delta_{t-1}-\delta_t$ yields the linear recursion $\delta_t\le(1-1/L)\delta_{t-1}+\epsilon_t$ for each step $t=1,\ldots,L$. Unrolling inductively from $\delta_0=F(S^\star)$ gives Eq.~\eqref{eq:additive-search-error}.
\end{proof}

\begin{proposition}[Separable Per-Image Cap Guarantee]
\label{prop:capped-greedy}
Under the per-image capped feasible family $\mathcal F_{\mathrm{cap}}=\{S\subseteq\mathcal C:|S|\le L,\ |S\cap\mathcal C_i|\le H_{\max}\ \forall i\}$, exact separable greedy selection returns $S_g^{\mathrm{cap}}$ satisfying
\[
F(S_g^{\mathrm{cap}})\ge\Bigl(1-\frac1e\Bigr)\,F(S_{\mathrm{cap}}^\star),
\]
relative to any capped optimum $S_{\mathrm{cap}}^\star\in\mathcal F_{\mathrm{cap}}$.
\end{proposition}
\begin{proof}
Because $F(S)=\sum_{i=1}^N F_i(S_i)$ separates additively by image, let $G_i^{(r)}$ denote image $i$'s exact local greedy prefix of length $r\le m_i=\min(|\mathcal C_i|,H_{\max})$ under consistent global tie breaking. Local marginal gains $g_{i,1}\ge\cdots\ge g_{i,m_i}\ge0$ are non-increasing, so the global max-heap merges these $N$ sequences and accumulates the largest $\min(L,\sum_i m_i)$ gains with multiplicity. Any feasible capped solution $S_{\mathrm{cap}}^\star\in\mathcal F_{\mathrm{cap}}$ allocates $k_i^\star=|S_i^\star|\le m_i$ windows to image $i$ with $\sum_i k_i^\star\le L$. Therefore, the union of local prefix gains $\{g_{i,r}:1\le r\le k_i^\star\}$ across all images contains at most $L$ terms and is upper-bounded by the global greedy sum:
\[
\begin{aligned}
F(S_g^{\mathrm{cap}})&\ge\sum_{i=1}^N\sum_{r=1}^{k_i^\star}g_{i,r}=\sum_{i=1}^N F_i\bigl(G_i^{(k_i^\star)}\bigr)\\
&\ge\Bigl(1-\frac1e\Bigr)\sum_{i=1}^N F_i(S_i^\star)=\Bigl(1-\frac1e\Bigr)F(S_{\mathrm{cap}}^\star),
\end{aligned}
\]
where the second inequality applies the exact single-image cardinality bound, corresponding to $\epsilon_t=0$ in Proposition~\ref{prop:approx-greedy}, to each image with $k_i^\star>0$, while terms with $k_i^\star=0$ vanish identically.
\end{proof}
Together, Propositions~\ref{prop:approx-greedy} and~\ref{prop:capped-greedy} show that single-step rounding discrepancies enter the global coverage bound only as a discounted additive sum $E_{\mathrm{LWS}}=\sum_{t=1}^L(1-1/L)^{L-t}\epsilon_t$ with early errors exponentially attenuated, while per-image latency caps preserve the full $(1-1/e)$ guarantee over the capped feasible set without requiring general matroid intersection solvers.

\smallskip\noindent\textbf{Empirical Discrete-Search Fidelity.}
To verify that nominal-to-rounded discretization error $E_{\mathrm{LWS}}$ is negligible on real detector outputs and that the default lattice captures nearly all coverage of a dense search space, we compare three discrete solver configurations: deployed \lws{} on the default lattice, exact same-pool greedy selection without nominal-zoom rounding across all $255$ dataset-detector-budget cells ($17$ frozen detectors at $B\in\{0.25,0.5,1,2,4\}$), and a $28{\times}$ denser reference lattice across Small checkpoints. The results are reported in Table~\ref{tab:fidelity-search-audit}, and Table~\ref{tab:theory-deployed-contract} summarizes the formal scope of idealized versus deployed \method{} solvers.

\begin{table}[!t]
\centering\small\setlength{\tabcolsep}{4pt}
\caption{\textbf{Discrete-Search Fidelity Across $255$ Cells.} Worst-case coverage ratio $F/F_{\mathrm{gr}}$ vs.\ exact greedy, maximum normalized search error $E_{\mathrm{LWS}}/\!\sum_j u_j$ in Proposition~\ref{prop:approx-greedy}, and $28{\times}$ finer-lattice lift $\Delta F$ across Small checkpoints.}
\label{tab:fidelity-search-audit}
\begin{nicetable}
\setlength{\tabcolsep}{3.0pt}
\begin{tabularx}{\columnwidth}{Xccc}
\toprule
\rowcolor{ntHeader}
 & Min.\ $F/F_{\mathrm{gr}}$ & Max.\ $E_{\mathrm{LWS}}/\!\sum u_j$ & $28{\times}$ Grid \\
\rowcolor{ntHeader}
\multirow{-2}{*}{Benchmark} & (\%)$\uparrow$ & (\%)$\downarrow$ & $\Delta F$ (\%)$\downarrow$ \\
\midrule
COCO & 99.69 & 3.80 & +0.91--2.95 \\
CrowdHuman & 99.87 & 0.29 & +0.49--2.47 \\
VisDrone-DET & 99.97 & 0.04 & +0.86--2.30 \\
\bottomrule
\end{tabularx}
\end{nicetable}

\end{table}

\begin{table*}[!t]
\centering\footnotesize\setlength{\tabcolsep}{4.5pt}
\caption{\textbf{Guarantee Scope Across Idealized and Deployed \method{} Solvers.} Separation between exact fixed-pool greedy bounds in Propositions~\ref{prop:greedy}--\ref{prop:capped-greedy} and deployed discrete, capped, micro-batch, or causal streaming solvers.}
\label{tab:theory-deployed-contract}
\begin{nicetable}
\begin{tabularx}{\textwidth}{llcc>{\raggedright\arraybackslash}X}
\toprule
\rowcolor{ntHeader}
Solver Variant & Pool \& Local Search & Budget Rule & Formal $F(S)$ Bound & Empirical Guarantee and Validation \\
\midrule
\multicolumn{5}{l}{\emph{Idealized Fixed-Pool Solvers}} \\
Exact Uncapped Greedy & Fixed $\mathcal C$, exact gain & $|S|\le L$ & $\ge(1{-}1/e)\,F(S^\star)$ & Monotone submodular floor in Proposition~\ref{prop:greedy} \\
Exact Capped Greedy & Fixed $\mathcal C$, exact gain & $|S|\le L,\;|S_i|\le H_{\max}$ & $\ge(1{-}1/e)\,F(S_{\mathrm{cap}}^\star)$ & Separable partition floor in Proposition~\ref{prop:capped-greedy} \\
\midrule
\multicolumn{5}{l}{\emph{Deployed Discrete and Streaming Solvers}} \\
\rowcolor{ntMethod}
\textbf{Offline \method{} (\lws{}+\sba{})} & Nominal sweep, rounded $\widehat z_c$ & $|S|\le L$ & $\ge(1{-}1/e)\,F(S^\star)-E_{\mathrm{LWS}}$ & Proposition~\ref{prop:approx-greedy}, with $F/F_{\mathrm{gr}}\ge 99.69\%$ in Table~\ref{tab:fidelity-search-audit} \\
Capped \method{} & Nominal sweep, rounded $\widehat z_c$ & $|S|\le L,\;|S_i|\le H_{\max}$ & -- & $-36\%$ to $-41\%$ p95 and $|\Delta\mathrm{AP}|\le 0.25$ at cap $4$ in Figure~\ref{fig:call-caps} \\
Block \sba{} ($M$ frames) & Per-block heap, no carryover & $|S_b|\le\lfloor BM\rfloor$ & Exact offline at $M{=}N$ & $M{\ge}16$ within $0.05$ AP of $M{=}N$ in Figure~\ref{fig:sba-blocks} \\
\rowcolor{ntMethod}
\textbf{Causal Online \method{}} & Causal quantile $\lambda_t$, $M_k^{\mathrm{train}}$ & Prefix $\le\lfloor Bt\rfloor$ & -- & $99.7\%$ budget use and $101/102$ wins vs.\ Tile in Figure~\ref{fig:online-deltas} \\
\bottomrule
\end{tabularx}
\end{nicetable}

\end{table*}

Across all $255$ cells in Table~\ref{tab:fidelity-search-audit}, deployed \lws{} retains at least $99.69\%$, $99.87\%$, and $99.97\%$ of exact same-pool greedy coverage $F_{\mathrm{gr}}$ on COCO, CrowdHuman, and VisDrone-DET, while bounding maximum normalized search error $E_{\mathrm{LWS}}/\sum_j u_j$ below $3.80\%$, $0.29\%$, and $0.04\%$. Expanding the candidate grid by $28{\times}$ increases surrogate coverage by only $0.49\%$ to $2.95\%$ across benchmarks. Because integer pixel rounding shifts window borders by at most half a pixel per side, nominal-zoom prefix sums closely track exact rounded-window gains, while Table~\ref{tab:theory-deployed-contract} keeps exact fixed-pool guarantees and deployed streaming controllers explicitly aligned with their respective operating regimes.

\section{Continuous Windows to Discrete Inference}
\label{app:discrete-inference}

While Appendix~\ref{app:theory} formulates coverage over continuous window coordinates, deploying \method{} inside a real-time detection pipeline requires resolving four algorithmic problems: preserving exact boundary alignment under integer pixel rounding and detector-specific preprocessing, searching dense multi-zoom candidate lattices without quadratic CPU overhead, connecting offline global heap allocation to causal streaming thresholds, and truncating fused multi-view predictions to $K$ slots without distorting within-class confidence rankings. We analyze these four correspondences below.

\subsection{Coordinate Mapping and Boundary Rules}
\label{app:coordinates}

To guarantee that the spatial region credited during planning coincides with the tensor slice evaluated by the frozen detector, we first formalize the discrete coordinate mapping between native images and crop views. Let an integer-rounded crop window $c=[x_0,y_0,x_1,y_1]\subseteq[0,W_i]\times[0,H_i]$ have pixel area $|c|=(x_1-x_0)(y_1-y_0)$. When $c$ is passed through the detector inference wrapper, any detector-specific resize or isotropic letterbox padding applied to fit $(W_D,H_D)$ is inverted back to local crop coordinates $(x,y)$ and translated by $(x_0,y_0)$ into native source-image coordinates $(x+x_0,y+y_0)$. Inside \prf{} interior admission, Eq.~\eqref{eq:app-positive-region} then tests containment strictly in native image pixels.

Contracting internal crop edges by $\delta_{\mathrm{img}}$ rejects seam-bisected fragments whose boxes touch an internal window boundary, whereas exempting outer source-canvas borders preserves genuine objects touching the image frame. Furthermore, because rounding continuous lattice coordinates to integer pixels slightly alters window area, \lws{} recomputes the realized integer-crop zoom
\begin{equation}
\widehat z_c=\sqrt{\frac{W_i H_i}{\max(|c|,1)}},
\label{eq:app-rounded-zoom}
\end{equation}
and re-evaluates exact interior containment and closure quality $q_{cj}$ at $\widehat z_c$ prior to max-heap insertion. Recomputing gain at $\widehat z_c$ ensures that the planner scores the exact discrete pixel slice sent to the GPU and prevents distinct continuous lattice points that round to the same integer box from consuming duplicate calls.

\subsection{Dual-Origin 2D Difference-Array Sweep}
\label{app:algorithmic-identities}

At each allocation step on image $i$ with $n_i$ coarse hypotheses, the planner solves $c_i^\star\in\arg\max_{c\in\mathcal C_i\setminus S_i}\Delta(c\mid S_i)$ across the discrete zoom lattice $\mathcal Z$ in Table~\ref{tab:constants} and dense spatial translations. Naive pairwise evaluation between $n_i$ hypotheses and $G_z$ candidate origins at zoom $z$ requires $O(n_i G_z)$ operations per zoom. We decouple scale closure from spatial containment so all $G_z$ lattice origins at zoom $z$ are scored simultaneously in linear grid time.

Fix nominal zoom $z\in\mathcal Z$ on image $(W_i,H_i)$ with window size $w_z=\min(W_i,W_i/z)$ and $h_z=\min(H_i,H_i/z)$. Stepping origins by $(\delta_x,\delta_y)$ with right and bottom border clipping forms a $G_z=N_x N_y$ lattice whose integer-rounded windows induce monotone positive-region limits $\underline{p}_{k_x}^x,\overline{p}_{k_x}^x$ and $\underline{p}_{k_y}^y,\overline{p}_{k_y}^y$ via Eq.~\eqref{eq:app-positive-region}.

Because both $\underline{p}_k^x$ and $\overline{p}_k^x$ are monotone non-decreasing in the origin index $k$, the set of lattice origins $(k_x,k_y)$ whose positive region $\mathcal P_c$ strictly encloses a given hypothesis box $b_j=[x_{j,1},y_{j,1},x_{j,2},y_{j,2}]$ forms a single dual axis-aligned index rectangle $[k_{j,x}^-,k_{j,x}^+]\times[k_{j,y}^-,k_{j,y}^+]$ given in closed form by
\begin{equation}
\begin{aligned}[b]
k_{j,x}^-&=\min\{k:\overline{p}_k^x>x_{j,2}\}, &
k_{j,x}^+&=\max\{k:\underline{p}_k^x<x_{j,1}\},\\
k_{j,y}^-&=\min\{k:\overline{p}_k^y>y_{j,2}\}, &
k_{j,y}^+&=\max\{k:\underline{p}_k^y<y_{j,1}\}.
\end{aligned}
\label{eq:app-dual-bounds}
\end{equation}
Monotonicity allows each of the four index bounds in Eq.~\eqref{eq:app-dual-bounds} to be located by binary search in $O(\log N_x+\log N_y)$ time. Denoting the inclusive index corners by $(l_x,r_x,l_y,r_y)=(k_{j,x}^-,k_{j,x}^+,k_{j,y}^-,k_{j,y}^+)$ and the hypothesis's zoom-$z$ closure credit by $v_j(z\mid S_i)=u_j R_j(S_i)h(a_j,z)$, each hypothesis with a nonempty dual rectangle writes four corner impulses into a zero-initialized $(N_x+1)\times(N_y+1)$ difference array $D^{(z)}$:
\begin{equation}
\begin{aligned}[b]
(l_x,l_y),\ (r_x+1,r_y+1)&:\quad +v_j(z\mid S_i),\\
(l_x,r_y+1),\ (r_x+1,l_y)&:\quad -v_j(z\mid S_i).
\end{aligned}
\label{eq:app-diff-impulses}
\end{equation}
Executing a single 2D prefix sum over $D^{(z)}$ then recovers the exact nominal marginal gain of every lattice window $c=c(k_x,k_y,z)$ simultaneously:
\begin{equation}
\begin{aligned}[b]
G^{(z)}(k_x,k_y)&=\sum_{u=0}^{k_x}\sum_{v=0}^{k_y}D^{(z)}(u,v)\\
&=\sum_{j:i(j)=i}v_j(z\mid S_i)\,\mathbf{1}[\,b_j\subseteq\mathcal P_c\,].
\end{aligned}
\label{eq:app-prefix-sum}
\end{equation}
Through Eqs.~\eqref{eq:app-dual-bounds}--\eqref{eq:app-prefix-sum}, scoring a zoom level takes $O(n_i\log G_z+G_z)$ scalar operations rather than $O(n_i G_z)$ pairwise checks. With rounded-zoom re-evaluation via Eq.~\eqref{eq:app-rounded-zoom} and $C_i=\sum_{z\in\mathcal Z}O(n_i\log G_z+G_z)$ per multi-zoom scan, collection initialization, $Q_B\le L$ single-image refreshes on popped images $i_t$, and max-heap updates run in total CPU time
\begin{equation}
T_{\mathrm{plan}}=\sum_{i=1}^N C_i+\sum_{t=1}^{Q_B}C_{i_t}+O(Q_B\log N),
\label{eq:app-total-complexity}
\end{equation}
using only $O(\sum_i n_i+N+Q_B+\max_z G_z)$ memory without ever materializing a $|\mathcal C_i|\times n_i$ incidence matrix. This algorithmic reduction explains why multi-zoom planning adds under $1.8$\,ms per image on a single CPU core in Table~\ref{tab:efficiency-profiling}, keeping CPU planning minor beside GPU inference.

\subsection{Global Threshold Equivalence of SBA}
\label{app:offline-threshold}

Offline benchmarks provide all $N$ images upfront so \sba{} can pop the highest marginal gains across the entire collection via a global max-heap, whereas streaming deployments must decide how many crops to allocate to each arriving frame $t$ immediately. To bridge offline and online allocation, we study the structural relationship between global heap selection and per-image marginal-gain thresholding.

Because the collection objective $F(S)=\sum_{i=1}^N F_i(S_i)$ separates additively across images and exact local greedy selection on each submodular component $F_i$ produces a monotonically non-increasing sequence of marginal gains, merging the $N$ per-image gain sequences inside a global max-heap for $L$ steps is mathematically equivalent to applying a single global gain cutoff $\lambda_L$ equal to the $L$-th popped key: every image $i$ receives the exact prefix of local greedy windows whose marginal gains satisfy $\Delta(c\mid S_i)\ge\lambda_L$ under consistent tie resolution at the boundary. While nominal-to-rounded zoom re-evaluation ($\widehat z_c$) in discrete \lws{} can introduce minor local non-monotonicities quantified in Appendix~\ref{sec:implementation-gap}, this threshold equivalence provides the direct theoretical justification for our causal online controller in Appendix~\ref{app:online-effective}: to approximate collection-wide \sba{} on an arriving stream without future lookahead, the online allocator simply replaces the unknown offline key $\lambda_L$ with a causal sliding-window quantile estimate $\lambda_t$ of recent marginal gains.

\subsection{Within-Class Order Invariance of PRF Squeeze}
\label{app:fusion-order}

In the final stage of Positive-Region Fusion (\prf{}), the expanded per-image candidate pool $\mathcal D_{2K}^{(i)}$ (containing up to $2K$ NMS-filtered detections with fused confidences $s_d$) must be truncated to the evaluation cap of $K$ outputs. On multi-class benchmarks, high-zoom re-observation of dense crowds multiplies medium-confidence predictions of dominant categories (such as pedestrians or cars), which can push valid detections of rarer categories out of a raw score top-$K$ cut. At the same time, altering predicted confidence scores to rebalance categories could corrupt precision-recall ordering. We therefore examine how class-mass balanced squeezing in Eq.~\eqref{eq:squeeze} affects per-class detection rankings.

For each candidate $d\in\mathcal D_{2K}^{(i)}$ of predicted category $k_d$, \prf{} computes the class-mass balanced selection key
\begin{equation}
v_d=\frac{s_d}{\sqrt{\max(M_{k_d},\epsilon_M)}},
\label{eq:app-squeeze-key}
\end{equation}
where $\epsilon_M>0$ in Table~\ref{tab:constants} is the stability floor and $M_k$ sums the fused scores of class $k$ across $\bigcup_i\mathcal D_{2K}^{(i)}$. Crucially, \prf{} uses $v_d$ strictly to determine which $K$ detections per image enter the bounded output set $\mathcal D_K^{(i)}$, while emitting their original fused confidence scores $s_d$ unchanged. Because any two detections $d,d'$ belonging to the same category $k_d=k_{d'}=k$ are divided by the identical positive constant $\sqrt{\max(M_k,\epsilon_M)}>0$, Eq.~\eqref{eq:app-squeeze-key} satisfies the exact equivalence
\[
v_d>v_{d'}\iff s_d>s_{d'}.
\]
Consequently, class-balanced squeezing preserves within-class confidence rankings verbatim and reduces identically to standard score top-$K$ truncation on single-class datasets ($C=1$). This property helps our overall framework by protecting rare categories from top-$K$ eviction on imbalanced multi-class benchmarks such as VisDrone-DET without perturbing per-class precision-recall curves or single-class CrowdHuman results.

\section{From Coverage to Detection AP}
\label{app:coverage-to-ap}

The optimization guarantees in Appendix~\ref{app:coverage-guarantees} apply to the pre-inference geometric surrogate $F(S)$, whereas object detection is evaluated after crop inference by Average Precision ($\mathrm{AP@}K$). Because $F(S)$ is computed from coarse predictions alone before any crop is executed, increasing $F(S)$ cannot unconditionally guarantee an increase in AP if crop passes introduce high-scoring false positives or disrupt precision rankings inside the top-$K$ output limit. In this section, we study the theoretical and empirical coupling between pre-inference coverage $F(S)$ and post-inference detection AP: we derive an exact four-channel decomposition of interpolated COCO AP to reveal how planning and fusion interact, prove conditional rank-alignment and AP-transfer bounds, and validate these relationships across multi-detector plan suites.

\subsection{Four-Channel AP Decomposition and Conditional Bounds}
\label{sec:rank-alignment-theory}

\noindent\textbf{Exact Four-Channel Decomposition of Interpolated COCO AP.}
Standard COCO evaluation computes area under a 101-point interpolated precision-recall envelope averaged across categories and IoU thresholds, which obscures why naive cropping can degrade AP even when it recovers new true positives. To isolate the exact mechanisms through which a re-observation plan $S$ alters final accuracy relative to the coarse pass $\varnothing$, we decompose interpolated $\Delta\mathrm{AP}_{\mathrm{COCO}}@K(S)=\mathrm{AP}_{\mathrm{COCO}}@K(S)-\mathrm{AP}_{\mathrm{COCO}}@K(\varnothing)$ at the instance level.

For each category $k\in\{1,\ldots,C\}$ with $G_k>0$ nonignored ground-truth instances $\mathcal G_k$ and IoU thresholds $\mathcal U=\{0.50,0.55,\ldots,0.95\}$, let $I_{g,u}(S)\in\{0,1\}$ indicate whether target $g\in\mathcal G_k$ is matched at IoU $\ge u$ within the top-$K$ outputs of plan $S$. Under deterministic descending confidence sorting, let $w_{g,u}(S)\in[0,1]$ denote the running precision at $g$'s matched prediction rank (or zero if $I_{g,u}(S)=0$). With uniform instance weight $\bar w_k=(C G_k|\mathcal U|)^{-1}$, define non-interpolated instance AP and the 101-point precision-envelope interpolation residual over $\mathcal R_{101}=\{0,0.01,\ldots,1.00\}$ as
\begin{align*}
\mathrm{AP}_{\mathrm{ni}}@K(S)&=\sum\nolimits_{k,g,u}\bar w_k I_{g,u}(S)w_{g,u}(S),\\
\mathcal I_{\mathrm{env}}(S)&=\mathrm{AP}_{\mathrm{COCO}}@K(S)-\mathrm{AP}_{\mathrm{ni}}@K(S).
\end{align*}
Partitioning the instance contribution difference $I_{g,u}(S)w_{g,u}(S)-I_{g,u}(\varnothing)w_{g,u}(\varnothing)$ across newly gained matches ($I_{g,u}(S)=1,I_{g,u}(\varnothing)=0$), lost coarse matches ($I_{g,u}(S)=0,I_{g,u}(\varnothing)=1$), and retained matches ($I_{g,u}(S)=I_{g,u}(\varnothing)=1$) yields the exact four-channel decomposition
\begin{equation}
\begin{aligned}[b]
\Delta\mathrm{AP}_{\mathrm{COCO}}@K(S)={}&\mathcal L_{\mathrm{pos}}(S)-\mathcal E_{\mathrm{loss}}(S)\\
&+\mathcal D_{\mathrm{rank}}(S)+\Delta\mathcal I_{\mathrm{env}}(S),
\end{aligned}
\label{eq:ap-three-channel}
\end{equation}
where $[x]_+=\max(x,0)$, $\Delta I_{g,u}(S)=I_{g,u}(S)-I_{g,u}(\varnothing)$, $\Delta w_{g,u}(S)=w_{g,u}(S)-w_{g,u}(\varnothing)$, and the four signed channels are given by
\begin{equation}
\begin{aligned}[b]
\mathcal L_{\mathrm{pos}}(S)&=\sum_{k,g,u}\!\bar w_k\,w_{g,u}(S)\,[\Delta I_{g,u}(S)]_+\ge0,\\
\mathcal E_{\mathrm{loss}}(S)&=\sum_{k,g,u}\!\bar w_k\,w_{g,u}(\varnothing)\,[-\Delta I_{g,u}(S)]_+\ge0,\\
\mathcal D_{\mathrm{rank}}(S)&=\sum_{k,g,u}\!\bar w_k\,I_{g,u}(S)I_{g,u}(\varnothing)\,\Delta w_{g,u}(S),\\
\Delta\mathcal I_{\mathrm{env}}(S)&=\mathcal I_{\mathrm{env}}(S)-\mathcal I_{\mathrm{env}}(\varnothing).
\end{aligned}
\label{eq:ap-channel-defs}
\end{equation}
Equations~\eqref{eq:ap-three-channel}--\eqref{eq:ap-channel-defs} clarify the exact functional division of labor inside \method{}. Shared-Budget Complementary Planning (\scp{}) maximizes $F(S)$ to place crops where magnified views recover new true positives or sharpen localization across IoU thresholds, driving the positive recall-recovery channel $\mathcal L_{\mathrm{pos}}(S)$. However, if post-inference fusion admits seam-clipped false positives or allows one crowded category to monopolize the $K$-slot cap, valid coarse detections are evicted ($\mathcal E_{\mathrm{loss}}(S)>0$) and false positives inserted ahead of retained true positives depress running precision ($\mathcal D_{\mathrm{rank}}(S)<0$). Positive-Region Fusion (\prf{}) directly guards against both loss channels by enforcing interior containment $\mathcal P_c$, downweighting superseded coarse boxes rather than deleting them, and balancing class quotas.

When \prf{} stabilizes eviction and rank drift, net AP gain $\Delta\mathrm{AP@}K(S)$ concentrates around an increasing concave function of surrogate coverage $F(S)$. We therefore study two conditional propositions that formalize how surrogate ranking and greedy approximation transfer to detection AP whenever plan residuals are bounded by $\bar\zeta$.

\begin{proposition}[Ordinal Rank Alignment Under Perturbed Monotone Coupling]
\label{prop:rank-alignment}
Let plans $S^{(1)},\ldots,S^{(m)}$ ($m\ge2$) satisfy $F_1<\cdots<F_m$ with distinct AP gains $Y_a=\Delta\mathrm{AP@}K(S^{(a)})$ lying in a uniform band $|Y_a-\Psi(F_a)|\le\bar\zeta$ around a strictly increasing link $\Psi$ for $\bar\zeta\ge0$. Defining ambiguity neighborhood counts $b_a^-=\bigl|\{b<a:\Psi(F_a)-\Psi(F_b)\le2\bar\zeta\}\bigr|$, $b_a^+=\bigl|\{b>a:\Psi(F_b)-\Psi(F_a)\le2\bar\zeta\}\bigr|$, $b_a=\max(b_a^-,b_a^+)$, $b_{\max}=\max_a b_a$, and total ambiguous pairs $D_{\bar\zeta}=\bigl|\{(a,b):1\le a<b\le m,\,\Psi(F_b)-\Psi(F_a)\le2\bar\zeta\}\bigr|$, the ascending rank $\hat r_a$ of $Y_a$ satisfies $|\hat r_a-a|\le b_a$, and rank correlations obey
\begin{equation}
\begin{aligned}[b]
\tau_K&\ge1-\frac{4D_{\bar\zeta}}{m(m-1)},\\
\rho_S&\ge1-\frac{6\sum_{a=1}^m b_a^2}{m(m^2-1)}\ge1-\frac{6b_{\max}^2}{m^2-1},
\end{aligned}
\label{eq:spearman-lower-bound}
\end{equation}
between $F(S)$ and $\mathrm{AP@}K(S)$.
\end{proposition}
\begin{proof}
1.~\emph{Individual rank displacement.} For any pair $b<a$ satisfying $\Psi(F_a)-\Psi(F_b)>2\bar\zeta$, the triangle inequality guarantees $Y_a-Y_b\ge\Psi(F_a)-\Psi(F_b)-2\bar\zeta>0$. Thus $Y_b<Y_a$ holds for at least $(a-1)-b_a^-$ lower-surrogate plans, and symmetrically $Y_b>Y_a$ holds for at least $(m-a)-b_a^+$ higher-surrogate plans, confining the realized rank to $\hat r_a\in[a-b_a^-,a+b_a^+]\subseteq[a-b_a,a+b_a]$.

\noindent 2.~\emph{Correlation bounds.} Because only pairs inside $D_{\bar\zeta}$ can reverse order, at most $D_{\bar\zeta}$ of the $m(m-1)/2$ pairs are discordant, establishing the Kendall $\tau_K$ bound. Substituting $(\hat r_a-a)^2\le b_a^2\le b_{\max}^2$ into the untied Spearman identity $\rho_S=1-\frac{6}{m(m^2-1)}\sum_{a=1}^m(\hat r_a-a)^2$ yields Eq.~\eqref{eq:spearman-lower-bound}.
\end{proof}

\begin{proposition}[Conditional AP-Gain Transfer From Local Surrogate Greedy]
\label{prop:ap-transfer}
Under Proposition~\ref{prop:approx-greedy} with weighted search error $E_{\mathrm{LWS}}\allowbreak=\sum_{t=1}^L(1-1/L)^{L-t}\epsilon_t$ and $S_{\mathrm{AP}}^\star\in\arg\max_{|S|\le L}\Delta\mathrm{AP@}K(S)$, if $|\Delta\mathrm{AP@}K(S)\allowbreak-\Phi_{\Delta}(F(S))|\le\bar\zeta$ holds across $|S|\le L$ for a non-decreasing concave link $\Phi_{\Delta}:[0,\infty)\to[0,\infty)$ with $\Phi_{\Delta}(0)=0$, then
\begin{equation}
\begin{aligned}[b]
\Delta\mathrm{AP@}K(S_L)\ge{}&\Bigl(1-\frac1e\Bigr)\Delta\mathrm{AP@}K(S_{\mathrm{AP}}^\star)\\
&-\Bigl[\Bigl(2-\frac1e\Bigr)\bar\zeta+\Phi_{\Delta}(E_{\mathrm{LWS}})\Bigr].
\end{aligned}
\label{eq:conditional-ap-transfer}
\end{equation}
\end{proposition}
\begin{proof}
Concavity of $\Phi_{\Delta}$ with $\Phi_{\Delta}(0)=0$ implies star-shaped scaling $\Phi_{\Delta}(\alpha x)\ge\alpha\Phi_{\Delta}(x)$ for $\alpha\in[0,1]$ and subadditivity $\Phi_{\Delta}(u-e)\ge\Phi_{\Delta}(u)-\Phi_{\Delta}(e)$ for $u\ge e\ge0$. Setting $\alpha=1-1/e$ and noting $F(S^\star)\ge F(S_{\mathrm{AP}}^\star)$, Proposition~\ref{prop:approx-greedy} gives $F(S_L)\ge\alpha F(S_{\mathrm{AP}}^\star)-E_{\mathrm{LWS}}$. Applying monotonicity, subadditivity (since $\Phi_{\Delta}(F(S_L))\ge0$ holds directly whenever $\alpha F(S_{\mathrm{AP}}^\star)<E_{\mathrm{LWS}}$), star-shaped scaling, and $|\Delta\mathrm{AP@}K-\Phi_{\Delta}(F)|\le\bar\zeta$ at both $S_L$ and $S_{\mathrm{AP}}^\star$ yields Eq.~\eqref{eq:conditional-ap-transfer}.
\end{proof}

\begin{remark}[Conditional Boundaries and Worst-Case Counterexamples]
Propositions~\ref{prop:rank-alignment} and~\ref{prop:ap-transfer} make explicit why surrogate optimization transfers to detection accuracy only conditionally on bounded link residuals $\bar\zeta$. Two analytical counterexamples under frozen \prf{} and standard COCO evaluation illustrate what happens when those coupling assumptions are violated. First, if a crop enclosing a valid small coarse hypothesis injects $K$ higher-scoring false positives that also pass interior and scale checks, top-$K$ truncation evicts the coarse true positive via $\mathcal E_{\mathrm{loss}}$, causing $F(S)$ to increase while $\mathrm{AP@}100$ drops from $100$ to $0$. Second, if an isolated small target fails to emit any coarse hypothesis above $s_{\min}$, every candidate window has $F(S)\equiv0$, yet an externally supplied crop over that target raises $\mathrm{AP@}100$ from $0$ to $100$ via $\mathcal L_{\mathrm{pos}}$. Thus, $F(S)$ provides an effective planning objective precisely because \prf{} suppresses false-positive eviction and natural small targets cluster near weak coarse cues.
\end{remark}

\subsection{Empirical Link Calibration, Plan Ranking, and AP Attribution}
\label{app:theory-link-audits}

\begin{figure*}[!t]
\centering
\includegraphics[width=\textwidth]{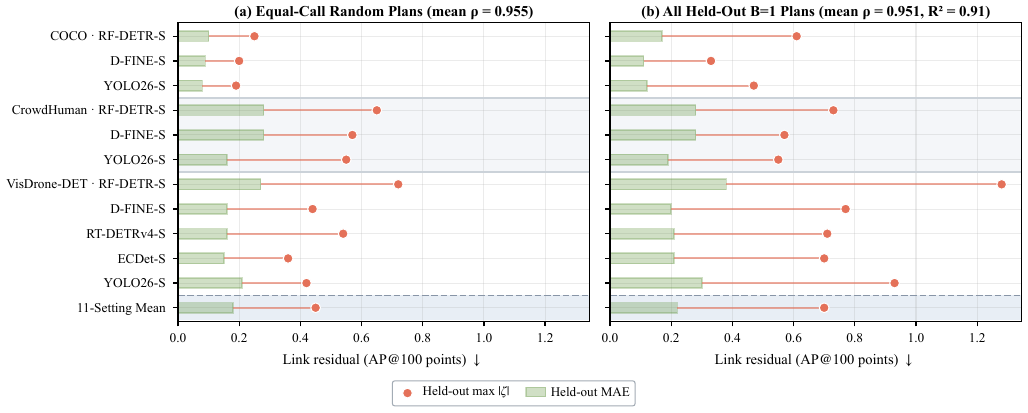}
\caption{\textbf{Out-of-Sample Concave Link Calibration Across $11$ Small-Detector Settings.} Held-out MAE (bars) and maximum prediction residual $|\zeta|$ (circles) on unseen equal-call random plans (left) and $B{=}1$ held-out plans (right) across COCO, CrowdHuman, and VisDrone-DET.}
\label{fig:calibration-residuals}
\end{figure*}

\begin{figure*}[!t]
\centering
\begin{subfigure}[b]{0.485\textwidth}
  \centering
  \includegraphics[width=\textwidth]{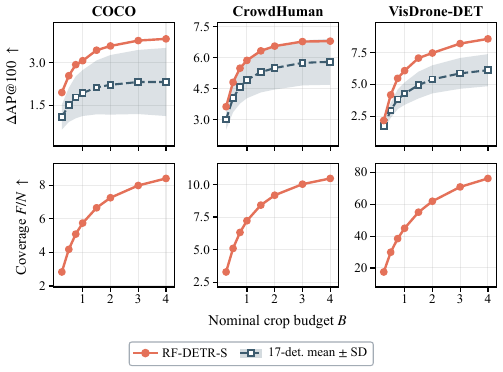}
  \caption{Multi-budget AP gains and coverage objective $F/N$}
  \label{fig:coverage-scaling}
\end{subfigure}\hfill
\begin{subfigure}[b]{0.485\textwidth}
  \centering
  \includegraphics[width=\textwidth]{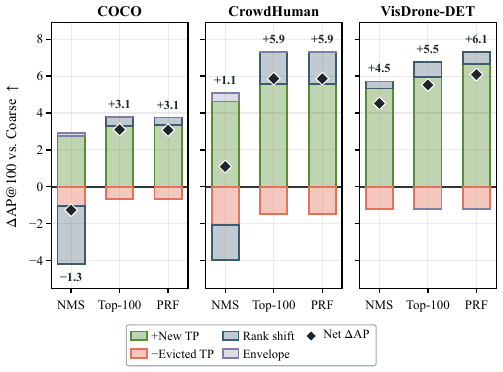}
  \caption{Four-channel AP decomposition on identical $B{=}1$ views}
  \label{fig:ap-attribution}
\end{subfigure}
\caption{\textbf{Surrogate Coverage Scaling and Four-Channel AP Attribution.} \textbf{(a)}~Concave per-image coverage $F/N$ tracks RF-DETR-S and $17$-detector mean$\pm$SD gains across $B\in[0.25,4]$. \textbf{(b)}~Exact channel decomposition via Eq.~\eqref{eq:ap-three-channel} on matched $B{=}1$ crops shows \prf{} eliminates eviction ($-\mathcal E_{\mathrm{loss}}$) and negative rank drift ($\mathcal D_{\mathrm{rank}}$).}
\label{fig:surrogate-and-attribution}
\end{figure*}

\noindent\textbf{Link Calibration and Equal-Call Ranking.}
To test whether the monotone concave link $\Phi_{\Delta}$ in Propositions~\ref{prop:rank-alignment} and~\ref{prop:ap-transfer} generalizes to unseen schedules at fixed call budgets, we fit $\Phi_{\Delta}$ across $11$ Small-detector settings and evaluate two held-out plan suites: equal-call random plans and $B{=}1$ baseline/ablation plans. The results are reported in Figure~\ref{fig:calibration-residuals}. Across all $11$ settings, the calibrated link achieves a mean out-of-sample MAE of $0.18$ AP on equal-call plans and $0.22$ AP on $B{=}1$ plans, with $R^2=0.91$ and Spearman rank correlation $\rho_S=0.95$. Holding call counts strictly equal at $B{=}1$ across all $17$ detectors likewise yields mean rank correlations of $\rho_S=0.95$, $0.97$, and $0.88$ on COCO, CrowdHuman, and VisDrone-DET, while $28{\times}$ grid refinement shifts validation AP by at most $0.56$ points, an order of magnitude smaller than removing multiplicative residual discounting.

\smallskip\noindent\textbf{Multi-Budget Scaling, AP Attribution, and Residual Regimes.}
To examine how planned coverage translates into post-inference accuracy across merging rules, we evaluate multi-budget surrogate trajectories over $B\in[0.25,4]$ and compare three post-processors on identical $B{=}1$ \method{} crop views: plain NMS, raw score top-$K$ truncation, and full \prf{}. The results are reported in Figure~\ref{fig:surrogate-and-attribution}. First, Figure~\ref{fig:coverage-scaling} shows that per-image coverage $F(S)/N$ grows concavely with $B$, driving monotonic RF-DETR-S and $17$-detector mean AP gains across all three benchmarks. Second, decomposing identical $B{=}1$ views via Eq.~\eqref{eq:ap-three-channel} in Figure~\ref{fig:ap-attribution} explains why fusion alignment is essential: plain NMS suffers severe coarse-match eviction ($\mathcal E_{\mathrm{loss}}=1.02$--$2.06$ AP) and negative precision rank drift ($\mathcal D_{\mathrm{rank}}=-3.15$ on COCO and $-1.91$ on CrowdHuman), pulling net COCO accuracy $1.26$ AP below Coarse. Compared with plain NMS and score top-$K$, \prf{} turns $\mathcal D_{\mathrm{rank}}$ positive across all three benchmarks ($+0.42$ to $+1.73$ AP), lifts VisDrone-DET $\mathcal L_{\mathrm{pos}}$ from $5.95$ to $6.67$ AP via class-balanced squeezing, and raises $11$-setting mean net $B{=}1$ gain from $+1.32$ AP to $+3.88$ AP. Across cropped frames, realized true-positive gains follow $F_i$, while deviations stem mainly from top-$K$ slot saturation, recovery of adjacent unsupported tiny instances on VisDrone-DET, or high-zoom background clutter as shown in Row~5 of Figure~\ref{fig:qualitative-examples}.

\section{More Quantitative Results}
\label{app:protocol}

This section details the evaluation protocol and presents supporting quantitative comparisons across detector families, official high-density output limits, open- versus closed-loop planning, native slicing pipelines, deployment latency profiling, and causal online streaming.

\subsection{Datasets, Frozen Checkpoints, and Evaluation Protocol}
\label{app:checkpoint-protocol}

\begin{figure*}[!t]
\centering
\begin{subfigure}[b]{0.325\textwidth}
  \centering
  \includegraphics[width=\textwidth]{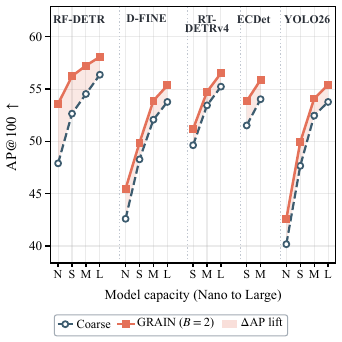}
  \caption{COCO val2017}
  \label{fig:boot-coco}
\end{subfigure}\hfill
\begin{subfigure}[b]{0.325\textwidth}
  \centering
  \includegraphics[width=\textwidth]{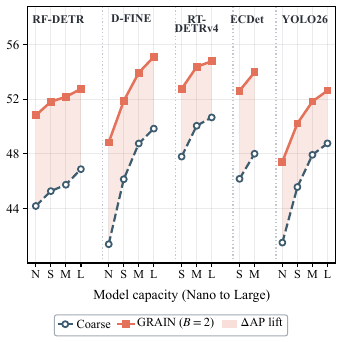}
  \caption{CrowdHuman validation}
  \label{fig:boot-ch}
\end{subfigure}\hfill
\begin{subfigure}[b]{0.325\textwidth}
  \centering
  \includegraphics[width=\textwidth]{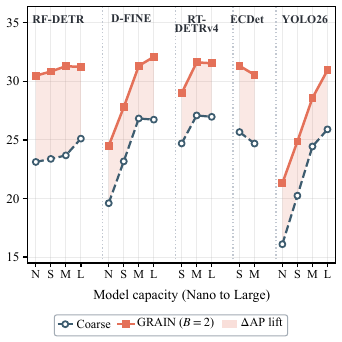}
  \caption{VisDrone-DET validation}
  \label{fig:boot-vd}
\end{subfigure}
\caption{\textbf{Accuracy Scaling Across Five Detector Families (Nano to Large).} Native Coarse (dashed) vs.\ \method{} at $B{=}2$ (solid) across all $17$ frozen checkpoints, where shaded coral bands highlight $\Delta$AP lift at every model capacity.}
\label{fig:cross-model-scaling}
\end{figure*}

\begin{table*}[!t]
\caption{\textbf{Paired Image-Bootstrap Gains Across \SWEEPdetectors{} Frozen Detectors.} Mean$\pm$SD AP@100 gains over native Coarse ($\Delta$Coarse$\uparrow$) and call-matched Tile ($\Delta$Tile$\uparrow$) at $B\in\{1,2\}$. Bold marks the best gain per family, and the bottom rows report the \SWEEPdetectors{}-detector mean$\pm$SD and win count ($>0$).}
\label{tab:sweep-bootstrap-family}
\centering
\begingroup
\newcommand{\bsd}[1]{{\textcolor{black!55}{\fontsize{5.6}{6.6}\selectfont$\pm$#1}}}
\scriptsize
\setlength{\extrarowheight}{0.65pt}
\setlength{\tabcolsep}{0.4pt}
\newcolumntype{B}{>{\centering\arraybackslash\cellcolor{ntMethodSoft}}X}
\begin{nicetable}
\begin{tabularx}{\textwidth}{llYYBBYYBBYYBB}
\toprule
\rowcolor{ntHeader}
 & & \multicolumn{4}{c}{COCO val2017} & \multicolumn{4}{c}{CrowdHuman validation} & \multicolumn{4}{c}{VisDrone-DET validation} \\
\tintrule\cmidrule(lr){3-6}\cmidrule(lr){7-10}\cmidrule(lr){11-14}
\rowcolor{ntHeader}
Family & Size & \multicolumn{2}{c}{$\Delta$Coarse$\uparrow$} & \multicolumn{2}{c}{\cellcolor{ntMethod}$\Delta$Tile$\uparrow$} & \multicolumn{2}{c}{$\Delta$Coarse$\uparrow$} & \multicolumn{2}{c}{\cellcolor{ntMethod}$\Delta$Tile$\uparrow$} & \multicolumn{2}{c}{$\Delta$Coarse$\uparrow$} & \multicolumn{2}{c}{\cellcolor{ntMethod}$\Delta$Tile$\uparrow$} \\
\rowcolor{ntHeader}
 & & $B{=}1$ & $B{=}2$ & \cellcolor{ntMethod}$B{=}1$ & \cellcolor{ntMethod}$B{=}2$ & $B{=}1$ & $B{=}2$ & \cellcolor{ntMethod}$B{=}1$ & \cellcolor{ntMethod}$B{=}2$ & $B{=}1$ & $B{=}2$ & \cellcolor{ntMethod}$B{=}1$ & \cellcolor{ntMethod}$B{=}2$ \\
\midrule
\multirow{4}{*}{RF-DETR \venue{(ICLR'26)}} & N & \textbf{+4.49}\bsd{0.20} & \textbf{+5.69}\bsd{0.22} & \textbf{+1.85}\bsd{0.18} & \textbf{+1.84}\bsd{0.16} & \textbf{+5.90}\bsd{0.18} & \textbf{+6.64}\bsd{0.18} & +3.71\bsd{0.11} & +2.86\bsd{0.09} & +5.75\bsd{0.25} & +7.36\bsd{0.26} & +0.71\bsd{0.23} & +0.52\bsd{0.21} \\
 & S & +3.08\bsd{0.17} & +3.60\bsd{0.19} & +1.23\bsd{0.14} & +0.91\bsd{0.13} & +5.86\bsd{0.18} & +6.55\bsd{0.19} & \textbf{+3.81}\bsd{0.12} & \textbf{+2.91}\bsd{0.10} & \textbf{+6.08}\bsd{0.27} & +7.47\bsd{0.28} & \textbf{+0.82}\bsd{0.26} & +0.39\bsd{0.22} \\
 & M & +2.41\bsd{0.15} & +2.72\bsd{0.16} & +1.10\bsd{0.12} & +0.79\bsd{0.13} & +5.74\bsd{0.18} & +6.44\bsd{0.19} & +3.67\bsd{0.12} & +2.86\bsd{0.10} & +5.84\bsd{0.26} & \textbf{+7.61}\bsd{0.28} & +0.75\bsd{0.23} & \textbf{+0.77}\bsd{0.19} \\
 & L & +1.66\bsd{0.13} & +1.72\bsd{0.14} & +0.96\bsd{0.10} & +0.72\bsd{0.10} & +5.27\bsd{0.18} & +5.88\bsd{0.18} & +3.50\bsd{0.11} & +2.68\bsd{0.09} & +4.85\bsd{0.26} & +6.12\bsd{0.29} & +0.28\bsd{0.23} & $-$0.01\bsd{0.18} \\
\midrule
\multirow{4}{*}{D-FINE \venue{(ICLR'25)}} & N & \textbf{+2.48}\bsd{0.15} & \textbf{+2.87}\bsd{0.16} & \textbf{+1.10}\bsd{0.13} & \textbf{+0.72}\bsd{0.12} & \textbf{+6.59}\bsd{0.19} & \textbf{+7.47}\bsd{0.20} & \textbf{+3.93}\bsd{0.13} & \textbf{+3.02}\bsd{0.10} & +4.05\bsd{0.21} & +4.90\bsd{0.21} & \textbf{+0.96}\bsd{0.19} & \textbf{+0.65}\bsd{0.18} \\
 & S & +1.26\bsd{0.15} & +1.59\bsd{0.16} & +0.56\bsd{0.13} & +0.47\bsd{0.15} & +5.14\bsd{0.17} & +5.74\bsd{0.17} & +3.34\bsd{0.11} & +2.48\bsd{0.09} & +3.60\bsd{0.22} & +4.62\bsd{0.23} & +0.60\bsd{0.16} & +0.41\bsd{0.15} \\
 & M & +1.64\bsd{0.15} & +1.82\bsd{0.15} & +1.00\bsd{0.12} & +0.71\bsd{0.13} & +4.67\bsd{0.17} & +5.19\bsd{0.17} & +3.21\bsd{0.10} & +2.38\bsd{0.08} & +3.58\bsd{0.22} & +4.49\bsd{0.23} & +0.57\bsd{0.18} & +0.30\bsd{0.17} \\
 & L & +1.44\bsd{0.15} & +1.58\bsd{0.16} & +0.68\bsd{0.09} & +0.39\bsd{0.11} & +4.76\bsd{0.17} & +5.26\bsd{0.17} & +3.18\bsd{0.10} & +2.32\bsd{0.08} & \textbf{+4.39}\bsd{0.22} & \textbf{+5.36}\bsd{0.25} & +0.74\bsd{0.17} & +0.39\bsd{0.15} \\
\midrule
\multirow{3}{*}{RT-DETRv4 \venue{(ECCV'26)}} & S & \textbf{+1.43}\bsd{0.15} & \textbf{+1.55}\bsd{0.18} & \textbf{+0.81}\bsd{0.12} & \textbf{+0.52}\bsd{0.14} & \textbf{+4.42}\bsd{0.16} & \textbf{+4.95}\bsd{0.16} & \textbf{+3.06}\bsd{0.10} & \textbf{+2.32}\bsd{0.08} & +3.30\bsd{0.21} & +4.29\bsd{0.22} & +0.30\bsd{0.16} & +0.19\bsd{0.16} \\
 & M & +1.16\bsd{0.14} & +1.30\bsd{0.14} & +0.66\bsd{0.12} & +0.49\bsd{0.14} & +3.88\bsd{0.15} & +4.32\bsd{0.16} & +2.90\bsd{0.09} & +2.19\bsd{0.08} & \textbf{+3.68}\bsd{0.23} & +4.52\bsd{0.25} & \textbf{+0.81}\bsd{0.18} & +0.44\bsd{0.17} \\
 & L & +1.14\bsd{0.17} & +1.29\bsd{0.17} & +0.54\bsd{0.16} & +0.24\bsd{0.11} & +3.75\bsd{0.15} & +4.12\bsd{0.16} & +2.69\bsd{0.09} & +1.90\bsd{0.07} & +3.58\bsd{0.22} & \textbf{+4.57}\bsd{0.23} & +0.55\bsd{0.18} & \textbf{+0.56}\bsd{0.17} \\
\midrule
\multirow{2}{*}{ECDet \venue{(TMLR'26)}} & S & \textbf{+1.98}\bsd{0.14} & \textbf{+2.32}\bsd{0.17} & \textbf{+1.05}\bsd{0.13} & \textbf{+0.75}\bsd{0.13} & \textbf{+5.67}\bsd{0.19} & \textbf{+6.45}\bsd{0.19} & \textbf{+3.60}\bsd{0.12} & \textbf{+2.83}\bsd{0.09} & +4.49\bsd{0.23} & +5.66\bsd{0.25} & +0.88\bsd{0.22} & +0.81\bsd{0.19} \\
 & M & +1.60\bsd{0.15} & +1.85\bsd{0.15} & +0.62\bsd{0.12} & +0.47\bsd{0.11} & +5.30\bsd{0.18} & +5.97\bsd{0.18} & +3.51\bsd{0.11} & +2.66\bsd{0.09} & \textbf{+4.72}\bsd{0.24} & \textbf{+5.87}\bsd{0.25} & \textbf{+1.02}\bsd{0.20} & \textbf{+0.89}\bsd{0.18} \\
\midrule
\multirow{4}{*}{YOLO26 \venue{(arXiv'26)}} & N & \textbf{+2.09}\bsd{0.16} & \textbf{+2.41}\bsd{0.16} & \textbf{+1.01}\bsd{0.15} & \textbf{+0.70}\bsd{0.13} & \textbf{+5.33}\bsd{0.18} & \textbf{+5.93}\bsd{0.18} & \textbf{+3.39}\bsd{0.11} & \textbf{+2.65}\bsd{0.10} & +4.06\bsd{0.26} & \textbf{+5.21}\bsd{0.26} & \textbf{+0.88}\bsd{0.23} & \textbf{+0.71}\bsd{0.22} \\
 & S & +2.02\bsd{0.17} & +2.27\bsd{0.18} & \textbf{+1.01}\bsd{0.15} & +0.62\bsd{0.16} & +4.24\bsd{0.16} & +4.66\bsd{0.17} & +2.84\bsd{0.10} & +2.11\bsd{0.09} & +3.63\bsd{0.23} & +4.62\bsd{0.24} & +0.28\bsd{0.24} & +0.00\bsd{0.23} \\
 & M & +1.56\bsd{0.14} & +1.62\bsd{0.15} & +0.84\bsd{0.12} & +0.50\bsd{0.12} & +3.58\bsd{0.15} & +3.91\bsd{0.16} & +2.66\bsd{0.10} & +1.93\bsd{0.09} & +3.13\bsd{0.23} & +4.13\bsd{0.24} & +0.17\bsd{0.22} & $-$0.04\bsd{0.19} \\
 & L & +1.38\bsd{0.11} & +1.63\bsd{0.13} & +0.83\bsd{0.10} & +0.61\bsd{0.11} & +3.51\bsd{0.15} & +3.87\bsd{0.16} & +2.48\bsd{0.09} & +1.88\bsd{0.09} & \textbf{+4.07}\bsd{0.24} & +5.05\bsd{0.27} & +0.79\bsd{0.21} & +0.50\bsd{0.19} \\
\midrule
\rowcolor{ntHeader}
\multicolumn{2}{l}{\textbf{17-Det.\ Mean}} & \textbf{+1.93}\bsd{0.82} & \textbf{+2.22}\bsd{1.05} & \textbf{+0.93}\bsd{0.30} & \textbf{+0.67}\bsd{0.33} & \textbf{+4.92}\bsd{0.89} & \textbf{+5.49}\bsd{1.04} & \textbf{+3.26}\bsd{0.42} & \textbf{+2.47}\bsd{0.37} & \textbf{+4.28}\bsd{0.88} & \textbf{+5.40}\bsd{1.10} & \textbf{+0.65}\bsd{0.25} & \textbf{+0.44}\bsd{0.28} \\
\multicolumn{2}{l}{\textit{Wins ($>0$)}} & \textbf{17 / 17} & \textbf{17 / 17} & \textbf{17 / 17} & \textbf{17 / 17} & \textbf{17 / 17} & \textbf{17 / 17} & \textbf{17 / 17} & \textbf{17 / 17} & \textbf{17 / 17} & \textbf{17 / 17} & \textbf{17 / 17} & \textbf{14 / 17} \\
\bottomrule
\end{tabularx}
\end{nicetable}
\endgroup
\end{table*}

\noindent\textbf{Benchmarks and Statistical Protocol.}
We evaluate on Microsoft COCO val2017 ($5{,}000$ images)~\cite{lin2014microsoft}, CrowdHuman validation ($4{,}370$ images, full-body \texttt{fbox})~\cite{shao2018crowdhuman}, VisDrone-DET validation ($548$ images) and held-out test-dev ($1{,}610$ images)~\cite{zhu2021detection}, and AI-TOD-v2 validation ($2{,}804$ images) and held-out test ($14{,}018$ images)~\cite{xu2022detecting}. Standard COCO-style AP averages over IoU $0.50{:}0.05{:}0.95$ at $K{=}100$ unless an official higher capacity ($K{=}500$) is specified, CrowdHuman $\mathrm{MR}^{-2}$ reports log-average miss rate over $[10^{-2},1]$ FPPI at IoU $0.5$, and paired uncertainty estimates use $2{,}000$ image-bootstrap resamples.

\smallskip\noindent\textbf{Frozen Checkpoints, Hardware, and Licenses.}
Coarse and crop passes share identical frozen weights and native input canvas $(W_D,H_D)$. COCO uses official pretrained weights across all $17$ checkpoints, while CrowdHuman, VisDrone-DET, and AI-TOD-v2 checkpoints are trained once under each family's standard recipe and frozen. Accuracy sweeps run in PyTorch FP32, and end-to-end latency profiling executes TensorRT FP16 engines on a single NVIDIA RTX~4090 paired with single-thread Numba CPU planning. All datasets and pretrained detector families are used under their public research licenses, and source code with frozen manifests will be released.

\subsection{Cross-Detector Scaling and Official-Limit Evaluation}
\label{app:revision-evidence}

\noindent\textbf{Architecture Scaling Across Five Detector Families.}
To assess whether shared-budget re-observation generalizes across detector paradigms and model capacities, we compare \method{} at $B\in\{1,2\}$ against native Coarse and call-matched Tile under shared \prf{} across $17$ frozen Nano-to-Large checkpoints spanning RF-DETR~\cite{robinson2026rf}, D-FINE~\cite{peng2025d}, RT-DETRv4~\cite{liao2026rt}, ECDet~\cite{liu2026edgecrafter}, and YOLO26~\cite{jocher2026yolo26}. The results are reported in Figure~\ref{fig:cross-model-scaling} and Table~\ref{tab:sweep-bootstrap-family}. At $B{=}1$, \method{} outperforms call-matched Tile on all $17$ detectors across all three benchmarks by $+0.93$, $+3.26$, and $+0.65$ AP@100 on average, and at $B{=}2$ it gains $+2.22$, $+5.49$, and $+5.40$ AP@100 over Coarse while winning $48$ of $51$ Tile comparisons (with the three dense VisDrone-DET near-ties within $-0.04$ to $0.00$ AP). On COCO and CrowdHuman, gains over Coarse peak on Nano checkpoints and taper modestly as larger backbones resolve medium instances coarsely, whereas on aerial VisDrone-DET gains remain equally strong or peak on Medium and Large models because distant tiny vehicles remain pixel-limited across all backbone scales.

\begin{table*}[!t]
\centering\scriptsize\setlength{\tabcolsep}{2.4pt}
\caption{\textbf{Official High-Density Capacities and Zero-Tuning AI-TOD-v2 Transfer (RF-DETR-S).} \textbf{(a)}~CrowdHuman AP$\uparrow$ and official canvas-clipped $\mathrm{MR}^{-2}\downarrow$ at $K{=}100$ and $K{=}500$. \textbf{(b)}~AI-TOD-v2 validation and held-out test AP$\uparrow$ at $K{=}100$ and $K{=}500$ without parameter retuning. Bold marks the best score per budget group.}
\label{tab:official-limits}
\begin{minipage}[t]{0.485\textwidth}
\vspace{0pt}\centering
\begin{nicetable}
\setlength{\extrarowheight}{0.5pt}
\begin{tabularx}{\linewidth}{lcYYYY}
\toprule
\rowcolor{ntHeader}
\multicolumn{6}{l}{\textbf{(a) CrowdHuman Validation}} \\
\tintrule\cmidrule(lr){1-6}
\rowcolor{ntHeader}
 & & \multicolumn{2}{c}{$K{=}100$} & \multicolumn{2}{c}{$K{=}500$} \\
\tintrule\cmidrule(lr){3-4}\cmidrule(lr){5-6}
\rowcolor{ntHeader}
Pipeline & $B$ & AP$\uparrow$ & MR$^{-2}\downarrow$ & AP$\uparrow$ & MR$^{-2}\downarrow$ \\
\midrule
Coarse & 0 & 45.27 & 44.44 & 46.85 & 44.17 \\
\cmidrule(lr){1-6}
Tile~\cite{akyon2022sahi} & 1 & 47.32 & 40.81 & 48.81 & 40.41 \\
\rowcolor{ntMethod}
\textbf{\method{}} & 1 & \textbf{51.13} & \textbf{36.22} & \textbf{53.03} & \textbf{35.45} \\
\cmidrule(lr){1-6}
Tile~\cite{akyon2022sahi} & 2 & 48.91 & 39.26 & 50.53 & 38.77 \\
SAHI+\cgs{}+\sba{} & 2 & 49.48 & 38.40 & 51.24 & 37.88 \\
Dyn.\ Tiling+\cgs{}+\sba{} & 2 & 50.84 & 37.13 & 52.70 & 36.41 \\
DAHI+\cgs{}+\sba{} & 2 & 51.55 & 36.18 & 53.44 & 35.44 \\
\rowcolor{ntMethod}
\textbf{\method{}} & 2 & \textbf{51.83} & \textbf{35.46} & \textbf{53.73} & \textbf{34.61} \\
\cmidrule(lr){1-6}
\rowcolor{ntMethod}
\textbf{\method{}} & 4 & \textbf{52.07} & 35.97 & \textbf{54.05} & 35.08 \\
\bottomrule
\end{tabularx}
\end{nicetable}
\end{minipage}\hfill
\begin{minipage}[t]{0.495\textwidth}
\vspace{0pt}\centering
\begin{nicetable}
\setlength{\extrarowheight}{0.5pt}
\begin{tabularx}{\linewidth}{lcYYYY}
\toprule
\rowcolor{ntHeader}
\multicolumn{6}{l}{\textbf{(b) AI-TOD-v2 Validation \& Held-Out Test}} \\
\tintrule\cmidrule(lr){1-6}
\rowcolor{ntHeader}
 & & \multicolumn{2}{c}{Validation} & \multicolumn{2}{c}{Held-Out Test} \\
\tintrule\cmidrule(lr){3-4}\cmidrule(lr){5-6}
\rowcolor{ntHeader}
Pipeline & $B$ & AP$_{100}\uparrow$ & AP$_{500}\uparrow$ & AP$_{100}\uparrow$ & AP$_{500}\uparrow$ \\
\midrule
Coarse & 0 & 23.00 & 24.78 & 22.19 & 23.23 \\
\cmidrule(lr){1-6}
Tile~\cite{akyon2022sahi} & 0.5 & 23.71 & 25.67 & 22.91 & 24.13 \\
\rowcolor{ntMethod}
\textbf{\method{}} & 0.5 & \textbf{25.16} & \textbf{28.00} & \textbf{23.48} & \textbf{25.38} \\
\cmidrule(lr){1-6}
Tile~\cite{akyon2022sahi} & 1 & 24.36 & 26.64 & 23.43 & 24.88 \\
\rowcolor{ntMethod}
\textbf{\method{}} & 1 & \textbf{25.67} & \textbf{28.61} & \textbf{23.95} & \textbf{25.91} \\
\cmidrule(lr){1-6}
Tile~\cite{akyon2022sahi} & 2 & 25.29 & 27.80 & 23.60 & 25.28 \\
SAHI+\cgs{}+\sba{} & 2 & \textbf{26.25} & \textbf{29.07} & 24.07 & 25.92 \\
Dyn.\ Tiling+\cgs{}+\sba{} & 2 & 25.80 & 28.77 & \textbf{24.37} & \textbf{26.32} \\
DAHI+\cgs{}+\sba{} & 2 & 26.00 & 29.02 & 24.26 & 26.22 \\
\rowcolor{ntMethod}
\textbf{\method{}} & 2 & 25.76 & 28.77 & 24.07 & 26.04 \\
\bottomrule
\end{tabularx}
\end{nicetable}
\end{minipage}

\end{table*}

\smallskip\noindent\textbf{Official Detection Capacities and Held-Out Splits.}
To verify that gains are not an artifact of the uniform $K{=}100$ cap or validation-split tuning, we evaluate Coarse, Tile, external proposer pools augmented with \cgs{}+\sba{}, and \method{} at official high-density capacity ($K{=}500$) and on held-out test splits without parameter retuning. The results are reported in Table~\ref{tab:official-limits}. Across all $17$ detectors on VisDrone-DET at $K{=}500$, \method{} widens its mean lead over Tile to $+1.00$ AP$_{500}$ at $B{=}1$ and $+0.72$ AP$_{500}$ at $B{=}2$, winning all $34$ of $34$ cells. On CrowdHuman in Table~\ref{tab:official-limits}a, \method{} achieves the highest AP and lowest $\mathrm{MR}^{-2}$ across budgets at both $K{=}100$ and $K{=}500$. On AI-TOD-v2 validation and held-out test in Table~\ref{tab:official-limits}b, \method{} beats Tile across $B\in\{0.5,1,2\}$, while adding \cgs{}+\sba{} lifts every external pool above Tile at $B{=}2$ ($26.25/29.07$ validation AP for SAHI+\cgs{}+\sba{} and $24.37/26.32$ test AP for Dynamic Tiling+\cgs{}+\sba{}). Frozen parameters likewise transfer to held-out VisDrone-DET test-dev with $+0.92$ and $+0.70$ AP@100 gains over Tile at $B{=}1,2$.

\begin{table}[!ht]
\centering\scriptsize\setlength{\tabcolsep}{2.0pt}
\caption{\textbf{One-Shot Open-Loop \scp{} vs.\ Sequential Closed-Loop Replanning (RF-DETR-S).} Validation AP@100$\uparrow$, paired bootstrap $\Delta$AP (mean$\pm$SD), and $B{=}1$ window overlap against the one-shot schedule.}
\label{tab:crop-feedback}
\begingroup
\newcommand{\bsd}[1]{{\textcolor{black!55}{\fontsize{5.5}{6.5}\selectfont$\pm$#1}}}
\begin{nicetable}
\setlength{\tabcolsep}{2.5pt}
\setlength{\extrarowheight}{0.45pt}
\begin{tabularx}{\columnwidth}{llYYc}
\toprule
\rowcolor{ntHeader}
 & & \multicolumn{2}{c}{AP@100$\uparrow$ ($\Delta$AP)} & \\
\tintrule\cmidrule(lr){3-4}
\rowcolor{ntHeader}
Benchmark & Policy & $B{=}1$ & $B{=}2$ & \makecell[c]{Overlap\\(\%)} \\
\midrule
\multirow{3}{*}{COCO} & \cellcolor{ntMethod}\textbf{One-shot} & \cellcolor{ntMethod}\textbf{55.73} & \cellcolor{ntMethod}56.24 & \cellcolor{ntMethod}100.0 \\
 & Confirm. & \textbf{55.73} {\scriptsize(+0.01\bsd{0.03})} & \textbf{56.31} {\scriptsize(+0.07\bsd{0.04})} & 86.3 \\
 & Expand & \textbf{55.73} {\scriptsize(+0.01\bsd{0.03})} & 56.27 {\scriptsize(+0.03\bsd{0.05})} & 83.6 \\
\midrule
\multirow{3}{*}{CrowdHuman} & \cellcolor{ntMethod}\textbf{One-shot} & \cellcolor{ntMethod}51.13 & \cellcolor{ntMethod}51.83 & \cellcolor{ntMethod}100.0 \\
 & Confirm. & 51.20 {\scriptsize(+0.08\bsd{0.03})} & \textbf{51.84} {\scriptsize(+0.01\bsd{0.03})} & 69.9 \\
 & Expand & \textbf{51.21} {\scriptsize(+0.09\bsd{0.03})} & \textbf{51.84} {\scriptsize(+0.01\bsd{0.03})} & 69.5 \\
\midrule
\multirow{3}{*}{VisDrone-DET} & \cellcolor{ntMethod}\textbf{One-shot} & \cellcolor{ntMethod}29.46 & \cellcolor{ntMethod}30.84 & \cellcolor{ntMethod}100.0 \\
 & Confirm. & 29.52 {\scriptsize(+0.06\bsd{0.06})} & \textbf{31.00} {\scriptsize(+0.13\bsd{0.06})} & 93.4 \\
 & Expand & \textbf{29.54} {\scriptsize(+0.08\bsd{0.06})} & 30.98 {\scriptsize(+0.13\bsd{0.06})} & 93.2 \\
\midrule
\multirow{3}{*}{AI-TOD-v2} & \cellcolor{ntMethod}\textbf{One-shot} & \cellcolor{ntMethod}25.67 & \cellcolor{ntMethod}\textbf{25.76} & \cellcolor{ntMethod}100.0 \\
 & Confirm. & \textbf{25.78} {\scriptsize(+0.12\bsd{0.11})} & 25.65 {\scriptsize($-$0.21\bsd{0.36})} & 66.8 \\
 & Expand & 25.69 {\scriptsize(+0.03\bsd{0.11})} & 25.43 {\scriptsize($-$0.37\bsd{0.31})} & 66.5 \\
\bottomrule
\end{tabularx}
\end{nicetable}
\endgroup

\end{table}

\smallskip\noindent\textbf{Open-Loop versus Sequential Closed-Loop Planning.}
To test whether post-crop feedback improves over parallelizable one-shot planning, we compare open-loop \scp{} against two sequential closed-loop policies that re-plan after each crop execution: \emph{Confirmation}, which updates coarse residuals from observed crop matches, and \emph{Expand}, which also inserts newly discovered crop hypotheses. The results are reported in Table~\ref{tab:crop-feedback}. Compared with one-shot \scp{}, sequential replanning changes AP@100 by at most $+0.09$ points on COCO and CrowdHuman and $+0.13$ points on VisDrone-DET while trailing by $0.21$ to $0.37$ points on AI-TOD-v2 at $B{=}2$, with $66.5\%$ to $93.4\%$ of selected windows overlapping the one-shot plan. Deterministic geometric closure $q_{cj}$ thus captures virtually all sequential value while retaining parallel GPU batching.

\subsection{Native Pipelines and Deployment Efficiency}
\label{app:latency}

\begin{table}[!ht]
\centering\footnotesize\setlength{\tabcolsep}{3pt}
\caption{\textbf{Scale-Stratified Breakdown ($\mathrm{AP}_S$, $\mathrm{AP}_M$, $\mathrm{AP}_L$) on Frozen RF-DETR-S.} Native slicing raises small-object recall yet degrades $\mathrm{AP}_M/\mathrm{AP}_L$ via border-fragment overwriting, whereas call-matched \method{} ($B{=}4,10$) boosts $\mathrm{AP}_S$ while preserving medium and large objects.}
\label{tab:native-scale-failure}
\begin{nicetable}
\begin{tabularx}{\columnwidth}{lcYYY}
\toprule
\rowcolor{ntHeader}
Pipeline & Calls & AP$_S\uparrow$ & AP$_M\uparrow$ & AP$_L\uparrow$ \\
\midrule
\multicolumn{5}{l}{\emph{COCO val2017}} \\
Coarse & 1.0 & 31.61 & 58.04 & \textbf{73.00} \\
SAHI~\cite{akyon2022sahi} & 5.0 & 30.93 & 48.65 & 56.22 \\
DAHI~\cite{suarez2026dahi} & 10.6 & 32.58 & 36.52 & 38.92 \\
Dynamic Tiling~\cite{nguyen2023dynamic} (full) & 11.0 & 38.34 & 51.28 & 45.13 \\
Dynamic Tiling~\cite{nguyen2023dynamic} (tiles) & 10.0 & 39.36 & 41.78 & 10.54 \\
\rowcolor{ntMethod}
\textbf{\method{} ($B{=}4$)} & 5.0 & 42.03 & \textbf{58.15} & 72.55 \\
\rowcolor{ntMethod}
\textbf{\method{} ($B{=}10$)} & 11.0 & \textbf{42.14} & 58.09 & 72.10 \\
\midrule
\multicolumn{5}{l}{\emph{CrowdHuman validation}} \\
Coarse & 1.0 & 27.76 & 42.76 & 50.66 \\
SAHI~\cite{akyon2022sahi} & 5.0 & 25.87 & 23.38 & 19.08 \\
DAHI~\cite{suarez2026dahi} & 9.4 & 31.85 & 28.42 & 24.56 \\
Dynamic Tiling~\cite{nguyen2023dynamic} (full) & 11.4 & 43.99 & 47.77 & 35.25 \\
Dynamic Tiling~\cite{nguyen2023dynamic} (tiles) & 10.4 & 43.92 & 43.17 & 19.11 \\
\rowcolor{ntMethod}
\textbf{\method{} ($B{=}4$)} & 5.0 & \textbf{47.15} & \textbf{53.63} & \textbf{52.63} \\
\rowcolor{ntMethod}
\textbf{\method{} ($B{=}10$)} & 11.0 & 47.02 & 53.60 & 52.61 \\
\bottomrule
\end{tabularx}
\end{nicetable}

\end{table}

\begin{table*}[!t]
\caption{\textbf{GPU and CPU Deployment Profiling.} \textbf{(a)}~VisDrone-DET GPU execution under sequential (\emph{Static}) vs.\ $M{=}16$ dynamic crop batching (\emph{Batched}). \textbf{(b)}~Single-crop TensorRT FP16 stage share and full-split single-thread CPU planning latency.}
\label{tab:efficiency-profiling}
\centering\footnotesize
\begin{minipage}[t]{0.488\textwidth}
  \vspace{0pt}\centering\setlength{\tabcolsep}{3pt}
  \begin{nicetable}
\begin{tabularx}{\linewidth}{lYYY}
\toprule
\rowcolor{ntHeader}
\multicolumn{4}{l}{\textbf{(a) VisDrone-DET Dynamic Crop Batching (Mean / p95 ms)}} \\
\tintrule\cmidrule(lr){1-4}
\rowcolor{ntHeader}
Pipeline (Budget) & Static$\downarrow$ & Batched ($M{=}16$)$\downarrow$ & Speedup$\uparrow$ \\
\midrule
Tile ($B{=}1$) & 3.17 / 3.29 & 1.86 / 1.91 & $1.71\times$ \\
\rowcolor{ntMethod}
\textbf{\method{} ($B{=}1$)} & 3.17 / 4.75 & 1.86 / 2.41 & $1.70\times$ \\
Tile ($B{=}2$) & 4.74 / 4.77 & 2.70 / 2.74 & $1.76\times$ \\
\rowcolor{ntMethod}
\textbf{\method{} ($B{=}2$)} & 4.75 / 6.47 & 2.71 / 3.39 & $1.75\times$ \\
\rowcolor{ntMethod}
\textbf{\method{} ($B{=}2$, cap $4$)} & 4.74 / 6.46 & 2.71 / 3.37 & $1.75\times$ \\
SAHI $2{\times}2$ ($B{=}4$) & 7.90 / 7.99 & 4.38 / 4.44 & $1.80\times$ \\
\bottomrule
\end{tabularx}
\end{nicetable}

\end{minipage}\hfill
\begin{minipage}[t]{0.488\textwidth}
  \vspace{0pt}\centering\setlength{\tabcolsep}{3pt}
  \begin{nicetable}
\setlength{\tabcolsep}{2.2pt}
\setlength{\extrarowheight}{2.2pt}
\begin{tabularx}{\linewidth}{>{\raggedright\arraybackslash}Xcc}
\toprule
\rowcolor{ntHeader}
\multicolumn{3}{l}{\textbf{(b) Single-Crop GPU \& CPU Planning Breakdown}} \\
\tintrule\cmidrule(lr){1-3}
\rowcolor{ntHeader}
Execution Stage & Mean (ms)$\downarrow$ & Share / Scope \\
\midrule
Crop slice, resize, normalize & 0.04 & $2.5\%$ of crop pass \\
TensorRT FP16 model pass & 1.33 & $87.6\%$ of crop pass \\
Post-decode \& D2H transfer & 0.15 & $9.9\%$ of crop pass \\
\midrule
\scp{} CPU plan (COCO, $B{=}2$) & 0.60 & Full split, 1 thread \\
\scp{} CPU plan (VisDrone-DET, $B{=}2$) & 1.71 & Full split, 1 thread \\
\bottomrule
\end{tabularx}
\end{nicetable}

\end{minipage}
\end{table*}

\noindent\textbf{Scale Diagnostics Against Complete Native Pipelines.}
At matched total detector calls in Table~\ref{tab:native-main}, \method{} outperforms native uniform and adaptive slicing in SAHI~\cite{akyon2022sahi}, ASAHI~\cite{zhang2023adaptive}, DAHI~\cite{suarez2026dahi}, DAASD~\cite{liu2026daasd}, Dynamic Tiling~\cite{nguyen2023dynamic}, ROI-Gated SAHI~\cite{riyadh2026roi}, and ViCrop-Det~\cite{wang2026vicrop}, as well as learned cluster proposing in ClusDet~\cite{yang2019clustered}. To isolate the underlying scale behavior, we compare Coarse, native SAHI, native DAHI, full and tile-only Dynamic Tiling, and call-matched \method{} ($B{=}4,10$) across $\mathrm{AP}_S$, $\mathrm{AP}_M$, and $\mathrm{AP}_L$ on frozen RF-DETR-S. The results are reported in Table~\ref{tab:native-scale-failure}. While native slicing lifts $\mathrm{AP}_S$, unguarded border fragments overwrite intact coarse predictions and collapse $\mathrm{AP}_L$ from $73.00$ to $10.54$--$56.22$ on COCO and from $50.66$ to $19.08$--$35.25$ on CrowdHuman. By contrast, \method{} lifts $\mathrm{AP}_S$ to $42.03$ and $47.15$ at $B{=}4$ while preserving $\mathrm{AP}_M$ and $\mathrm{AP}_L$ via shared $\mathcal P_c$ admission and scale gates.

\smallskip\noindent\textbf{Dynamic Crop Batching and CPU Planning Overhead.}
To quantify GPU runtime gains from open-loop scheduling, we compare sequential single-crop execution (\emph{Static}) against $M{=}16$ dynamic crop batching (\emph{Batched}) on VisDrone-DET and profile TensorRT FP16 GPU stages alongside single-thread Numba CPU planning. The results are reported in Table~\ref{tab:efficiency-profiling}. In Table~\ref{tab:efficiency-profiling}a, $M{=}16$ dynamic batching accelerates \method{} by $1.70{\times}$ at $B{=}1$ and $1.75{\times}$ at $B{=}2$, cutting $B{=}2$ p95 GPU latency from $6.47$ to $3.39$\,ms. In Table~\ref{tab:efficiency-profiling}b, TensorRT model inference accounts for $87.6\%$ of single-crop runtime ($0.04$\,ms for slicing and resizing), while single-thread CPU planning adds only $0.60$\,ms per image on COCO and $1.71$\,ms per image on VisDrone-DET.

\subsection{Causal Online Streaming Without Test-Split Statistics}
\label{app:online-effective}

\begin{figure*}[!t]
\centering
\includegraphics[width=\textwidth]{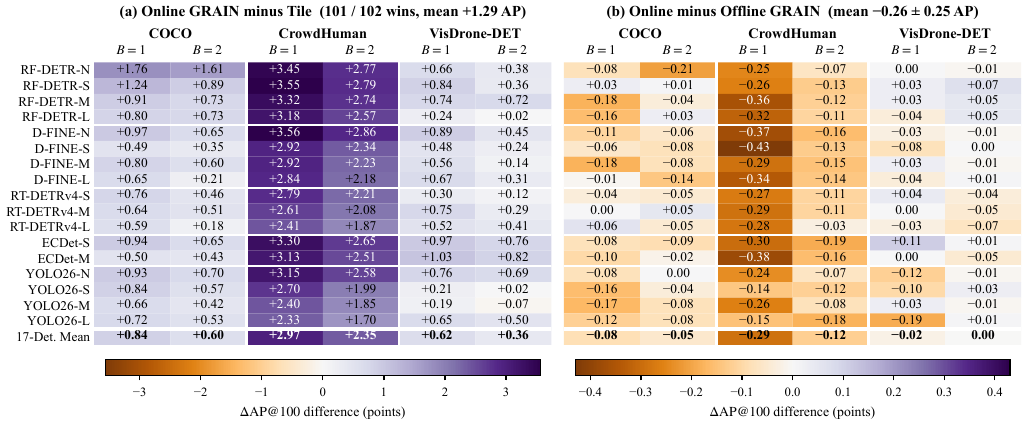}
\caption{\textbf{Causal Online Streaming Across All $17$ Frozen Detectors.} Validation AP@100 differences at $B\in\{1,2\}$ using causal quantile $\lambda_t$ and frozen training class prior $N_k^{\mathrm{train}}$, comparing \emph{Online minus Tile} on the left ($101/102$ wins) and \emph{Online minus Offline \method{}} on the right (mean $-0.26\pm 0.25$ AP).}
\label{fig:online-deltas}
\end{figure*}

\begin{table*}[!t]
\centering\footnotesize\setlength{\tabcolsep}{3.8pt}
\caption{\textbf{Zero-Test-Statistic Deployment Across Budget Controllers and Squeeze Priors.} \textbf{(a)}~Orthogonal ablation on frozen RF-DETR-S decoupling budget allocation from squeeze class priors. \textbf{(b)}~Win counts over call-matched Tile and paired $\Delta$AP vs.\ Offline \method{} across all $102$ detector-dataset-budget cells.}
\label{tab:zero-test-stats}
\begin{nicetable}
\setlength{\extrarowheight}{0.45pt}
\begin{tabularx}{\textwidth}{lllYYYYYY}
\toprule
\rowcolor{ntHeader}
\multicolumn{9}{l}{\textbf{(a) RF-DETR-S Orthogonal Decoupling of Budget Control and Squeeze Priors (AP@100$\uparrow$)}} \\
\tintrule\cmidrule(lr){1-9}
\rowcolor{ntHeader}
 & & & \multicolumn{2}{c}{COCO} & \multicolumn{2}{c}{CrowdHuman} & \multicolumn{2}{c}{VisDrone-DET} \\
\tintrule\cmidrule(lr){4-5}\cmidrule(lr){6-7}\cmidrule(lr){8-9}
\rowcolor{ntHeader}
Deployment Regime & Budget Controller & Squeeze Prior & $B{=}1$ & $B{=}2$ & $B{=}1$ & $B{=}2$ & $B{=}1$ & $B{=}2$ \\
\midrule
Offline reference & Global \sba{} ($M{=}N$) & Test collection $M_k^{\mathrm{test}}$ & 55.73 & 56.24 & \textbf{51.13} & \textbf{51.83} & 29.46 & 30.84 \\
Causal (prefix squeeze) & Causal quantile $\lambda_t$ & Stream prefix $M_k^{\le t}$ & 55.75 & 56.19 & 50.87 & 51.65 & \textbf{29.47} & \textbf{30.90} \\
\midrule
\rowcolor{ntMethod}
\textbf{Micro-batch (\method{})} & Block \sba{} ($M{=}16$) & Frozen train $M_k^{\mathrm{train}}$ & \textbf{55.77} & \textbf{56.25} & 51.02 & 51.70 & 29.31 & 30.61 \\
\rowcolor{ntMethod}
\textbf{Causal stream (\method{})} & Causal quantile $\lambda_t$ & Frozen train $M_k^{\mathrm{train}}$ & 55.75 & 56.19 & 50.87 & 51.65 & 29.29 & 30.68 \\
\midrule
Single-image quota & Per-image cap ($M{=}1$) & Frozen train $M_k^{\mathrm{train}}$ & 54.95 & 55.84 & 48.56 & 50.24 & 29.00 & 30.48 \\
Single-image Tile & Uniform $2{\times}2$ grid & Frozen train $M_k^{\mathrm{train}}$ & 54.49 & 55.33 & 47.32 & 48.91 & 28.46 & 30.26 \\
\bottomrule
\end{tabularx}
\vspace{3pt}
\begin{tabularx}{\textwidth}{lYY}
\toprule
\rowcolor{ntHeader}
\multicolumn{3}{l}{\textbf{(b) Causal Online Summary Across $17$ Detectors at $B\in\{1,2\}$ ($102$ Budget Cells)}} \\
\tintrule\cmidrule(lr){1-3}
\rowcolor{ntHeader}
Evaluation Scope & Online Wins vs.\ Call-Matched Tile$\uparrow$ & Online $-$ Offline \method{} $\Delta$AP (Mean$\pm$SD) \\
\midrule
\rowcolor{ntMethod}
\textbf{All settings ($102$ cells)} & \textbf{101 / 102} & $-0.26 \pm 0.25$ \\
COCO ($34$ cells) & 34 / 34 & $-0.12 \pm 0.08$ \\
CrowdHuman ($34$ cells) & 34 / 34 & $-0.20 \pm 0.10$ \\
VisDrone-DET ($34$ cells) & 33 / 34 & $-0.47 \pm 0.33$ \\
\bottomrule
\end{tabularx}
\end{nicetable}

\end{table*}

To evaluate strictly causal deployment on live streams without future frames or test-split class masses $M_k^{\mathrm{test}}$, we compare six deployment regimes decoupling budget control from squeeze class priors: Offline \sba{} with $M_k^{\mathrm{test}}$, causal quantile $\lambda_t$ under the sliding-window constants in Table~\ref{tab:constants} paired with stream-prefix $M_k^{\le t}$ or frozen training priors $M_k^{\mathrm{train}}$, $M{=}16$ Block~\sba{} with $M_k^{\mathrm{train}}$, single-image quotas ($M{=}1$), and single-image Tile. The results are reported in Figure~\ref{fig:online-deltas} and Table~\ref{tab:zero-test-stats}. Across all $102$ detector-dataset-budget cells in Figure~\ref{fig:online-deltas} and Table~\ref{tab:zero-test-stats}b, causal online \method{} utilizes $99.7\%$ of the call budget and beats call-matched Tile in $101$ of $102$ cells by $+1.29$ AP@100 on average, trailing Offline \method{} by only $-0.26\pm0.25$ AP@100. In Table~\ref{tab:zero-test-stats}a, frozen training priors match test-collection squeezing within $0.23$ AP under $M{=}16$ Block~\sba{} and $0.26$ AP under causal streaming, while $M{=}16$ Block~\sba{} adapts faster under abrupt scene shifts whenever a $16$-frame buffer is permissible.

\section{GRAIN Algorithm and Protocol Constants}
\label{app:implementation}

Algorithm~\ref{alg:grain-unified} unifies the collection-wide Offline solver and the zero-test-statistic causal Online controller under one frozen-detector interface. Given coarse predictions filtered at confidence floor $s_{\min}$, Line~3 computes scale-calibrated resolution-deficit weights $u_j$ and initializes multiplicative residuals $R_j\gets 1$ without accessing internal backbone features. Lines~4--6 execute additive-separable global max-heap scheduling in Offline mode, refreshing only the popped image $i^\star$ after updating $R_j$ so all accepted crop windows are finalized prior to parallel GPU batching. Lines~7--10 run causal Online streaming within the prefix allowance $\lfloor Bt\rfloor$ by tracking a sliding-window quantile threshold $\lambda_t$ over recent marginal gains. Lines~11--14 execute the frozen detector on accepted crops, project boxes back to native image coordinates, enforce interior admission $\mathcal P_c$ and scale gates, and apply order-preserving class-mass squeezing in Appendix~\ref{app:fusion-order} to return per-image top-$K$ detections. Because both Offline heap pops and Online threshold checks evaluate only coarse hypotheses and scalar residual arrays, planning requires zero backbone gradients, auxiliary heads, or test-split ground truth. When a live stream starts from cold initialization prior to warmup, the Online controller falls back to deterministic per-image top-gain allocation within the prefix allowance $\lfloor Bt\rfloor$ and enforces the per-frame cap so early arrivals incur no unbudgeted latency spikes. Table~\ref{tab:constants} serves as the single reference for all frozen Offline and Online protocol constants across the paper.

\begin{table}[!tp]
\caption{\textbf{Frozen Planner, Streaming, and Fusion Constants.} Canonical parameter values shared across all $17$ detectors and four benchmarks, where only domain scale $\tau$ varies by benchmark as analyzed in Appendix~\ref{app:tau}.}
\label{tab:constants}
\centering\scriptsize\setlength{\tabcolsep}{3.0pt}
\begin{nicetable}
\begin{tabularx}{\columnwidth}{l>{\raggedright\arraybackslash}X}
\toprule
\rowcolor{ntHeader}
Parameter / Module & Frozen Protocol Value and Rule \\
\midrule
\multicolumn{2}{l}{\emph{Stage I: Planning (\scp{}) and Budget Controllers}} \\
Target scale $\tau$ & $24$, $32$, $12$, and $12$\,px on COCO, CrowdHuman, VisDrone-DET, and AI-TOD-v2 \\
Zoom lattice $\mathcal Z, \delta_{x,y}$ & $2^{k/4}$ for $k{=}1,\ldots,16$ with $\tfrac18$-window origin step, border-clipped \\
Coarse floor $s_{\min}$ & $s_{\min}=0.05$, with planner saturation cutoff $z_c a_j<4\tau$ \\
Offline / online budget & $L=\lfloor BN\rfloor$ with cap $H_{\max}$, and online window $256$, warmup $20$, horizon $50$, cap $4B$ \\
\midrule
\multicolumn{2}{l}{\emph{Stage II: Fusion (\prf{}) and Tiling Baseline}} \\
Interior margin $\delta_{\mathrm{img}}$ & $\delta_{\mathrm{img}}=2$\,px inward contraction on $\mathcal P_c$, with outer image borders exempt \\
Scale gates $4\tau, 2\tau, \gamma$ & Crop gate $a_{\mathrm{g}}<4\tau$, and covered coarse attenuation $\gamma=0.5$ on $a_{\mathrm{g}}<2\tau$ \\
Pool \& squeeze $K, \epsilon_M$ & Class NMS IoU $0.6$ to $2K{=}200$, and $M_k^{-1/2}$ squeeze to $K{=}100$ with $\epsilon_M{=}10^{-6}$ \\
Tile comparator & Per-image $2{\times}2$ grid with $20\%$ spatial overlap \\
\bottomrule
\end{tabularx}
\end{nicetable}

\end{table}

\begin{table*}[!t]
\caption{\textbf{Summary of Core Notation.} Mathematical symbols grouped by modeling stage, with physical units or domains and defining references.}
\label{tab:app-symbols}
\centering\small
\begin{nicetable}
\setlength{\extrarowheight}{1.65pt}
\setlength{\tabcolsep}{4pt}
\begin{tabularx}{\textwidth}{l>{\raggedright\arraybackslash}Xcc}
\toprule
\rowcolor{ntHeader}
\textbf{Symbol} & \textbf{Definition} & \textbf{Unit / Domain} & \textbf{Reference} \\
\midrule
\multicolumn{4}{l}{\emph{1.\ Collection Geometry, Budgets, and Resolution-Deficit Surrogate}} \\
$N,\; B,\; L$ & Image count, nominal per-image crop budget, and integer collection call allowance $\lfloor BN\rfloor$. & count, calls/img, calls & Section~\ref{sec:method} \\
$W_i, H_i,\; W_D, H_D$ & Native resolution of image $i$ and fixed input canvas of the frozen detector $D$. & native / input px & Appendix~\ref{app:symbols} \\
$b_j,\; s_j,\; a_j,\; \tau$ & Coarse box, confidence score, effective input side length, and domain half-saturation scale. & box, $[0,1]$, input px & Eq.~\eqref{eq:app-input-scale} \\
$\phi(s),\; r(a),\; u_j$ & Confidence salience $\sqrt{s}(1{-}s/2)$, Hill support $a^2/(a^2{+}\tau^2)$, and deficit weight $\phi(s_j)[1{-}r(a_j)]$. & dimensionless & Eq.~\eqref{eq:app-weight-expanded} \\
$\delta_{\mathrm{img}},\; \mathcal P_c$ & Inward internal-edge margin and contracted interior positive region of crop window $c$. & native px, region & Eq.~\eqref{eq:app-positive-region} \\
$z_c,\; \widehat z_c,\; \pi_{cj},\; q_{cj}$ & Nominal zoom, integer-rounded zoom, containment $\mathbf{1}[b_j\subseteq\mathcal P_c]$, and closure quality $\pi_{cj}h(a_j,z_c)$. & ratios, $[0,1]$ & Eqs.~\eqref{eq:app-closure}, \eqref{eq:app-rounded-zoom} \\
\midrule
\multicolumn{4}{l}{\emph{2.\ Multi-Zoom Lattice Sweep (\lws{}) and Shared-Budget Allocation (\sba{})}} \\
$S_i,\; S,\; \mathcal C_i$ & Selected crop plan on image $i$, collection plan $\bigcup_i S_i$, and image-$i$ candidate window pool. & window sets & Appendix~\ref{app:theory} \\
$R_j(S),\; F(S),\; \Delta(c\mid S)$ & Multiplicative residual $\prod_{c\in S}(1{-}q_{cj})$, collection coverage $\sum_j u_j[1{-}R_j(S)]$, and marginal gain. & $[0,1]$, closure & Eqs.~\eqref{eq:app-remaining}--\eqref{eq:app-marginal-gain} \\
$v_j(z\mid S),\; G_z$ & Hypothesis $j$'s remaining zoom-$z$ credit and discrete lattice origin count $N_x N_y$. & closure, count & Appendix~\ref{app:algorithmic-identities} \\
$D^{(z)},\; G^{(z)},\; T_{\mathrm{plan}}$ & Dual-origin 2D difference array, prefix-summed lattice gain grid, and total CPU planning time. & closure, complexity & Eqs.~\eqref{eq:app-diff-impulses}--\eqref{eq:app-total-complexity} \\
$H_{\max},\; \lambda_L$ & Optional per-image call cap and offline global marginal-gain threshold. & calls, closure & Appendices~\ref{app:offline-threshold}, \ref{sec:implementation-gap} \\
\midrule
\multicolumn{4}{l}{\emph{3.\ Submodular Optimization Guarantees and AP-Coupling Theory}} \\
$\epsilon_t,\; E_{\mathrm{LWS}}$ & Step-$t$ discrete lattice gain shortfall and discounted cumulative search error $\sum_t(1{-}1/L)^{L-t}\epsilon_t$. & weighted closure & Proposition~\ref{prop:approx-greedy} \\
$\Psi,\; \Phi_\Delta,\; \bar\zeta$ & Increasing rank-alignment link, concave AP-transfer link, and uniform residual bound. & functions, AP pts & Propositions~\ref{prop:rank-alignment}--\ref{prop:ap-transfer} \\
$\mathcal L_{\mathrm{pos}},\; \mathcal E_{\mathrm{loss}}$ & Signed COCO AP channels for newly matched targets and evicted coarse matches under cap $K$. & AP pts & Eq.~\eqref{eq:ap-channel-defs} \\
$\mathcal D_{\mathrm{rank}},\; \Delta\mathcal I_{\mathrm{env}}$ & Signed COCO AP channels for retained-match precision rank shifts and 101-point envelope delta. & AP pts & Eq.~\eqref{eq:ap-channel-defs} \\
\midrule
\multicolumn{4}{l}{\emph{4.\ Positive-Region Fusion (\prf{}) and Causal Online Streaming}} \\
$K,\; \gamma,\; \mathcal D_{2K}^{(i)}$ & Output evaluation cap, covered-coarse attenuation multiplier, and expanded per-image NMS pool. & count, ratio, set & Section~\ref{sec:method}, Table~\ref{tab:constants} \\
$M_k,\; v_d$ & Fused class-$k$ score mass and order-preserving class-balanced ranking key. & score, key & Eq.~\eqref{eq:app-squeeze-key} \\
$M_k^{\mathrm{train}},\; N_k^{\mathrm{train}}$ & Frozen training-split coarse score mass and training instance-count class priors. & score, count & Appendix~\ref{app:online-effective} \\
$t,\; Q,\; \lambda_t$ & Stream arrival index, cumulative spent crop calls, and causal sliding-window quantile threshold. & index, calls, closure & Algorithm~\ref{alg:grain-unified} \\
\bottomrule
\end{tabularx}
\end{nicetable}
\end{table*}

\begin{figure}[!tp]
\begin{grainalg}{\method{} Inference (Offline and Online Modes)}{alg:grain-unified}
\item \textbf{Inputs:} Frozen detector $D$, domain scale $\tau$, budget $B$, cap $H_{\max}$, output limit $K$, mode $\in\{\mathrm{Offline},\mathrm{Online}\}$, Online training prior $M_k^{\mathrm{train}}$, and frozen constants from Table~\ref{tab:constants}.
\item \textbf{Output:} Per-image top-$K$ fused detections $\mathcal D_K^{(i)}$ for all $i$.
\item Run coarse pass $D(I_i)$ on all frames in Offline mode or arrival $I_t$ in Online mode, filter $s_j\ge s_{\min}$, compute $u_j$ via Eq.~\eqref{eq:app-weight-expanded}, and set $R_j\gets 1$.
\item \textbf{if} $\mathrm{mode}=\mathrm{Offline}$ \textbf{then}
\item \algind $S\gets\varnothing$, and \textbf{for} $i=1,\ldots,N$ \textbf{do} score $(c_i^\star, g_i^\star)$ via 2D sweep in Eq.~\eqref{eq:app-prefix-sum}, update $\widehat z_{c_i^\star}$ via Eq.~\eqref{eq:app-rounded-zoom}, and push to max-heap. \textbf{end for}
\item \algind \textbf{while} $|S|<\lfloor BN\rfloor$ and heap non-empty \textbf{do} pop $(i^\star, c)$, update $S\gets S\cup\{c\}$ and $R_j\gets R_j(1-q_{cj})$, and refresh $i^\star$ if $|S_{i^\star}|<H_{\max}$. \textbf{end while}
\item \textbf{else} \hfill{\scriptsize$\triangleright$ \textit{Causal Online streaming}}
\item \algind Update sliding-window gain buffer and causal quantile threshold $\lambda_t$ using the window rules in Table~\ref{tab:constants}.
\item \algind \textbf{for} each arrival $I_t$ \textbf{do} sweep local greedy list and accept longest prefix with gain $\ge\lambda_t$ within remaining prefix budget $\lfloor Bt\rfloor-Q$ and per-frame caps. \textbf{end for}
\item \textbf{end if}
\item Execute frozen $D$ on accepted crops $c\in S$ and project boxes to native coordinates as detailed in Appendix~\ref{app:coordinates}.
\item Enforce $\mathcal P_c$ admission via Eq.~\eqref{eq:app-positive-region}, crop scale gate, $\gamma$-attenuation of covered coarse boxes, and class-wise NMS to form $\mathcal D_{2K}^{(i)}$ following Table~\ref{tab:constants}.
\item Select top-$K$ detections via Eq.~\eqref{eq:app-squeeze-key} using collection $M_k$ in Offline mode or frozen $M_k^{\mathrm{train}}$ in Online mode, preserving fused scores $s_d$.
\item \textbf{Return} fused detection sets $\{\mathcal D_K^{(i)}\}$.
\end{grainalg}
\end{figure}

\section{Symbols Table}
\label{app:symbol-table}

Table~\ref{tab:app-symbols} summarizes the core mathematical notation, physical domains, and defining references across our resolution-deficit surrogate, multi-zoom lattice sweep, submodular optimization bounds, four-channel AP decomposition, and post-inference fusion rules. Every symbol matches the main-paper formulation in Section~\ref{sec:method} without redefinition across offline, micro-batch, and causal streaming regimes, while all fixed numerical hyperparameter values are consolidated once in Table~\ref{tab:constants}. Across the four modeling stages in Table~\ref{tab:app-symbols}, uppercase calligraphic letters denote spatial regions, candidate pools, or signed AP channels, whereas lowercase symbols denote per-hypothesis scales, closure weights, and stream indices. All set cardinalities, runtime complexities, and AP differences follow standard COCO evaluation conventions unless noted otherwise.

\section{Additional Ablation Studies}
\label{app:review-replay}
\label{app:round3}

We conduct controlled ablations across the four core design axes of \method{}: component interactions and fusion rules under matched views, collection demand skew and extended budgets, domain target-scale alignment and hyperparameter stability, and per-image latency caps with streaming block sizes.

\subsection{Module Interactions and Fixed-View Fusion Controls}
\label{app:factorial-controls}

\begin{figure}[!t]
\centering
\includegraphics[width=\columnwidth]{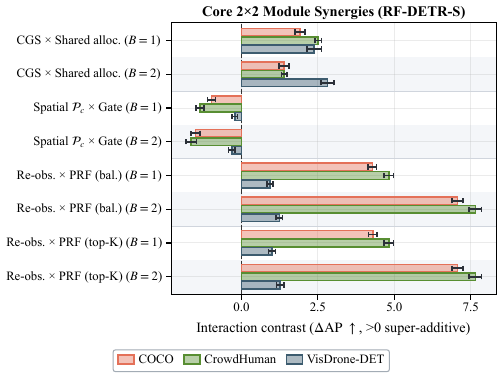}
\caption{\textbf{Factorial $2{\times}2$ Interaction Contrasts on RF-DETR-S.} Paired image-bootstrap mean$\pm$SD bars at $B\in\{1,2\}$. Positive bars to the right of the dashed zero line indicate super-additive module synergy.}
\label{fig:factorial-interactions}
\end{figure}

\noindent\textbf{Scoring-by-Allocation Synergy and Equal-Call Controls.}
To test whether planning and fusion modules act super-additively and whether shared allocation still leads when matched to the exact realized call count of per-image quotas, we evaluate paired $2{\times}2$ interaction contrasts $I=(A_{1,1}-A_{0,1})-(A_{1,0}-A_{0,0})$ on frozen RF-DETR-S across four factor pairs: \cgs{} crossed with shared allocation, spatial $\mathcal P_c$ planning crossed with fusion gates, and re-observation crossed with \prf{} under class-balanced or score top-$K$ selection. The results are reported in Figure~\ref{fig:factorial-interactions}. First, the \cgs{}-by-sharing contrast reaches $+1.39\pm0.16$ to $+2.81\pm0.21$ AP@100 across COCO, CrowdHuman, and VisDrone-DET at $B\in\{1,2\}$, and the re-observation-by-\prf{} contrast reaches $+0.93\pm0.10$ to $+7.65\pm0.20$ AP@100, exceeding zero by more than eight bootstrap standard deviations. Second, capping shared allocation at the exact realized call count of per-image quotas still improves AP@100 by $+0.78$ and $+0.48$ on COCO, $+2.42$ and $+1.48$ on CrowdHuman, and $+0.26$ and $+0.16$ on VisDrone-DET at $B{=}1,2$. Sharing calls without multiplicative residual discounting causes the heap to re-crop the same dense cluster, dropping COCO $B{=}2$ AP to $53.43$ below the $54.42$ quota baseline, whereas \cgs{} without sharing strands surplus calls on sparse frames.

\begin{table}[!t]
\centering\footnotesize\setlength{\tabcolsep}{2.8pt}
\caption{\textbf{Continuous Residual Contraction vs.\ Hard Overlap Vetoes.} Validation AP@100$\uparrow$ on frozen RF-DETR-S at $B\in\{1,2\}$ under identical \prf{} fusion.}
\label{tab:nonresidual}
\begingroup
\newcolumntype{A}{>{\hsize=0.86\hsize\centering\arraybackslash}X}
\newcolumntype{B}{>{\hsize=1.04\hsize\centering\arraybackslash}X}
\newcolumntype{C}{>{\hsize=1.10\hsize\centering\arraybackslash}X}
\begin{nicetable}
\setlength{\tabcolsep}{1.6pt}
\begin{tabularx}{\columnwidth}{lAABBCC}
\toprule
\rowcolor{ntHeader}
 & \multicolumn{2}{c}{COCO} & \multicolumn{2}{c}{CrowdHuman} & \multicolumn{2}{c}{VisDrone-DET} \\
\tintrule\cmidrule(lr){2-3}\cmidrule(lr){4-5}\cmidrule(lr){6-7}
\rowcolor{ntHeader}
Planner Rule & $B{=}1$ & $B{=}2$ & $B{=}1$ & $B{=}2$ & $B{=}1$ & $B{=}2$ \\
\midrule
No residual ($R_j\equiv 1$) & 53.16 & 53.27 & 46.37 & 46.85 & 23.87 & 24.02 \\
Hard IoU veto ($>0.5$) & 54.10 & 54.48 & 48.45 & 49.61 & 25.33 & 26.76 \\
Hard IoS veto ($>0.5$) & 55.49 & 56.03 & 50.47 & 51.30 & 28.01 & 29.71 \\
\midrule
\rowcolor{ntMethod}
\textbf{Continuous $R_j$ (\method{})} & \textbf{55.73} & \textbf{56.24} & \textbf{51.13} & \textbf{51.82} & \textbf{29.46} & \textbf{30.84} \\
\bottomrule
\end{tabularx}
\end{nicetable}
\endgroup

\end{table}

\smallskip\noindent\textbf{Residual Discounting versus Hard Overlap Vetoes.}
To test whether multiplicative residual contraction $R_j(S)=\prod_{c\in S}(1-q_{cj})$ can be replaced by spatial exclusion heuristics under identical frozen RF-DETR-S weights and \prf{}, we compare four planner rules at $B\in\{1,2\}$: removing residual updates ($R_j\equiv 1$), rejecting candidates with $\mathrm{IoU}>0.5$ against earlier picks on the same image, rejecting candidates with intersection-over-smaller $\mathrm{IoS}>0.5$, and continuous residual contraction in \method{}. The results are reported in Table~\ref{tab:nonresidual}. Removing residual updates drops $B{=}2$ AP@100 from $30.84$ to $24.02$ on VisDrone-DET and from $51.82$ to $46.85$ on CrowdHuman. Compared with hard IoU and IoS vetoes, continuous $R_j(S)$ improves AP@100 by $+1.63$ to $+4.13$ and by $+0.21$ to $+1.45$ across all six dataset-budget cells. Hard spatial vetoes block overlapping multi-zoom windows that resolve disjoint target subsets in dense crowds, whereas $R_j(S)$ discounts only hypotheses already covered.

\begin{figure*}[!t]
\centering
\begin{subfigure}[b]{0.485\textwidth}
  \centering
  \includegraphics[width=\textwidth]{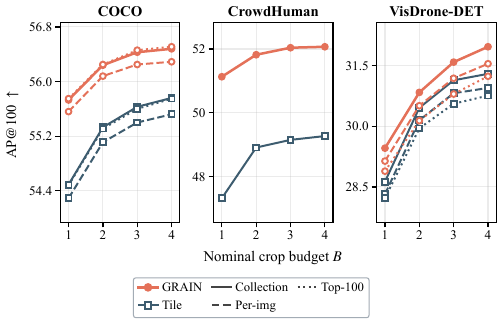}
  \caption{Extended budgets up to complete $2{\times}2$ tiling ($B\le 4$)}
  \label{fig:complete-grid}
\end{subfigure}\hfill
\begin{subfigure}[b]{0.485\textwidth}
  \centering
  \includegraphics[width=\textwidth]{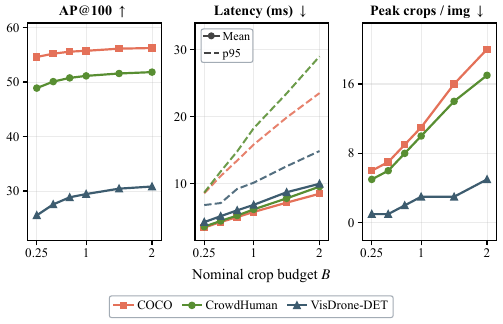}
  \caption{Sub-unit to $B{=}2$ AP, RTX~4090 latency, and peak load}
  \label{fig:lws-fractional-budget}
\end{subfigure}
\caption{\textbf{Budget Scaling From Fractional Allowances to $B{=}4$ on RF-DETR-S.} \textbf{(a)}~\method{} retains a consistent lead over Tile even at $B{=}4$ across all three output-selection rules. \textbf{(b)}~AP@100, mean/p95 GPU+CPU latency, and peak crops per image over $B\in[\tfrac14,2]$.}
\label{fig:budget-complete-fractional}
\end{figure*}

\smallskip\noindent\textbf{Fixed-View Fusion Components and External Merging Rules.}
To isolate each post-processing design choice without planner confounding, we hold $B{=}2$ RF-DETR-S crop views fixed and compare internal \prf{} ablations alongside external merging baselines including plain NMS, Soft-NMS~\cite{bodla2017softnms}, WBF~\cite{solovyev2021weighted}, SAHI~\cite{akyon2022sahi}, and AutoFocus~\cite{najibi2019autofocus}. The results are reported in Table~\ref{tab:ablation} and Figure~\ref{fig:ablation}c. Within \prf{}, disabling interior admission $\mathcal P_c$ causes the largest accuracy drop of $-4.02$, $-6.58$, and $-0.68$ AP@100 on COCO, CrowdHuman, and VisDrone-DET, disabling scale-gated coarse supersession costs $0.29$ to $0.52$ AP@100, and class-balanced squeezing adds $+0.70$ AP@100 on multi-class VisDrone-DET. Against external merging rules on the exact same views, full \prf{} leads the strongest baseline by $+0.53$ to $+0.96$ AP@100 because $\mathcal P_c$ rejects seam-bisected fragments without merging adjacent boxes.

\subsection{Budget Scaling and Complete-Grid Analysis}
\label{app:review-allocation}

\noindent\textbf{Demand Skew and Unsupervised Density Concentration.}
To verify that coarse deficit weights capture true small-object density without ground-truth supervision, we inspect per-image call allocations of frozen RF-DETR-S at $B{=}1$ across COCO, CrowdHuman, and VisDrone-DET. Across the validation splits, \sba{} skips roughly $60\%$ of sparse COCO and CrowdHuman frames and concentrates calls onto high-density scenes, yielding Spearman correlations of $\rho=0.65$ to $0.78$ with ground-truth small-object counts and doubling \method{}'s AP gain over Tile on crowded CrowdHuman frames relative to sparser scenes.

\smallskip\noindent\textbf{Complete Grids and Fractional Budgets.}
To evaluate \method{} from sub-unit budgets up to exhaustive grid coverage, we compare \method{} and uniform $2{\times}2$ Tile across collection-balanced, per-image, and score top-$100$ selection rules for $B\in\{1,2,3,4\}$, and measure accuracy, latency, and peak per-image load over $B\in[\tfrac14,2]$. The results are reported in Figure~\ref{fig:budget-complete-fractional}. In Figure~\ref{fig:complete-grid}, even at $B{=}4$ where Tile evaluates all four slices on every image, \method{} leads by $+0.72$, $+2.80$, and $+0.67$ AP@100 ($56.48$, $52.07$, and $31.97$) across all three selection rules because multi-zoom \lws{} windows resolve tiny clusters while skipping resolved frames. In Figure~\ref{fig:lws-fractional-budget}, spending only $B{=}\tfrac14$ extra calls per image already surpasses one-crop Tile ($B{=}1$) on COCO ($54.60$ vs.\ $54.49$) and CrowdHuman ($48.89$ vs.\ $47.32$).

\subsection{Target Scale, Label-Free Calibration, and Hyperparameter Sensitivity}
\label{app:tau}
\label{sec:archive-reproducibility}
\label{app:review-constants}

\begin{figure*}[!t]
\centering
\begin{subfigure}[b]{0.605\textwidth}
  \centering
  \includegraphics[width=\textwidth]{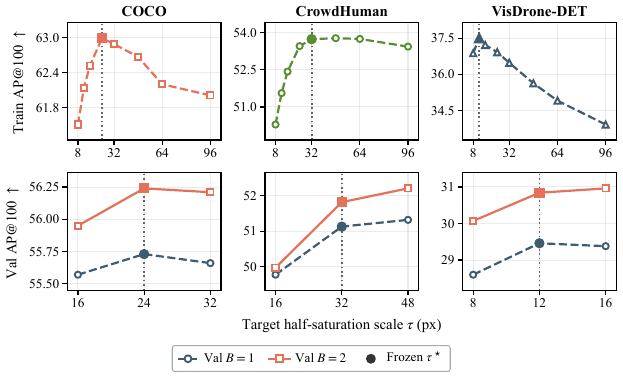}
  \caption{Training-subset and full-validation $\tau$ sweeps}
  \label{fig:target-scale}
\end{subfigure}\hfill
\begin{subfigure}[b]{0.37\textwidth}
  \centering
  \includegraphics[width=\textwidth]{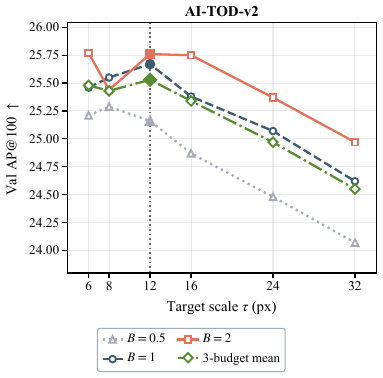}
  \caption{AI-TOD-v2 validation sweep}
  \label{fig:aitod-scale}
\end{subfigure}
\caption{\textbf{Domain Target-Scale Alignment ($\tau$).} \textbf{(a)}~RF-DETR-S AP@100 across training subsets (top) and full validation splits at $B\in\{1,2\}$ (bottom), where vertical lines and filled markers indicate frozen protocol $\tau^\star$ from Table~\ref{tab:constants}. \textbf{(b)}~AI-TOD-v2 validation sweep across budgets.}
\label{fig:tau-sweeps}
\end{figure*}

\noindent\textbf{Domain Target-Scale Alignment.}
To examine sensitivity to the sole domain parameter $\tau$ in Eq.~\eqref{eq:app-weight-expanded}, whose frozen values are listed in Table~\ref{tab:constants}, we sweep $\tau$ across training subsets and full validation splits on RF-DETR-S. The results are reported in Figure~\ref{fig:tau-sweeps}. Empirical optima align directly with characteristic object scale across aerial, multi-scale, and full-body pedestrian benchmarks, and shifting $\tau$ by one grid step changes training-subset AP@100 by at most $0.29$ to $0.62$ points, allowing a single RF-DETR-S setting per domain to transfer across all $17$ detectors.

\smallskip\noindent\textbf{Label-Free Calibration and Hyperparameter Stability.}
To test sensitivity when labeled calibration splits are unavailable or auxiliary constants in Table~\ref{tab:constants} are perturbed, we evaluate unlabeled scale estimation, post-hoc score calibration, and single-axis constant sweeps on RF-DETR-S at $B{=}2$. Setting $\tau$ to the $30$th percentile of coarse box scales over $256$ unlabeled frames matches the frozen protocol $\tau$ within $0.32$ AP@100 across all four benchmarks, while Temperature, Platt, or isotonic score calibration changes AP@100 by at most $0.03$ points. Varying interior margins $\delta_{\mathrm{img}}\in\{1,2,4\}$\,px, attenuation multipliers $\gamma\in[0.1,0.6]$, supersession gates $\eta\in[2\tau,4\tau]$, or squeeze exponents $\alpha\in[0.25,0.50]$ shifts $B{=}2$ AP@100 by only $0.05$ to $0.31$ points, whereas boundary degeneracies at $\delta_{\mathrm{img}}{=}0$ or $\gamma{=}0$ degrade precision or recall sharply.

\subsection{Per-Image Call Caps and Streaming Block Horizon}
\label{app:call-caps}

\begin{figure*}[!t]
\centering
\begin{subfigure}[b]{0.485\textwidth}
  \centering
  \includegraphics[width=\textwidth]{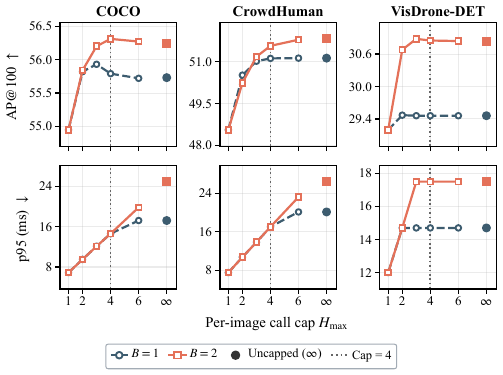}
  \caption{Per-image call caps $H_{\max}$ vs.\ modeled p95 latency}
  \label{fig:call-caps}
\end{subfigure}\hfill
\begin{subfigure}[b]{0.485\textwidth}
  \centering
  \includegraphics[width=\textwidth]{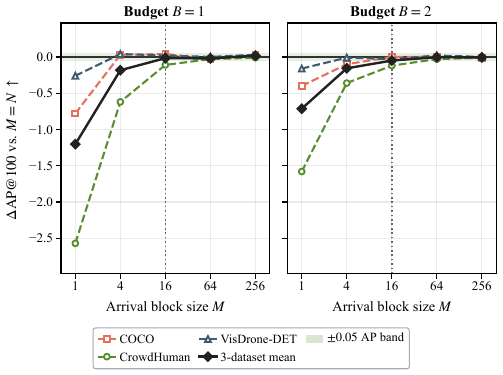}
  \caption{Micro-batch arrival horizon $M$ vs.\ offline \sba{}}
  \label{fig:sba-blocks}
\end{subfigure}
\caption{\textbf{Tail-Latency Capping and Micro-Batch Convergence on RF-DETR-S.} \textbf{(a)}~$H_{\max}{=}4$ cuts $B{=}2$ p95 latency by $36\%$--$41\%$ within $0.25$ AP of uncapped \sba{}. \textbf{(b)}~Block size $M{\ge}16$ converges within the shaded $\pm 0.05$ AP parity band of offline \sba{} ($M{=}N$).}
\label{fig:caps-and-blocks}
\end{figure*}

\noindent\textbf{Bounding Tail Latency via Per-Image Call Caps.}
Because unconstrained sharing can assign up to $20$ crops to an unusually crowded frame, we test the cap guarantee of Proposition~\ref{prop:capped-greedy} by sweeping $H_{\max}\in\{1,2,3,4,6,\infty\}$ on RF-DETR-S at $B\in\{1,2\}$. The results are reported in Figure~\ref{fig:call-caps}. Enforcing $H_{\max}{=}4$ at $B{=}2$ cuts modeled p95 tail latency by $41\%$ on COCO ($24.9$ to $14.6$\,ms) and $36\%$ on CrowdHuman ($26.4$ to $17.0$\,ms) while keeping AP@100 within $0.25$ points of uncapped \sba{} ($56.31$, $51.57$, and $30.85$ on COCO, CrowdHuman, and VisDrone-DET). Multiplicative residual contraction front-loads each image's highest-gain windows into the first few picks, bounding tail latency without sacrificing collection accuracy.

\smallskip\noindent\textbf{Convergence of Bounded Micro-Batch Horizons.}
To quantify how many arriving frames are needed for batched streaming to match offline sharing, we partition validation streams into independent blocks of $M\in\{1,4,16,64,256,N\}$ frames with block allowance $\lfloor BM\rfloor$. The results are reported in Figure~\ref{fig:sba-blocks}. Whereas single-image quotas ($M{=}1$) trail offline \sba{} ($M{=}N$) by $-1.20$ and $-0.72$ AP@100 on average at $B{=}1,2$, micro-batches of $M{=}16$ frames converge within $-0.01$ to $-0.05$ AP@100 across all three benchmarks, and $M{\ge}64$ converges within $\pm0.02$ AP@100.

\makeatletter\def\Hy@backout#1{}\makeatother
\section{Full Related Work}
\label{app:full-related-work}

\noindent\textbf{High-Resolution Detectors and Tiny-Object Representations.}
Detecting small and occluded instances across wide scenes remains challenging because downsampling high-resolution frames into a fixed detector canvas strips away fine spatial detail~\cite{cheng2023towards}. Multi-scale architectures and training curricula seek to preserve small-object representations within a single forward pass, as Feature Pyramid Networks~\cite{lin2017feature} fuse top-down semantics with high-resolution stride maps and SNIP~\cite{singh2018analysis} restricts gradient updates to scale-matched instances. More recent detectors strengthen internal query routing and feature fidelity: DEIM~\cite{huang2025deim}, Mr.~DETR~\cite{zhang2025mr}, and PaQ-DETR~\cite{kang2026paq} refine dense matching, multi-route supervision, and quality-aware dynamic queries, while SET~\cite{sun2025set}, Gaussian position modeling~\cite{bian2025feature}, BDNet~\cite{guan2026bdnet}, and YOLO-Master~\cite{lin2026yolo} incorporate spectral enhancement, continuous positional priors, dual backbones, and mixture-of-experts routing. Despite improving feature utilization on a fixed input canvas, single-pass detectors remain bounded by physical pixel support when distant targets span only a few input pixels, and adopting any new architecture requires full retraining. \method{} is orthogonal to these designs because it keeps detector weights frozen and selectively magnifies under-resolved regions at inference time, improving all five detector families in Table~\ref{tab:family} and Appendix~\ref{app:revision-evidence}.

\smallskip\noindent\textbf{Uniform and Content-Adaptive Training-Free Tiling.}
Re-evaluating a frozen detector on magnified image crops recovers small targets without retraining~\cite{unel2019power}, yet every crop requires an extra forward pass. Uniform slicing pipelines trade compute for spatial resolution across the whole image: SAHI~\cite{akyon2022sahi} evaluates fixed overlapping tiles and merges projected boxes, while ASAHI~\cite{zhang2023adaptive} scales slice counts and dimensions to input resolution. Because both methods ignore scene content, they spend identical calls on empty backgrounds and dense crowds. Content-adaptive training-free planners instead guide crops using coarse predictions: DAHI~\cite{suarez2026dahi} samples windows from a coarse density field, DAASD~\cite{liu2026daasd} varies multi-scale tile granularity over a coarse detection grid, GOIS~\cite{muzammul2025gois} selects guided slices around coarse cues, and Dynamic Tiling~\cite{nguyen2023dynamic} places zoomed views around boundary-risk boxes with quadrant fallbacks. However, planning each image in isolation prevents these methods from transferring unused calls from sparse frames to crowded scenes or discounting multi-zoom overlap on a common scale. \method{} overcomes both limits by casting re-observation as collection-wide allocation over a calibrated deficit-closure objective, outperforming call-matched tiling in Table~\ref{tab:family} and upgrading the window pools of SAHI, DAHI, and Dynamic Tiling in Table~\ref{tab:similar-main} and Appendix~\ref{app:revision-evidence}.

\smallskip\noindent\textbf{Learned Zoom, Active Perception, and Dynamic Computation.}
A complementary literature learns where to zoom or how much computation to spend per input~\cite{han2022dynamic}. Dedicated zoom and density proposers train auxiliary modules alongside the detector: SNIPER~\cite{singh2018sniper} and AutoFocus~\cite{najibi2019autofocus} cascade multi-scale chips, ClusDet~\cite{yang2019clustered}, DMNet~\cite{li2020density}, and coarse-grained density networks~\cite{duan2021coarse} predict crowd cluster regions, and UFPMP-Det~\cite{huang2022ufpmp}, CZ~Det~\cite{meethal2023cascaded}, QueryDet~\cite{yang2022querydet}, and adaptive warps~\cite{wang2026adaptive} pack patches, refine cascaded crops, route sparse feature queries, or deform image grids. Active perception methods instead formulate sequential view selection via reinforcement learning or visual attention, including recurrent glimpse policies~\cite{mnih2014recurrent}, Dynamic Zoom-in~\cite{gao2018dynamic}, PatchDrop~\cite{uzkent2020learning}, AdaZoom~\cite{xu2022adazoom}, grounded multimodal cropping~\cite{zhao2026learning}, and ZoomEarth~\cite{liu2026zoomearth}. Internal dynamic-execution models route compute inside the network through spatial halting in SACT~\cite{figurnov2017spatially}, early exits in MSDNet~\cite{huang2018multi}, glance-and-focus classification~\cite{wang2020glance}, edge latency schedulers~\cite{jiang2021flexible}, patch recycling~\cite{qin2025no}, proposal gating in ROI-Gated SAHI~\cite{riyadh2026roi}, cross-attention entropy gating in ViCrop-Det~\cite{wang2026vicrop}, and decoder trajectory reliability in CQTR~\cite{shi2026cqtr}. These approaches require retraining, internal feature access, or sequential per-crop decisions that impede GPU batching and leave budgets unshared across images. \method{} operates strictly as a black-box wrapper on coarse boxes and scores, shares a conserved call budget across frames, and outperforms native proposer and gating pipelines at matched calls in Table~\ref{tab:native-main} and Appendix~\ref{app:latency}.

\smallskip\noindent\textbf{Submodular Coverage and Fast Spatial Window Search.}
Selecting crop windows to resolve uncovered objects under a call budget forms a cardinality-constrained coverage problem. Classical greedy maximization of monotone submodular functions guarantees a $(1-1/e)$ approximation~\cite{nemhauser1978analysis,krause2014submodular} accelerated by lazy queues~\cite{minoux1978accelerated} and CELF~\cite{leskovec2007cost}, whereas adaptive submodular policies~\cite{golovin2011adaptive} update posteriors sequentially after each observation. For spatial localization, Efficient Subwindow Search~\cite{lampert2008beyond} finds one globally optimal box per image via branch-and-bound, while summed-area tables~\cite{crow1984summed,viola2001rapid} evaluate rectangular sums in constant time. Yet generic set coverage neither calibrates how zoom closes scale-dependent deficit nor avoids quadratic scoring across multi-zoom lattices, single-window branch-and-bound does not support iterative residual updates, and sequential policies block parallel batching. \method{} addresses these challenges through an additively separable geometric closure surrogate that lets Shared-Budget Allocation (\sba{}) refresh only the popped image in a global max-heap prior to crop execution, while Lattice Window Sweep (\lws{}) uses dual-origin 2D difference arrays in Appendix~\ref{app:discrete-inference} to score multi-zoom grids in linear lattice time.

\smallskip\noindent\textbf{Crowded-Scene Suppression and Multi-View Fusion.}
Projecting overlapping crops back into native coordinates introduces duplicate boxes and seam-truncated fragments that compete for bounded top-$K$ evaluation slots~\cite{dollar2012pedestrian}. Single-view post-processors resolve intra-view overlap: greedy NMS~\cite{neubeck2006efficient}, Soft-NMS~\cite{bodla2017softnms}, and Adaptive NMS~\cite{liu2019adaptive} decay or threshold overlapping boxes, learned NMS~\cite{hosang2017learning} and CrowdDet~\cite{chu2020detection} decode crowded proposals end-to-end, and WBF~\cite{solovyev2021weighted} and MoCaE~\cite{oksuz2023mocae} combine box coordinates or mixture confidences without modeling crop-border truncation. Multi-view heuristics such as greedy slice union in SAHI~\cite{akyon2022sahi}, border-band pruning in AutoFocus~\cite{najibi2019autofocus} and DAHI~\cite{suarez2026dahi}, and dense map fusion in SCAHI~\cite{becker2026scahi} filter views independently of the planner's credit region and allow abundant crowded classes to evict rare categories at cap $K$. Positive-Region Fusion (\prf{}) resolves both failure modes from sparse boxes and scores by enforcing the shared interior region $\mathcal P_c$ and scale gates used in planning, downweighting superseded coarse predictions as fallbacks, and applying order-preserving class-mass squeezing in Appendix~\ref{app:fusion-order} and Appendix~\ref{app:factorial-controls}. Table~\ref{tab:method-taxonomy} compares these paradigms across five deployment properties.

\begin{table*}[!t]
\centering\small
\caption{\textbf{Capability Taxonomy Across Detection and Re-observation Paradigms.} Comparison across five deployment properties, where \cmark{} denotes full support, \pmark{} denotes partial support or an auxiliary module, \xmark{} denotes unsupported, and -- denotes not applicable to single-pass detectors.}
\label{tab:method-taxonomy}
\begin{nicetable}
\setlength{\tabcolsep}{3.8pt}
\begin{tabularx}{\textwidth}{l>{\raggedright\arraybackslash}Xccccc}
\toprule
\rowcolor{ntHeader}
Paradigm & Representative Works & \makecell{Frozen\\Weights} & \makecell{Black-Box\\Inputs} & \makecell{Shared\\Budget} & \makecell{Aligned\\Fusion} & \makecell{Parallel\\Batching} \\
\midrule
High-res.\ detectors & FPN~\cite{lin2017feature}, DEIM~\cite{huang2025deim}, SET~\cite{sun2025set}, BDNet~\cite{guan2026bdnet}, PaQ-DETR~\cite{kang2026paq}, YOLO-Master~\cite{lin2026yolo} & \xmark & \xmark & \xmark & -- & \cmark \\
Uniform tiling & Power of Tiling~\cite{unel2019power}, SAHI~\cite{akyon2022sahi}, ASAHI~\cite{zhang2023adaptive} & \cmark & \cmark & \xmark & \xmark & \cmark \\
Adaptive tiling & DAHI~\cite{suarez2026dahi}, DAASD~\cite{liu2026daasd}, Dynamic Tiling~\cite{nguyen2023dynamic} & \cmark & \cmark & \xmark & \pmark & \cmark \\
Proposal / attn.\ gating & ROI-Gated SAHI~\cite{riyadh2026roi}, ViCrop-Det~\cite{wang2026vicrop} & \pmark & \xmark & \xmark & \xmark & \cmark \\
Learned zoom heads & ClusDet~\cite{yang2019clustered}, DMNet~\cite{li2020density}, QueryDet~\cite{yang2022querydet}, CZ~Det~\cite{meethal2023cascaded} & \xmark & \xmark & \xmark & \pmark & \pmark \\
Sequential RL zoom & Dynamic Zoom-in~\cite{gao2018dynamic}, PatchDrop~\cite{uzkent2020learning}, AdaZoom~\cite{xu2022adazoom}, ZoomEarth~\cite{liu2026zoomearth} & \xmark & \xmark & \xmark & \xmark & \xmark \\
\midrule
\rowcolor{ntMethod}
\textbf{Shared re-observation} & \textbf{\method{} (Ours)} & \cmark & \cmark & \cmark & \cmark & \cmark \\
\bottomrule
\end{tabularx}
\end{nicetable}

\end{table*}

\section{More Qualitative Results}
\label{app:current-qualitative}

\begin{figure*}[!t]
\centering
\includegraphics[width=0.98\textwidth]{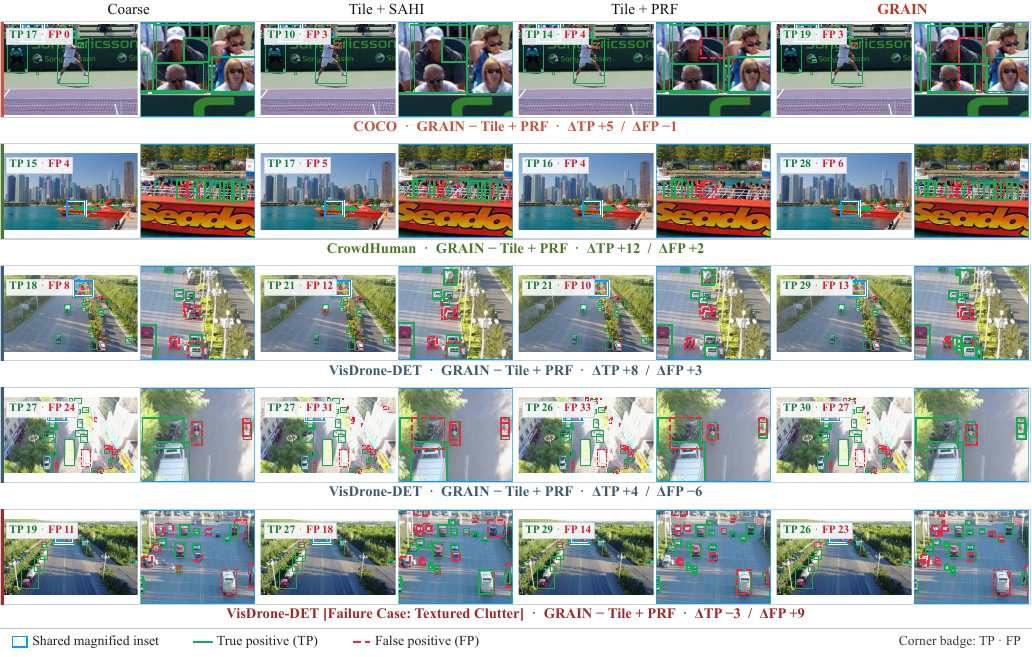}
\caption{\textbf{Complete Detection Pipelines at $B{=}2$ (Frozen RF-DETR-S).} Overview and shared magnified inset (blue box) with full-image TP/FP corner badges across Coarse, Tile+SAHI, Tile+\prf{}, and \method{}. Row~5 highlights an adverse textured VisDrone-DET scene.}
\label{fig:qualitative-examples}
\end{figure*}

\begin{figure*}[!t]
\centering
\includegraphics[width=0.98\textwidth]{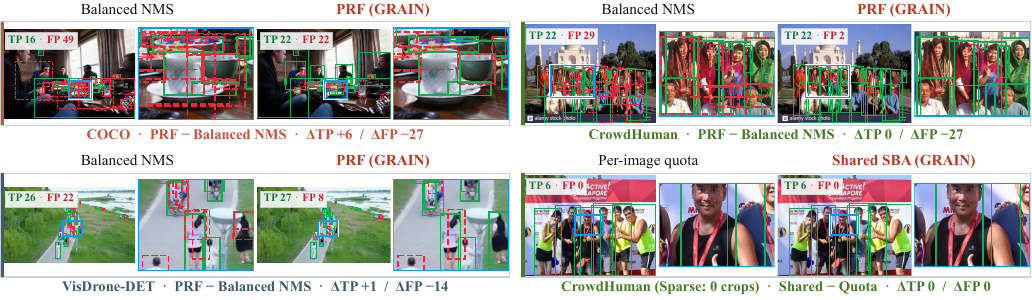}
\caption{\textbf{Isolated Module Controls (Part I): Seam Suppression and Sparse Skipping ($B{=}2$, RF-DETR-S).} Balanced NMS vs.\ \prf{} across COCO, CrowdHuman, and VisDrone-DET, and per-image quota vs.\ shared \sba{} on a sparse CrowdHuman frame ($0$ crops).}
\label{fig:qualitative-controls}
\label{fig:qualitative}
\label{fig:current-pipeline}
\end{figure*}

\begin{figure*}[!tp]
\centering
\includegraphics[width=0.98\textwidth]{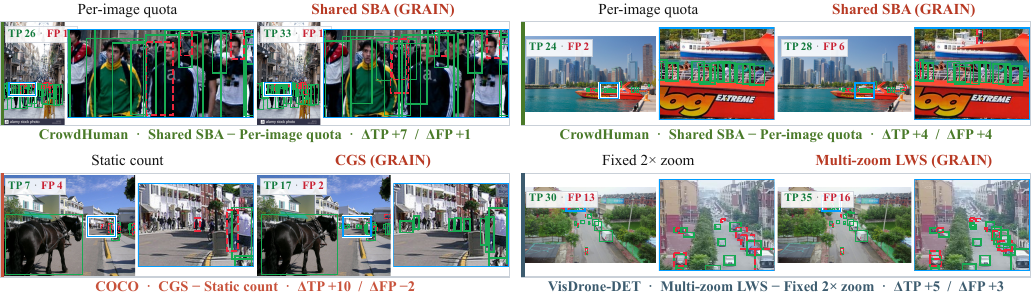}
\caption{\textbf{Isolated Module Controls (Part II): Crowded Reallocation, Residual Discounting, and Multi-Zoom Search ($B{=}2$, RF-DETR-S).} Per-image quota vs.\ shared \sba{} on dense CrowdHuman crowds (top), static counts vs.\ \cgs{} on COCO (bottom left), and fixed $2{\times}$ zoom vs.\ multi-zoom \lws{} on VisDrone-DET (bottom right).}
\label{fig:qualitative-controls-alloc}
\end{figure*}

\begin{figure*}[!tp]
\centering
\includegraphics[width=0.98\textwidth]{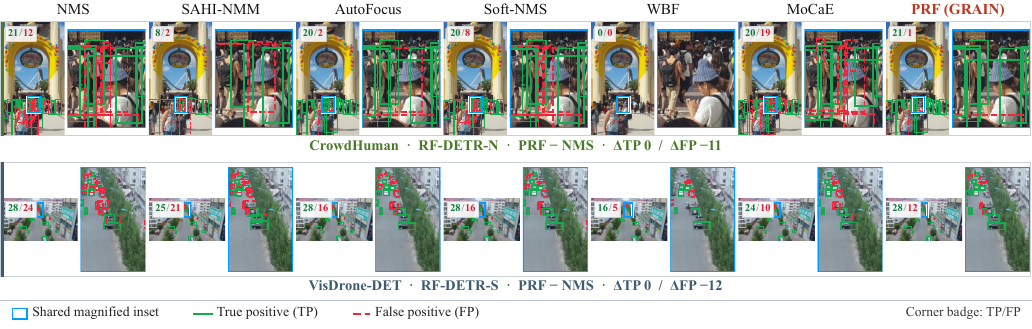}
\caption{\textbf{Fixed-View Comparison of Seven Post-Inference Fusion Rules ($B{=}1$).} Identical coarse and crop predictions on CrowdHuman (RF-DETR-N) and VisDrone-DET (RF-DETR-S).}
\label{fig:qualitative-fusion}
\label{fig:qualitative-panels}
\end{figure*}

To visualize how planning and fusion decisions govern instance-level precision and recall, we render qualitative comparisons across end-to-end pipelines, isolated module controls, and fixed-view merging rules under class-aware greedy matching at confidence $\ge0.3$, IoU $\ge0.5$, and $K{=}100$, omitting ignored regions. The results are shown in Figures~\ref{fig:qualitative-examples}--\ref{fig:qualitative-fusion}.

\smallskip\noindent\textbf{End-to-End Pipelines and Adverse Clutter.}
To compare end-to-end detection quality across domains, Figure~\ref{fig:qualitative-examples} evaluates Coarse, Tile+SAHI~\cite{akyon2022sahi}, Tile+\prf{}, and \method{} on frozen RF-DETR-S at $B{=}2$. Upgrading uniform Tile from greedy SAHI union to \prf{} restores suppressed crowd true positives, and replacing uniform Tile with \method{} redirects calls onto unresolved clusters to increase net true positives across COCO, CrowdHuman, and VisDrone-DET. Conversely, Row~5 isolates an adverse VisDrone-DET intersection where high-zoom crops over textured road markings trigger background false positives, illustrating the eviction and rank-drift channels $\mathcal E_{\mathrm{loss}}$ and $\mathcal D_{\mathrm{rank}}$ in Eq.~\eqref{eq:ap-three-channel}.

\smallskip\noindent\textbf{Module Controls and Fixed-View Fusion.}
To isolate individual components visually, Figures~\ref{fig:qualitative-controls} and~\ref{fig:qualitative-controls-alloc} vary one planning or fusion mechanism at a time at $B{=}2$: \prf{} eliminates seam-truncated duplicate boxes retained by balanced NMS, shared \sba{} skips resolved sparse frames ($0$ crops) to fund crowded scenes, \cgs{} steers second crops away from already covered clusters, and multi-zoom \lws{} resolves distant vehicles missed at fixed $2{\times}$ zoom. On identical $B{=}1$ views in Figure~\ref{fig:qualitative-fusion}, plain NMS and MoCaE~\cite{oksuz2023mocae} retain border duplicates while SAHI~\cite{akyon2022sahi} and WBF~\cite{solovyev2021weighted} suppress crowded true positives, whereas \prf{} preserves dense recall while pruning seam fragments.

\section{Societal Impacts}
\label{app:societal-impacts}
\label{app:audit-boundaries}

\noindent\textbf{Positive Applications and Broader Impacts.}
By improving small- and crowded-object recall on unmodified frozen detectors without retraining or architectural changes, \method{} lowers inference compute and energy requirements for resource-constrained aerial inspection, traffic safety monitoring, infrastructure maintenance, and search-and-rescue assessment, enabling practitioners to deploy high-resolution perception on standard commodity GPUs.

\smallskip\noindent\textbf{Ethical Considerations and Potential for Misuse.}
Because \method{} automatically directs high-zoom re-observation onto dense crowds and distant small targets, it increases the effective spatial resolution at which pedestrians and vehicles are localized in wide-area imagery. If deployed without oversight in public camera networks, higher-recall pedestrian localization could facilitate automated crowd surveillance or intrusive tracking, and may propagate any subgroup or environmental biases already present in the underlying frozen detector checkpoint.

\smallskip\noindent\textbf{Mitigation and Operational Safeguards.}
\method{} introduces no new learned weights or biometric identification capabilities beyond the base detector and operates strictly within user-specified call budgets and per-image caps. Practical deployments in human-centric environments should pair budgeted re-observation with domain-specific privacy governance, strict access controls, and subgroup error evaluation on representative local scenes before deployment.

\end{document}